\documentclass[compsoc,conference,a4paper,10pt,times]{IEEEtran}
\IEEEoverridecommandlockouts
\usepackage{amsmath,amssymb,amsfonts}
\usepackage{algorithm}
\usepackage[noend]{algpseudocode}
\usepackage{mathtools}
\usepackage{array}
\usepackage{multirow}
\usepackage{booktabs}
\usepackage{xcolor}
\usepackage{cite}
\usepackage{graphicx}
\usepackage{textcomp}
\usepackage{bmpsize}
\usepackage{lipsum}
\usepackage[colorlinks=true,urlcolor=black]{hyperref}

\usepackage[font=normalsize]{caption}
\usepackage{subcaption}
\usepackage{url}
\usepackage{amsthm}
\usepackage{bbm}
\usepackage{bm}

\newtheorem*{remark}{Remark}

\newtheorem*{definition}{Definition}

\newcommand{\defeq}{\ensuremath{\triangleq}}

\usepackage{stfloats}
\usepackage{lscape}
\usepackage{amsfonts}
\usepackage{amssymb}
\usepackage{times}
\usepackage{epsfig}
\usepackage{latexsym}
\usepackage{epstopdf}
\usepackage{verbatim}
\usepackage{units}
\usepackage{placeins}
\usepackage{afterpage}
\usepackage{dsfont}
\usepackage{soul}
\usepackage{multicol}
\usepackage[cmintegrals]{newtxmath}
\usepackage{anyfontsize}
\usepackage{amsmath}

\usepackage[normalem]{ulem}
\DeclareMathOperator*{\argmax}{arg\,max}
\useunder{\uline}{\ul}{}
\def\BibTeX{{\rm B\kern-.05em{\sc i\kern-.025em b}\kern-.08em
    T\kern-.1667em\lower.7ex\hbox{E}\kern-.125emX}}
\begin{document}

\title{Aggressive, Imperceptible, or Both: \\
Architecture-Aware Hybrid Byzantines\\ 
in Federated Learning  }
\author{
\IEEEauthorblockN{
Emre Özfatura$^{1*}$,
Kerem Özfatura$^{2*}$,
Baturalp Büyükateş$^{4}$,
Mert Coşkuner$^{1}$,
Alptekin Küpçü$^{2}$,
Deniz Gündüz$^{3}$
}

\IEEEauthorblockA{$^{1}$Sabancı University}
\IEEEauthorblockA{$^{2}$Koç University}
\IEEEauthorblockA{$^{3}$Imperial College London}
\IEEEauthorblockA{$^{4}$University of Birmingham}
}

\maketitle

\def\thefootnote{*}
\footnotetext{These authors contributed equally to this work.}
\def\thefootnote{\arabic{footnote}}

\maketitle

\begin{abstract}
In federated learning (FL), profiling and verifying each client is inherently difficult, which introduces a significant security vulnerability: malicious clients, commonly referred to as \emph{Byzantines}, can degrade the accuracy of the global model by submitting poisoned updates during training. To mitigate this, the aggregation process at the parameter server must be robust against such adversarial behaviour. Most existing defences approach the Byzantine problem from an \emph{outlier detection} perspective, treating malicious updates as statistical anomalies and ignoring the internal structure of the trained neural network (NN). Motivated by this, this work highlights the potential of leveraging side information tied to the NN architecture to design stronger, more targeted attacks. In particular, inspired by insights from sparse NNs, we introduce a hybrid sparse Byzantine attack. The attack consists of two coordinated components: (i) A sparse attack component that selectively manipulates parameters with higher sensitivity in the NN, aiming to cause maximum disruption with minimal visibility; (ii) A slow-accumulating attack component that silently poisons parameters over multiple rounds to evade detection. Together, these components create a strong but imperceptible attack strategy that can bypass common defences. We evaluate the proposed attack through extensive simulations and demonstrate its effectiveness against eight state-of-the-art defence mechanisms. Our code is available in \url{https://github.com/CRYPTO-KU/FL-Byzantine-Library}.
\end{abstract}


\section{Introduction}
Federated learning (FL) \cite{FedAVG1} has become the de-facto {\em collaborative learning} framework, where a large number of clients iteratively optimize a shared neural network (NN) by utilizing their local data. In each FL round, following local optimisation, participants seek consensus on the global model through the exchange of local models. While local optimisation helps protect data privacy by keeping personal data on-device, the consensus phase reduces generalization error by aggregating knowledge across clients. In this work, we focus on the {\em centralized} and {\em synchronous} FL setup \cite{FedAVG1,fedAVG_2}, where a central parameter server (PS) orchestrates the consensus phase by executing an aggregation policy for the local models received from the participating clients. Upon aggregation, the new model is conveyed to the clients for the next round of local optimisation/training.

FL is designed to enable collaborative model training across a large number of distributed clients. However, profiling and verifying every participant is often infeasible at scale. Moreover, in settings where the data is non-identically distributed among the clients, identifying honest and malicious behaviour becomes very hard. This opens the door for malicious clients who may attempt to undermine the learning process—either independently or through collusion with other adversarial participants. In the literature, such behaviour is referred to as a \emph{poisoning attack} \cite{backdoor_FL,AFLAL,FL_limitations,backdoor_norm}, where the malicious clients send modified models to the PS in compliance with a predefined attack strategy. These attacks can be {\em targeted } \cite{DBA,backdoor_FL,AFLAL}, where the objective is to cause the model to fail on specific tasks, such as miss-classification of certain labels, or {\em untargeted} \cite{Krum,IPM}, where the goal is to degrade the model’s overall performance across tasks \cite{AFLAL,FL_limitations,backdoor_norm,ALIE}. Such attacks are not always visible during training; the global model may appear to perform well on training and validation data but may lack generalization at test time. In particular, the global model may exhibit poor performance on input samples containing certain triggering patterns, a behaviour commonly associated with \textit{backdoor attacks} \cite{backdoor_FL}. In this work, we focus on \textit{untargeted model poisoning} attacks, often known as \emph{Byzantine attacks}.

In FL, to mitigate the detrimental impact of malicious clients, a fundamental defence strategy involves the design of a robust aggregator that can neutralize poisoned models \cite{Krum,RFA,Trimmed_mean}. Many of these methods treat malicious models as outliers and suggest filtering such statistically suspicious updates out before applying conventional aggregation operations such as averaging \cite{Bulyan,Krum,cosine_profiling,buyukates_fedsecurity,han2023kick}. Another group of approaches modify the aggregation rule itself to better represent benign clients—for example, using geometric or trimmed means \cite{Trimmed_mean,mean-median}. Rather than eliminating suspicious model updates, some methods attempt to sanitize all client models prior to aggregation, as suggested in \cite{CC,decentCC,ByzSGD_def,PODC-BYZ_DSGD}. Other more complicated defence strategies either utilize extra data at the PS \cite{FLtrust,zeno_def}, or profile the clients over time by assigning trust scores \cite{cosine_profiling, AFLAL}. Recent works have also explored a game-theoretic approach to dynamically select a different robust aggregator at each iteration \cite{game_theory}. 

From the attacker's perspective, an attack should be effective yet imperceptible \cite{ALIE,ndss2022_Attack,LMPA,game_theory}. To achieve this, adversaries often take advantage of the curse of dimensionality—the high dimensionality of model parameters—as well as the variation across benign client models, which arises from the heterogeneous data distributions across clients. As this statistical variation increases, so does the adversarial feasible attack space, enabling Byzantine clients to blend in with benign updates and avoid being flagged as outliers. In response, variance reduction techniques, originally developed to stabilize distributed learning, have been shown to effectively mitigate Byzantine threats by constraining the feasible attack space \cite{Bucketing,CC,sec.vr1,sec.vr2,sec.vr3,sec.vr4,sec.vr5}. One widely adopted strategy is to employ momentum stochastic gradient descent (SGD) as a local optimizer and to perform aggregation over the momentum terms. This approach has demonstrated increased robustness to Byzantine behaviour \cite{CC,rop,momentum_def,Byz.momentum}.

While many effective Byzantine attack strategies exist, a single, universally effective attack that simultaneously defeats a wide range of defences remains elusive. As we discuss later, some attacks are particularly effective against defences that detect index-wise outlier values, while others are tailored to circumvent mechanisms that utilise geometric distance metrics. Existing Byzantine schemes typically ignore the architecture of the NN under attack and tend to succeed only against specific defence classes. 

In this work, we explore whether it is possible to \textbf{design a universally effective yet computationally efficient Byzantine attack strategy} or not. 
We notice that depending on the architecture, \emph{certain weights may be more sensitive to perturbations}, and such structural prior knowledge can be leveraged when designing the attack. To the best of our knowledge, this is the first work that \textbf{explicitly exploits architectural characteristics of the target NN to guide the design of Byzantine attacks}. We propose a powerful and stealthy method, named the \emph{hybrid sparse Byzantine attack}, that exploits side information extracted from the target network architecture to craft perturbations that are both highly effective and difficult to detect. 

Our contributions are summarized as follows:
\begin{itemize}
\item Through extensive numerical experiments on state-of-the-art Byzantine attack and defence techniques, we visualize the current Byzantine attack landscape by identifying both the strengths and limitations of existing frameworks. Our goal is to design a universal versatile Byzantine framework.
\item We introduce a novel hybrid Byzantine attack strategy composed of two complementary components: one tailored to evade defences that rely on index-wise anomaly detection, and the other aimed at circumventing geometric distance-based defences. This hybrid design ensures effectiveness across a wide range of defence mechanisms, without requiring prior knowledge of the deployed defence mechanism.
\item Unlike the existing Byzantine attack designs oblivious to the underlying network topology, we incorporate side information extracted from the NN architecture. Using a network pruning framework, we identify weights that are potentially more sensitive to perturbations and leverage this information to distribute perturbation budget accordingly, which enable us to design attacks that are both stronger and more imperceptible.
\item We conduct extensive simulations across various NN architectures, datasets, and defence mechanisms. Our results show that the proposed attack can reduce test accuracy up to 55\% in the case of independent and identically distributed (IID) settings, and in non-IID scenarios, can even cause the global model to diverge entirely.
\end{itemize}

\section{Background and Related Work}\label{s:back}

\textbf{Notation:}
We use $\textbf{bold}$ case to denote vectors, e.g., $\boldsymbol{v}$, and capital calligraphic letters, e.g., $\mathcal{V}$, to denote sets. When we have an ordered set of vectors $\mathcal{V}=\left\{ \boldsymbol{v}_{1},\ldots,\boldsymbol{v}_{k}\right\}$, we use subscript index to identify the $i^{th}$ vector in the set, with $i\in[k]$, and  use double subscript $\mathbf{v}_{i,t}$, particularly when it is changing over time/iteration. For slicing operation, we use $[\cdot]$, such as $\mathbf{v}[j]$ to denote the $j^{th}$ index of a vector.  We use $|| \ \cdot \ ||_{p}$ to denote the $p$-norm of a vector, which, without a subscript $|| \ \cdot \  ||$ refers $l_{2}$ to norm. We use $< \cdot \ , \cdot  >$ for the inner product between two vectors. In Table \ref{tab:notations}, we list the widely used variables in this work.

\subsection{System Model}


\begin{table*}[t]
\caption{Widely used variables in this work.}
\label{tab:notations}
\centering
\begin{tabular}{|l|l|}
\hline
\textbf{Notation} & \textbf{Description} \\ \hline
$\mathcal{K}_{b}$, $\mathcal{K}_{m}$, and $\mathcal{K}= \mathcal{K}_{b} \cup \mathcal{K}_{m}$ & Set of benign, malicious, and all clients \\ \hline
$k_{b}$, $k_{m}$, and $k$ & Number of benign, malicious, and all clients \\ \hline
$\boldsymbol{\theta}_{i,t}$, $\mathbf{g}_{i,t}$, and $\mathbf{m}_{i,t}$ & Model, gradient, and momentum vector of client $i$ at iteration $t$ \\ \hline
$\tilde{\mathbf{m}}_{t}$, $\hat{\mathbf{m}}_{t}$, and $\bar{\mathbf{m}}_{t}$ & Reference, aggregated, and benign consensus momentum at iteration $t$ \\ \hline
$\delta$, $\mathbf{c}$ & Sparsity ratio and the corresponding mask vector \\ \hline
\end{tabular}
\end{table*}


The goal in FL is to solve the following parametrized optimisation problem over $k$ clients in a distributed manner:
\begin{equation}
\min_{\boldsymbol{\theta}\in\mathbb{R}^{d}} f(\boldsymbol{\theta})= \frac{1}{k}\sum^{k}_{i=1}\underbrace{\mathds{E}_{\zeta_{i} \sim \mathcal{D}_{i}}f(\boldsymbol{\theta},\zeta_{i})}_{\mathrel{\mathop:}=f_{i}(\boldsymbol{\theta})},\label{DSO}
\end{equation}
where $\boldsymbol{\theta}\in\mathbb{R}^{d}$ denotes the model parameters, e.g., weights of an NN, $\zeta_i$ is a random sample drawn from the local data distribution $\mathcal{D}_{i}$ of client $i$, and $f$ is the empirical loss function associated with the learning task. This problem is typically solved in an iterative manner. At each iteration, each client computes an updated local model $\boldsymbol{\theta}_{i}$ that minimizes its local loss function  $f_{i}(\boldsymbol{\theta}_{i})$, often by using SGD. The local models are then sent to the PS, which aggregates them to update the global model $\boldsymbol{\theta}$.

We note that there are multiple practical implementations of this two-step distributed optimization framework—comprising local optimization and global consensus—depending on several design choices. These include the selected local optimization method (such as SGD \cite{SGD.polyak}, Adam \cite{adamOPT}, or momentum SGD \cite{SGD.init}), the use of regularization strategies during local training \cite{FedADC,fedprox,fedDyn}, the aggregation strategy employed at the PS, the communication format between clients and the PS (including transmitting the full model \cite{FedDC}, model differences \cite{FedADC}, or momentum terms \cite{CC,momentum_def}), and how the global and local updates are coordinated \cite{FedADC,FedDC,SlowMo,scaffold}.

In the scope of this work, we focus on a specific setup, where \textbf{each client employs momentum SGD} for local optimisation. Each client first computes the gradient for a randomly sampled minibatch $\zeta_{i}$ as follows: 
\begin{equation} 
\mathbf{g}_{i}\gets \nabla_{\boldsymbol{\theta}}f_{i}(\boldsymbol{\theta},\zeta_{i}),
\end{equation} 
where $\boldsymbol{\theta}$ is the most recent global, i.e., consensus, model received from the PS. The local gradient $\mathbf{g}_{i}$ is then used to update the momentum term $\mathbf{m}_{i}$: 
\begin{equation} 
\mathbf{m}_{i} = (1 - \beta) \mathbf{g}_{i} + \beta \mathbf{m}_{i},
\end{equation} 
where $\beta$ is the momentum hyperparameter. Each client then sends its local momentum term to the PS for aggregation.

Although the techniques presented in this work can be extended to other federated optimisation setups, we design our attack scheme in this particular framework for two key reasons. First, the use of momentum helps accelerate gradient vectors in the right directions, thus \textbf{improving convergence speed} \cite{SGD.init,SGD.nesterov,SGD.polyak}. Second and more importantly, aggregating local momentum terms to compute the global model update contributes to robustness against Byzantine attacks. This is because \textbf{momentum aids in variance reduction}, effectively narrowing the feasible space for adversarial perturbations, thereby \textbf{making the attack problem more challenging} for Byzantine clients \cite{CC,rop,momentum_def,Byz.momentum,Bucketing,SCC}.

\textbf{Adversary Model:}
We assume that the adversary controls up to $k_{m}$ malicious clients out of the total $k$ clients (thus, the remaining $k-k_m$ clients are honest). Consistent with the standard threat model in the literature, we consider the case where the number of malicious clients is less than the number of benign ones, i.e., $(k_{m}/k) < 0.5$; which is also considered as an upper bound for Byzantine tolerance \cite{byzantine_orig}. Indeed, in our experimental evaluation, we keep this ratio much smaller. Following prior works \cite{ndss2022_Attack,backdoor_FL,ALIE,AFLAL,LMPA,IPM}, we assume that the adversary has access to the global model parameters broadcast at each iteration and can directly manipulate the gradients on compromised (malicious) clients. Additionally, similar to other works, \cite{ALIE,ndss2022_Attack,rop} we assume that the adversary can accurately predict the gradient of the benign clients and use those values to compute local momentums. We include an ablation study in \ref{sec:no_known_grad}, where malicious clients have no information on the benign client gradients and generate their attack based on their own gradients. Overall, we find that Byzantines can generate equally strong attacks, if not stronger, using only their local data. 

In our solution, Byzantines, i.e., malicious clients, construct their attacks in an aggregator-agnostic manner. That is, the Byzantine clients do not know which specific robust aggregation or defence mechanism is employed by the PS. Thus, the attack we design must be universally effective against a wide range of known defence mechanisms without requiring any aggregator-specific tuning or calibration.

The overall framework in Algorithm~\ref{code:robustFL} consists of Byzantine clients executing an attack strategy, denoted by $Attack(\cdot)$, and a robust aggregator, denoted by $AGG(\cdot)$, implemented at the PS. The attack function $Attack(\cdot)$ can be viewed as a mapping: 
\begin{equation} 
Attack(\cdot): \mathcal{H}_{t} \xrightarrow{} \left\{\mathbf{m}^{}_{i,t} \right\}_{i \in \mathcal{K}_{m}} \ \in \mathbb{R}^{d}, 
\end{equation} 
where $\mathcal{H}_{t}$ represents the history, i.e., all the information available to malicious clients up to iteration $t$, and $\mathcal{K}_{m}$ denotes the set of malicious clients.

\begin{algorithm}[t]
    \caption{FL with Robust Aggregator and Byzantines }\label{code:robustFL}
    \footnotesize
	\begin{algorithmic}[1]
	\State\textbf{Input:} Learning rate $\eta $,  $AGG(\cdot)$ , $Attack(\cdot)$
	\State \textbf{Output:} end consensus model: $\boldsymbol{\theta}_{T}$
		\For{$t=1,\ldots,T$}
		\State \textbf{Client side:}
		\For{$i=1,\ldots,k$} in parallel
		\State Receive: $\boldsymbol{\theta}_{t-1}$ from PS
		\If {$i\in\mathcal{K}_{b}$}
		\State Compute gradient estimate: $\mathbf{g}_{i,t}\gets \nabla_{\boldsymbol{\theta}}f_{i}(\boldsymbol{\theta}_{t-1},\zeta_{i,t})$
		\State Update Momentum:  $\mathbf{m}_{i,t} = (1-\beta) \mathbf{g}_{i,t} + \beta \mathbf{m}_{i,t-1}$
		\Else
		\State $\mathbf{m}_{i,t} \gets Attack(\mathcal{H}_{t})$
		\EndIf
		\EndFor
		\textbf{Server side:}
		\State{Aggregate local updates: $\tilde{\mathbf{m}}_{t}\gets AGG(\mathbf{{m}}_{1,t}, \ldots, \mathbf{m}_{k,t})$}
		\State Update the model: $\boldsymbol{\theta}_{t} \gets \boldsymbol{\theta}_{t-1} -  \eta_{t}\tilde{\mathbf{m}}_{t}$
		\EndFor
	\end{algorithmic}
\end{algorithm}

\subsection{Analysing Existing Defence Methods} \label{Sec:Defs}
Various defence mechanisms against Byzantines have been introduced to ensure the reliability of the learned model. Broadly speaking, these frameworks can be grouped into two categories: \textit{elimination} and \textit{sanitization}. Defence methods with elimination detect Byzantine updates under the assumption that they exhibit outlier behaviour compared to benign updates. These methods utilize various metrics for outlier detection such as geometric distance, index-wise statistics, and angular variance. Upon the elimination of the outlier values, the remaining elements (vectors or indices) are aggregated. The key drawbacks are the elimination of benign updates due to misclassification and the fact that these methods often require that the server knows the number of Byzantines. Alternatively, defence strategies in the second category utilize all updates but sanitize them to neutralize the detrimental impact of Byzantines through different mechanisms such as majority voting and geometric median. We list the defence mechanisms we consider in Table \ref{tab:def-comperisons}.

\begin{table*}[]
\centering
\caption{Defence methods we consider in this work.}
\label{tab:def-comperisons}
\begin{tabular}{@{}lll@{}}
\toprule
\textbf{\textbf{Defence framework}} & \textbf{\textbf{Identification Metric}} & \textbf{\textbf{Defence Technique}} \\ \midrule
Centered Clipping (CC) \cite{CC} & Euclidean distance & Sanitization (Clipping) \\
Krum \& Multi-Krum \cite{Krum} & Euclidean distance & Elimination \\
Bulyan \cite{Bulyan} & Euclidean distance & Elimination \\
Trimmed Mean \cite{Trimmed_mean} & Index-wise statistics  & Elimination \\
Centered Median \cite{Trimmed_mean} & Index-wise statistics  & Elimination (Index-wise median) \\
SignSGD \cite{SignSGD,signSGD_2} & Index-wise statistics & Sanitization (Majority voting) \\
RFA \cite{RFA} & Euclidean distance & Sanitization (Geometric median) \\
GAS \cite{GAS} & Euclidean distance (sub-vector) & Elimination (partitioned aggregation) \\
 \bottomrule
\end{tabular}
\end{table*}

In what follows, we give an overview of the elimination-based defences that use Euclidean distance and index-wise statistics, as they are relevant to our proposed attack design.


\textbf{(Multi) M-Krum:} The underlying idea behind the M-Krum mechanism \cite{Krum} is to identify where the benign vectors accumulate in close proximity. Under the assumption of $k_m$ Byzantines, considered as outliers, the first step is to form a set of neighbouring clients $\mathcal{S}_{i}$, for each vector $\mathbf{m}_{i}$, with cardinality $k^{\star}=k - k_{m}+ 2$, such that

\begin{equation}
   \mathcal{S}^{\star}_{i}= \underset{\mathcal{S},\vert\mathcal{S}\vert=k^{\star}}{\mathrm{argmin}} \sum_{j \in \mathcal{S}}   \left \| \mathbf{m}_{i} - \mathbf{m}_{j} \right \|^{2}.      
\end{equation}
Then, M-Krum identifies the vector $\mathbf{m}_{i}$ that has the minimum average distance to the vectors in its closest neighbourhood $\mathcal{S}^{\star}_{i}$ as the true center for the benign vectors, i.e.,
\begin{equation}
\mathrm{argmin}_{i \in \mathcal{K}}~ \sum_{j \in \mathcal{S}^{\star}_{i}}   \left \| \mathbf{m}_{i} - \mathbf{m}_{j} \right \|^{2}
\end{equation}
Multi-Krum is built on the same principle but chooses $n$ vectors instead of a single one, and outputs their average.

\textbf{Centered Clipping (CC):} Robust aggregators based on outlier elimination assume prior knowledge of the number of Byzantines and still risk misclassifying or under-utilizing benign clients \cite{Bucketing, ALIE}. To address this, CC \cite{CC} sanitizes all model updates by projecting each update $\mathbf{m}$ into a ball of radius $\tau$ centred at the reference model $\tilde{\mathbf{m}}$.
Accordingly, the clipping function can be formulated as
\begin{equation}
f_{CC}(\mathbf{m}\vert \tilde{\mathbf{m}},\tau ) = \tilde{\mathbf{m}} + \min\left\{1,\frac{\tau}{\vert\vert \tilde{\mathbf{m}} - \mathbf{m} \vert\vert}\right\} (\mathbf{m}-\tilde{\mathbf{m}}).
\end{equation}
By setting the previous aggregate $\hat{\mathbf{m}}_{t-1}$ as the reference model $\tilde{\mathbf{m}}_{t}$, CC can sanitize poisoned updates; however, it fails when Byzantines adapt perturbations to $\tilde{\mathbf{m}}_{t}$ \cite{rop}. We further discuss additional weaknesses of CC in Section \ref{sec:alie}.

\begin{algorithm}[t]
    \small
    \caption{Aggregation with Centered Clipping (CC)}
	\begin{algorithmic}[1]
     \State \textbf{Inputs:} ${\tilde{\mathbf{m}}}_{t}$, $\left\{\mathbf{m}_{i,t}\right\}_{i\in\mathcal{K}},\tau$
	 \For{$i=1,\ldots,k$} in parallel
     \State $\tilde{\mathbf{m}}_{i,t} = f_{CC}(\mathbf{m}_{i,t}\vert \tilde{\mathbf{m}}_{t},\tau)$
     \EndFor
     \State $\hat{\mathbf{m}}_{t}=\frac{1}{k}\sum_{i\in\mathcal{K}}\tilde{\mathbf{m}}_{i,t}$
	\end{algorithmic}
	\label{code:CC}
\end{algorithm}

\textbf{Robust-Federated-Averaging (RFA):} RFA is a geometric median-based robust aggregation method \cite{RFA}, where the consensus model $\mathbf{m}^{\star}$ is obtained as the solution to the following: 
\begin{equation}
    \mathrm{RFA}(\mathbf{m}_{1}, \dotso, \mathbf{m}_{n}) =\underset{\mathbf{m}^{\star}}{argmin}\sum\limits_{i=1}^{n}|| \mathbf{m}^{\star} - \mathbf{m}_{i}||_{2}.
\end{equation}
 It reduces the impact of contaminated updates by using the geometric median in place of the weighted arithmetic mean.

 
\textbf{SignSGD:} SignSGD \cite{SignSGD,signSGD_2} reduces the communication overhead in FL by updating the parameters according to a majority vote on the sign of the momentum terms. This majority voting also allows fault tolerance. The aggregation rule of SignSGD is given by:
\begin{equation} \label{eq: signSGD}
  \mathrm{SignSGD}(\mathbf{m}_{1}, \dotso, \mathbf{m}_{k})  = \mathrm{sign}\left( \sum_{i}^{k} \mathrm{sign}(\mathbf{m}_{i}) \right).
\end{equation}

\textbf{Coordinate-wise Median (CM):} CM employs the index-wise median of momentum vectors \cite{Trimmed_mean} such that we have $\hat{\mathbf{m}} = \mathrm{CM}(\mathbf{m}_{1}, \dotso, \mathbf{m}_{k})$ with 
\begin{equation} \label{eq:median}
    \hat{\mathbf{m}}[j] = \mathrm{median}( \mathbf{m}_{1}[j],\ldots,\mathbf{m}_{k}[j]).
\end{equation}

\textbf{Trimmed Mean (TM):} Rather than considering the median value as a representative as in CM, TM \cite{Trimmed_mean} discards outlier values and takes the mean of the remaining values, to achieve a more reliable consensus. Formally, under the assumption that the number of Byzantines, $k_{m}$, is known, for each coordinate $j$, TM computes the average after discarding the $k_{m}$ largest and smallest values, i.e.,
\begin{equation} 
[\mathrm{TM}(\mathbf{m}_{1}, \dotso, \mathbf{m}_{n})]_{j} = \frac{1}{k- 2k_{m}}\sum\limits_{i=k_{m+1}}^{k - k_{m-1}}[\mathbf{m_{\mathcal{S}_{j}(i)}}]_{j}\label{tm},
\end{equation}
where $\mathcal{S}_{j}$ denotes the ordered set formed by permuting the elements of the set of client indices $[n]$. Here, the order is determined by sorting the values at the $j$th coordinate.

\textbf{Bulyan:} Bulyan \cite{Bulyan} is designed as a two staged defence mechanism, where at the first stage $k-2k_{m}$ benign candidates are selected according to an elimination-based aggregation rule, e.g., M-Krum. Then, at the second stage, these candidates are aggregated according to TM.

\textbf{GAS:} GAS \cite{GAS} proposes enhancing defence robustness by partitioning the update vector into sub-vectors, with the number of partitions determined by the NN architecture size and typically increased for greater robustness.

\subsection{Analysing Existing Attack Methods}

\textbf{ALIE:} \label{sec:alie} 
Traditional aggregators such as M-Krum \cite{Krum}, TM \cite{Trimmed_mean}, and Bulyan \cite{Bulyan} assume that the selected set of momentums lie within a ball centered at the real mean, whose radius is a function of the number of benign clients. The ALIE attack \cite{ALIE} utilizes the index-wise mean $\bar{\mathbf{m}}$ and standard deviation $\bar{\boldsymbol{\sigma}}$ vectors of the benign clients to induce small but consistent perturbations to the parameters. By keeping the momentum values close to $\bar{\mathbf{m}}$, ALIE can steadily achieve an accumulation of error while concealing itself as a benign client during training. To avoid detection and staying close to the center of the ball, ALIE scales $\bar{\boldsymbol{\sigma}}$ with a parameter $z^{max}$ that depends on the number of benign and Byzantine clients. For this, let $s$ be the minimal number of benign clients that are required as supporters with $s = \left[ \frac{k}{2} + 1 \right] - k_{m}$. The attacker then leverages the cumulative standard normal function $\phi(z)$, and computes $z^{max}$ such that $s$ benign clients have a greater distance to the mean compared to the Byzantine clients. As a result, the benign clients are more likely to be classified as Byzantines. $z^{max}$ is given by: 
\begin{equation} \label{eq:z_max}
z^{max} = \mathrm{max} \left \{ z: \phi(z) < \frac{k-k_{m}-s}{k-k_{m}}  \right \}.
\end{equation}
Ultimately, $z^{max}$ is employed as a scaling parameter for $\bar{\boldsymbol{\sigma}}$ to perturb the mean of the benign clients:
\begin{equation}\label{eq:alie}
    \mathbf{m}_{i} = \bar{\mathbf{m}} - z^{max} \bar{\boldsymbol{\sigma}} \: , \forall i\in\mathcal{K}_{m}.
\end{equation}
Each individual Byzantine client generates an attack with a momentum value near  $\bar{\mathbf{m}}$, following (\ref{eq:alie}).

\textbf{Inner Product Manipulation (IPM):}  A necessary condition for convergence in FL is the positive alignment between the benign gradient vector $\bar{\mathbf{g}}$ and the output of the robust aggregation estimator \cite{IPM}. That is, the inner product between these two vectors must be positive:
\begin{equation}\label{ipm_orig}
 \langle \bar{\mathbf{g}}, AGG(\mathbf{g}_{i}: i \ \epsilon \ \mathcal{K} ) \rangle \geq 0,
\end{equation}
which ensures that the loss is steadily minimized over iterations. The IPM attack \cite{IPM} generates poisoned model updates at each iteration with the objective of violating condition (\ref{ipm_orig}). IPM sets the benign gradient with the inverse sign, $-\bar{\mathbf{g}}$, as its base for the attack and chooses a proper scaling parameter $z$ for imperceptibility so that the final version of the attack $-z\bar{\mathbf{g}}$ is not easily spotted but hinders the convergence.\footnote{When the aggregation is performed over momentum terms, the attack is performed as  $-z\bar{\mathbf{m}}$.} One of the major drawbacks of IPM is that it requires a large malicious client ratio $\frac{k_{m}}{k}$ to violate the condition in (\ref{ipm_orig}) \cite{rop}.

\textbf{Adaptive Optimized Attacks:} ALIE and IPM attacks utilize a fixed pre-determined scaling parameter throughout the training process. In \cite{ndss2022_Attack}, the authors employ the ALIE and IPM frameworks to derive the base form of the perturbation. Instead of using a fixed scaling parameter $z$, they adaptively adjust it by solving an optimization problem that maximizes $z$ under geometric distance–based imperceptibility constraints at each iteration.

\subsection{Trade-off Between Imperceptibility and Strength }\label{sec:revisiting_alie}

In this section, we inspect the trade-off between imperceptibility and attack strength considering the ALIE attack and well known robust aggregators to motivate the design of our proposed \textbf{hybrid sparse attack (HSA)}.

While end-to-end test accuracy provides the effectiveness of a defence, it offers limited insight into the internal dynamics of the aggregation process. To be more precise, the attack might be ineffective due to being easily spotted as an outlier, or can be imperceptive but too weak to meaningfully influence the convergence behaviour. Hence, to move beyond the standard black-box evaluation based on test accuracy and gain a more granular understanding of aggregator performance, we introduce an additional performance metric, {\em escape ratio}, which quantifies the proportion of malicious updates that successfully bypass the elimination stage of a given defence mechanism.

Throughout this section, \textbf{Acc.} denotes the test accuracy of the resulting global model after training under the corresponding attack/defence. When the escape ratio is below $100\%$, the defence filters part of the malicious signal (client updates and/or coordinate values, depending on the defence), and the reported accuracy is computed from the final model obtained from the resulting (filtered) aggregation trajectory.

We revisit the ALIE attack, and analyse the escape ratio under four different robust aggregators (CM, TM, Multi-Krum, and Bulyan) for varying $z$ values. Before the experimental setup and performance analysis, we briefly formalize the escape ratio metric for different robust aggregators. In TM, a Byzantine coordinate value that is not trimmed is considered an escape. Then, escape ratio is given as $e_{TM} = \frac{\sum_{j=1}^{d} n_{j}^m}{k_m \cdot d}$. We use a slightly modified formulation for CM, since for each index only the median value is selected, i.e.,  $e_{CM} = \frac{\sum_{j=1}^{d} n_{j}^m}{ d}$, where $n_{j}^m \in \left\{0,1\right\}$. In the case of Multi-Krum, the aggregator selects $n - m -2$ clients,  and  let $k_{m}^{krum}$  and $k_{b}^{krum}$ be the number of malicious and benign clients chosen by M-Krum respectively, then we define the escape ratio as $e_{krum} = \frac{k_{m}^{krum}}{k_m}$. Since Bulyan employs M-Krum and TM sequentially, we illustrate the escape ratio for each stage separately.

In our experiments, we consider an FL setup with ResNet-20 architecture on CIFAR-10 dataset \cite{CC,rop}, distributed over $k=25$ clients, with IID datasets, where $k_{m}=5$ ($k_m/k=20\%$) of them are malicious. The baseline accuracy without any malicious clients is $88.5\%$. In the first experiment, we consider two robust aggregation mechanisms, TM and CM against the ALIE with different scaling coefficient values $z\in \left\{0.25,0.5,1,1.5,2\right\}$. The corresponding results, test accuracy and escape ratio, are presented in Table \ref{tab:cm-tm}.
\begin{table}[]
\centering
\caption{
Escape ratios for ALIE with varying scaling coefficient values under CM and TM aggregators.
}
\label{tab:cm-tm}
\begin{tabular}{@{}ccccc@{}}
\hline
\multicolumn{1}{c}{\multirow{2}{*}{z}} & \multicolumn{2}{c}{\textbf{CM}} & \multicolumn{2}{c}{\textbf{TM}} \\ \cline{2-5} 
\multicolumn{1}{c}{} & \textbf{Esc. (\%)} & \textbf{Acc. (\%)} & \textbf{Esc. (\%)} & \textbf{Acc. (\%)} \\ \hline
0.25 & 66 & 45.43 & 99.9 & 72.7 \\
0.5 & 14 & 39.4 & 98 & 36.35 \\
1 & 0 & 48.3 & 63 & 47.9 \\
1.5 & 0 & 47.5 & 25 & 45.1 \\
2 & 0 & 49.3 & 7.5 & 42.9 \\ \hline
\end{tabular}
\end{table}
In Table \ref{tab:cm-tm}, ALIE with $z=0.5$ performs better than $z=0.25$, which implies that a \textbf{higher escape ratio does not imply a lower test accuracy}. In other words, attack performance depends on both imperceptibility and perturbation strength, which can be optimized by playing with scaling coefficient $z$. 

In the second experiment, we analyse two robust aggregators employing a Euclidean-distance outlier detection framework, M-Krum and Bulyan, as shown in Table~\ref{tab:Krum-Bulyan}. In the case of M-Krum, we observe that even with larger scaling, e.g., $z=0.5$ or $1$, ALIE is still imperceptible. However, at a certain level, e.g., $z=1.5$, ALIE becomes detectable and the attack is completely neutralized. For the Bulyan aggregator, which combines M-Krum and TM, we observe an identical behaviour on the client-level escape ratio (a sudden drop at $z=1.5$), but we also observe a gradual drop in the index-wise escape ratio. Here, we also highlight that the lowest accuracy is observed when $z=1$ and the index-wise escape ratio is $29\%$.

\begin{table}[]
\centering
\caption{
Client and index escape ratios for ALIE with varying scaling coefficient values under M-Krum and Bulyan.}
\label{tab:Krum-Bulyan}
\resizebox{\columnwidth}{!}{%
\begin{tabular}{cccccc}
\hline
\multirow{2}{*}{\textbf{z}} & \multicolumn{2}{c}{\textbf{M-Krum}} & \multicolumn{3}{c}{\textbf{Bulyan}} \\ \cline{2-6} 
 & \textbf{Client\%} & \textbf{Acc. \%} & \textbf{Client \%} & \textbf{Indices \%} & \textbf{Acc. \%} \\ \hline
0.25 & 100 & 55.75 & 100 & 73 & 47.13 \\
0.5 & 100 & 45.15 & 100 & 60 & 47.43 \\
1 & 100 & 45.32 & 100 & 29 & 32.79 \\
1.5 & 0 & 87.4 & 0 & 0 & 86.54 \\ \hline
\end{tabular}%
}
\end{table}

Based on these numerical experiments, we identify three key observations:
\begin{itemize}
\item ALIE is indeed imperceptible against robust aggregation methods CM, TM, M-Krum, Bulyan. However, the adversarial perturbation may not be sufficient to hinder the convergence of the model.
\item The use of a stronger perturbation, with a larger scaling coefficient $z$, makes the ALIE attack more visible. 
\item Finally, after a certain level, stronger perturbations can be fully detected by Euclidean distance-based investigations which leads to neutralization of the attack.
\end{itemize}
These observations explain the performance gain behind the dynamic/optimized attacks in \cite{ndss2022_Attack}. 

Next, we expand our analysis to two other important robust aggregators, CC and RFA. 
We note that, RFA assigns a weight to each client, based on Euclidean distance, to compute the geometric median. Thus, in Table \ref{tab:RFA-ablation}, we illustrate the weights assigned to Byzantine model updates, which indicate their imperceptibility. The decrease in the weights with respect to $z$ implies that RFA is sensitive to adversarial perturbation. However, since the adversarial updates still contribute to global model, we observe similar performance for different $z$ values, as the perturbation strength and the aggregation weight balance each other. 

\begin{table}[t]
\centering
\caption{Aggregation weights of Byzantine clients under RFA, for varying scaling coefficients of ALIE. 
The weight assigned to the benign clients is $0.2$.}
\label{tab:RFA-ablation}
\centering
\begin{tabular}{ccc}
\hline
z & \textbf{Weight} & \textbf{Accuracy (\%)} \\ \hline
0.25 & 0.16 & 48 \\
0.5 & 0.063 & 50 \\
1 & 0.026 & 54 \\
1.5 & 0.016 & 52 \\ \hline
\end{tabular}
\end{table}

Finally, we analyse the CC framework, which sanitizes models through vector-wise clipping. CC is shown to be effective against the ALIE attack \cite{CC}. However, certain limitations exist \cite{rop}. In particular, CC only scales the deviation from the reference model, i.e., the aggregate momentum from the previous round, and is therefore invariant to the direction of the perturbation. When a Byzantine client follows the ALIE strategy, it sends $\bar{\mathbf{m}} - z \bar{\boldsymbol{\sigma}}$ as a poisoned model update. The CC mechanism first measures the drift with respect to the reference update $\tilde{\mathbf{m}}_{t}$:
\begin{equation}
\mathbf{\Delta}^{ref}_{t}= \tilde{\mathbf{m}}_{t} - \bar{\mathbf{m}}_{t} +  z \bar{\boldsymbol{\sigma}}_{t},
\end{equation}
and then clips the drift $\mathbf{\Delta}^{ref}_{t}$ if its norm is larger than $\tau$, i.e., 
\begin{equation}
\tilde{\mathbf{\Delta}}^{ref}_{t}=
\min\left\{1,\frac{\tau}{\vert\vert \mathbf{\Delta}^{ref}_{t} \vert\vert}\right\}\mathbf{\Delta}^{ref}_{t},
\end{equation}
which we refer to as the \textit{effective perturbation}. Here, the clipping depends only on the norm of 
$\mathbf{\Delta}^{ref}_{t}$. Hence, when ALIE employs a larger scaling parameter $z$, CC can detect and clip the attack. However, this clipping does not necessarily alter the direction of the adversarial perturbation or prevent its temporal correlation, which, as illustrated in \cite{rop}, reinforces the time-coupled nature of ALIE and plays a key role in driving model divergence over many iterations. 
\begin{table}[]
\centering
\caption{ALIE attack with different scaling values $z$ against the CC aggregator. 
}
\label{tab:alieVsCC}
\begin{tabular}{@{}ccccc@{}}
\toprule
z & \textbf{Angle} & \textbf{Cos.} & \textbf{Norm} & \textbf{Acc. (\%)} \\ \midrule
0.25 & 92 & 0.82 & 0.4 & 82 \\
0.5 & 88 & 0.93 & 0.7 & 66 \\
1 & 77 & 0.97 & 1.2 & 50 \\
1.5 & 76 & 0.98 & 1.7 & 50 \\ \bottomrule
\end{tabular}
\end{table}

To clarify the discussion above, we conduct an experiment where we employ ALIE with scaling parameters $z\in \left\{0.25,0.5,1,1.5\right\}$ and use CC with $\tau=1$ as a robust aggregator. Then, for each scaling value, we measure the average norm of the drift $\vert \vert \mathbf{\Delta}^{ref}_{t}\vert\vert$, the expected angle between the effective perturbation $\tilde{\mathbf{\Delta}}^{ref}_{t}$ and the reference update $\tilde{\mathbf{m}}_{t}$, denoted by $\angle (\tilde{\mathbf{\Delta}}^{ref}_{t},\tilde{\mathbf{m}}_{t})$, and finally the temporal correlation of the effective perturbation over iterations, measured by the cosine similarity between the consecutive effective perturbations, $\mathrm{cos}(\tilde{\mathbf{\Delta}}^{ref}_{t}, \tilde{\mathbf{\Delta}}^{ref}_{t-1})$. The results are illustrated in Table \ref{tab:alieVsCC}, which clearly demonstrate that being clipped by the CC does not imply being sanitized properly. As the scaling parameter $z$  increases, CC starts clipping the perturbation $\mathbf{\Delta}^{ref}_{t}$, as the norm goes beyond $\tau=1$. However, even under clipping, the perturbation’s orthogonality remains preserved, which plays a crucial role in destabilizing the model, as shown in \cite{rop}. The results in 
The key observation here is that, as the scaling parameter $z$ increases, the time-coupled nature of ALIE, measured through cosine similarity, becomes more and more visible. Thus, we conclude that the CC strategy gives a false sense of security against ALIE, and the performance reported in \cite{CC} against ALIE is limited to $z$ values in certain range.

In conclusion, these numerical experiments suggest that \textbf{strict imperceptibility is not always necessary for a poisoning attack to have an impact}. In particular, within our evaluated settings, trading some imperceptibility for a stronger perturbation can increase the overall impact of the attack. This observation motivates our proposed HSA, which increases attack strength by allocating a larger perturbation to a small subset of locations, thereby seeking a favourable trade-off between perturbation strength and imperceptibility.

\section{Proposed Hybrid Sparse Attack (HSA)}\label{sec:design}
In this section, we first present the fundamental structure of our proposed HSA. We then highlight the key design principles that distinguish HSA from its counterparts, particularly ALIE, by balancing imperceptibility and attack strength. Finally, we draw an analogy between HSA and network pruning, discussing how side information about network topology can further enhance the effectiveness of HSA.

\subsection{The Fundamental Structure of HSA}

An effective Byzantine attack should be \textit{versatile} and \textit{balanced}. Versatility ensures that the Byzantine attack is capable of evading both index-wise and Euclidean distance-based defence mechanisms. A balanced attack should seek a trade-off between imperceptibility and strength to achieve the best performance against robust aggregators.

Our core idea is combining two orthogonal Byzantine strategies to form a versatile and balanced attack. In particular, we propose to generate two adversarial perturbations, each designed to target particular vulnerabilities, which are then combined to form the final adversarial perturbation such that:
\begin{equation}
\boldsymbol{\Delta}_{t} = \boldsymbol{\Delta}_{1,t}  + \boldsymbol{\Delta}_{2,t}.
\end{equation}
The first component, in line with ALIE, is equal to $\boldsymbol{\Delta}_{1,t} = - z_{1} \bar{\boldsymbol{\sigma}}$, which is small (imperceptible) but accumulates over time. The second component boosts attack performance by trading off strength against imperceptibility. For the second component, naively selecting a larger scaling coefficient $z_{2}$ would strengthen the attack but also make Byzantine updates detectable by index-wise or Euclidean distance outlier tests.

Instead, we propose using a larger scaling coefficient $z_{2}>z_{1}$ at only a small subset of locations (coordinates). Doing so raises the attack strength while keeping the Euclidean distance to benign model updates small enough to avoid detection. Formally, we leverage a sparse binary mask $\mathbf{c}\in\{0,1\}^{d}$ to generate the perturbation:
\begin{equation}
\mathbf{\Delta}_{t} = (z_{1}(1-\mathbf{c}) + z_{2}\mathbf{c})\odot \boldsymbol{\sigma}_{t}.
\end{equation}
The sparsity of $\mathbf{c}$ concentrates the larger scaling $z_2$ on a few locations, increasing per-coordinate impact without producing a large global deviation.

Following ALIE, we set $z_{1} = z^{\max}$ as in (\ref{eq:z_max}). On the other hand, for $z_{2}$ we consider an alternative strategy. From Chebyshev's inequality, we have:
\begin{equation}\label{eq:cheby}
Pr(\vert\bar{\mathbf{m}}[i]-\mathbf{m}[i]\vert > z_{2}\mathbf{\sigma}[i] ) \leq \frac{1}{z^{2}_{2}}.
\end{equation}
Hence, setting $z_{2} \approx \sqrt{2}$ would balance invisibility and strength of the attack on the sparse component. Thus, by setting $z_{2}=1.5$, we can strengthen the attack while the sparsity of the second component helps mask its global effect. We evaluate the impact of this scaling choice on imperceptibility and attack performance through extensive simulations in the next section. Before we describe how we select the sparse binary mask $\mathbf{c}$, we describe the dynamic version of the proposed HSA framework.

\textbf{Static to Dynamic HSA:} So far, we have discussed a static version of HSA, where we use a fixed $z_1$ and $z_2$. However, dynamic attacks can be formed by optimising the scaling coefficients at each iteration \cite{ndss2022_Attack}. In this section, we introduce a dynamic HSA (DHSA): $z_2$ remains fixed, but $z_{1}^{(t)}$ is reset at each iteration $t$ based on the benign updates $\mathcal{M}_{t}=\left\{\mathbf{m}_{i,t}\right\}_{i\in\mathcal{K}_{b}}$. For this we formulate the following optimisation problem:
\begin{equation}
\label{eq:DHSA-opt}
\begin{aligned}
\max_{\,z_{1}} \quad & z_{1} \\[4pt]
\text{s.t.} \quad 
& \frac{1}{k_{b}}\sum_{i\in \mathcal{K}_{b}}
  \Big\|\,\bar{\mathbf{m}}_{t}
  - z_{1}(1-\mathbf{c})\odot\boldsymbol{\sigma}_{t}
  - z_{2}\mathbf{c}\odot\boldsymbol{\sigma}_{t} \\[-2pt]
&\qquad\qquad\qquad\qquad\qquad\qquad\qquad
  - \mathbf{m}_{i,t}\,\Big\|_{2}
  \leq \Delta_{\mathrm{th}}, \\[4pt]
\end{aligned}
\end{equation}
with:
\begin{equation}
\Delta_{th} = \frac{1}{k_{b}-1}\left(\max_{i\in \mathcal{K}_{b}} \sum_{j\in\mathcal{K}_{b}}\vert\vert \mathbf{m}_{i,t} - \mathbf{m}_{j,t} \vert\vert\right).
\end{equation}
Here when $\mathbf{c} = \boldsymbol{1}_{d}$, our attack reduces to the method in \cite{ndss2022_Attack}, making their design a special case of our framework. That is, DHSA adds one more dimension to the optimisation problem through the sparsity mask $\mathbf{c}$. We also note that, as explained later, our method leverages the specific NN architecture being targeted in the attack design, which is reflected in the choice of $\mathbf{c}$, while the attack proposed in \cite{ndss2022_Attack} is architecture-agnostic.
The general form of HSA procedure is summarized in Algorithm \ref{code:HybridSparse}. 

\begin{algorithm}[t]
    \small
    \caption{Hybrid Sparse Attack (HSA)}
	\begin{algorithmic}[1]
    \Statex \textbf{Input:} initial model weights $\boldsymbol{\theta}_{0}\in \mathbb{R}^{d}$, binary sparse mask $\mathbf{c}\in \left\{0,1\right\}^d$
     \For{$t=1,\ldots,T$}
	  \State Compute/Estimate benign statistics $(\boldsymbol{\sigma}_{t}, \bar{\boldsymbol{m}}_{t})$
   \Statex \quad \textbf{Option 1:} Static policy
   \State Use fixed scaling $z^{(t)}_{1} = z^{max}_{1}$ and $z^{(t)}_{2} = z^{max}_{2}$
   \Statex \quad \textbf{Option 2:} Dynamic policy
   \State Set $z^{(t)}_{2} = z^{max}_{2}$, optimize $z^{(t)}_{1}$ via Min-Sum Policy \cite{ndss2022_Attack}
   \Statex \quad \textbf{Generate additive perturbation:} 
   \State $\mathbf{\Delta}_{t} = (z^{(t)}_{1}(1-\mathbf{c}) + z^{(t)}_{2}\mathbf{c})\odot \boldsymbol{\sigma}_{t}$
   \State{Byzantine update :} $\mathbf{m}_{adv} = \bar{\mathbf{m}}_{t} - \mathbf{\Delta}_{t}$
    \EndFor
     \end{algorithmic}
	\label{code:HybridSparse}
\end{algorithm}

\subsection{Network Topology-Informed Byzantines}

Existing attacks in the literature are oblivious to the network topology. By incorporating side information regarding the network topology, which hints the most sensitive weights, it is possible to design a stronger attack. Hence, with this side information, perturbation budget can be allocated among the NN weights in a more effective way, which is the core idea of our sparse binary mask design for the proposed HSA.

The key question is how to find a subset of weights such that attacking these weights simultaneously has the highest impact on the accuracy. A similar yet distinct problem has been studied in the literature as \emph{network pruning}, where the objective is to match the accuracy of a dense network with only a small subset of the NN weights.

Previous studies on network pruning have shown that, for moderate sparsity levels $\delta$, the vanilla approach of using random sub-networks can often match the performance of dense networks \cite{prune_connectivity2}. That is, for a given sparsity ratio $\delta$, we can  generate a binary value for each index i.e., for each $i\in[d]$:
\begin{equation}\label{eq:altsign}
\mathbf{c}[i] =  
  \left\{ \begin{array}{cl}
1, & \text{if } \ w.p. ~\delta\\
0, & \text{if } \ w.p. ~ 1-\delta
\end{array} \right.
\end{equation}
In our experiments, we observe that randomly generated sparse masks can serve as an effective candidate for our HSA as seen in the Appendix \ref{sec:network_pruning}. Our results indicate that at lower sparsity regimes, even randomly generated masks can perform similarly or better that more advanced approaches. Indeed, such masks perform well against a range of defence mechanisms. However, when we move our focus from moderate sparsity to high sparsity regime, to make the attack more imperceptible, randomly generated masks exhibit poor performance. 

\textbf{Critical Layers:} Another important consideration in the literature on sparse networks is the presence of sensitive layers, which are typically excluded from pruning because they are highly sensitive \cite{prune_force,prune_nodata,powerSGD}. These critical layers often include batch normalization layers and bias parameters. We denote the set of critical layers as $\mathcal{L}_{critical}$. From the perspective of a Byzantine attack, we argue that, analogous to keeping these layers dense in network pruning, the attack should particularly target these sensitive layers. Accordingly, we exclude all critical layers $l \in \mathcal{L}_{critical}$ from pruning, i.e., all corresponding entries in $\mathbf{c}$ are set to $1$. Further ablations on critical layers are included in Appendix \ref{sec:ablation_on_sparsity}. Overall we find that, including critical layers, strength of the HSA also increases against most of the aggregators especially on high sparsity regimes (i.e.,$\delta=5\times10^-3$ ). 

\textbf{Synaptic Connectivity:}
Existing works on NN pruning \cite{prune_force,prune_nodata,powerSGD,prune3,prune_connectivity,prune_connectivity2}, have shown that, in high-sparsity regimes, randomly generated sub-networks often underperform; hence, it is crucial to take into account synaptic connectivity in the underlying NN architecture, when designing the sparsity mask.

With these in mind, our objective is to find a small subset of NN parameters $\mathcal{M}$ with size 
$||\mathcal{M}|| = \delta \cdot d << d $ such that the NN model with only the subset of the parameters $\boldsymbol{\theta}_{\mathcal{M}}$, i.e., $\boldsymbol{\theta}[i]=0,i\in\mathcal{M}$ is sufficient to achieve a certain loss over the given dataset $\mathcal{D}$. Formally speaking, the objective can be formulated as:
\begin{equation} \label{prune}
\begin{aligned}
\underset{\mathbf{c}, \boldsymbol{\theta}}{\min} \quad 
& F(\mathbf{c} \odot \boldsymbol{\theta}; \mathcal{D}) 
= \frac{1}{|\mathcal{D}|} \sum_{\zeta \in \mathcal{D}} f(\mathbf{c} \odot \boldsymbol{\theta}; \zeta) \\
\text{s.t.} \quad 
& \|\mathbf{c}\|_{0} \le d \cdot \delta, \\
& \boldsymbol{\theta} \in \mathbb{R}^{d}.
\end{aligned}
\end{equation}
As discussed, even though a random sparse masks strategy is a feasible solution for the optimization problem in (\ref{prune}), it often underperforms in high sparsity regimes. Hence, for our HSA, we leverage a more advanced network pruning strategy named FORCE \cite{prune_force}. FORCE takes the synaptic connectivity of the network into account. In a nutshell, the FORCE algorithm iteratively eliminates network weights based on a predefined saliency measure, which is explained further in Appendix \ref{sec:network_pruning}. Another important reason for using the FORCE algorithm to design sparse masks is that it requires only a small subset of the data, which can be accessed by the colluding Byzantines. Thus, in our main analysis, we employ the FORCE algorithm to generate the sparse masks. However, in the Appendix \ref{sec:sparsity_structure}, we provide an extensive numerical analysis of the impact of sparse masks on the performance of HSA, where we show that most sparse masks are robust against robust aggregators.  Further, recent works have proposed more advanced pruning methodologies \cite{prune_at_init2,prune_at_init3,prune_layerwise}, which enable pruning at initialization with little to no data while achieving better performance. Nonetheless, within the scope of this paper, we do not consider these approaches and leave the exploration of optimal pruning strategies for attack design as an open research direction.

\begin{table*}[]
\centering
\normalsize
\caption{Performance comparison of eight attack methods 
against eight robust aggregators.  
We show the final test accuracy results for the ResNet-20 architecture with the CIFAR-10 dataset for {\em IID} split. 
}
\label{tab:Main-IID}
\resizebox{\textwidth}{!}{%
\begin{tabular}{cccccccl}
\hline
\multicolumn{1}{l}{\textbf{Attack - Defence}} & \textbf{Bulyan} \cite{Bulyan} & \textbf{CC} \cite{CC} & \textbf{CM} \cite{Trimmed_mean} & \textbf{M-Krum} \cite{Krum} & \textbf{RFA} \cite{RFA} & \textbf{TM} \cite{Trimmed_mean} & \textbf{GAS (Krum - Bulyan)} \cite{GAS}  \\ \hline
\textbf{No ATK} & 87.6 $\pm$ 1 & 87.4 $\pm$ 0.5 & 88 $\pm$ 0.1 & 89.2 $\pm$ 0.7 & 88.2 $\pm$ 0.3 & 87.8 $\pm$ 0.4 & 87.7 $\pm$ 0.3 $-$ 87.6 $\pm$ 0.4 \\
ALIE \cite{ALIE} & 47.1 $\pm$ 16 & 80.7 $\pm$ 1.4 & 45.1 $\pm$ 9.3 & 55.8 $\pm$ 14 & 48.7 $\pm$ 11 & 72.8 $\pm$ 2.5 & 81 $\pm$ 0.9 $-$ 81.1 $\pm$ 0.6 \\
Bit Flip  & 86.2 $\pm$ 0.6 & 82.2 $\pm$ 0.7 & 80.9 $\pm$ 0.8 & 87.8 $\pm$ 0.3 & 87.5 $\pm$ 0.7 & 80.1 $\pm$ 0.9 & 86.5 $\pm$ 0.5 $-$ 86.8 $\pm$ 0.8 \\
IPM \cite{IPM} & 70 $\pm$ 0.9 & 85.8 $\pm$ 0.3 & 54.5 $\pm$ 0.8 & 83.8 $\pm$ 0.1 & 66.4 $\pm$ 0.7 & 84.9 $\pm$ 0.1 & 85.3 $\pm$ 0.2 $-$ 85.6 $\pm$ 0.3 \\
Label Flip \cite{Label_flip1} & 85.8 $\pm$ 0.3 & 86.4 $\pm$ 0.6 & 79.9 $\pm$ 0.1 & 87.7 $\pm$ 0.1 & 86.3 $\pm$ 0.1 & 78.5 $\pm$ 0.3 & 87.1 $\pm$ 0.7 $-$ 86.9 $\pm$ 0.6 \\
ROP \cite{rop} & 41.5 $\pm$ 2 & 47.4 $\pm$ 0.9 & 79 $\pm$ 0.4 & 88.2 $\pm$ 0.5 & \textcolor{red}{\underline{43.2 $\pm$ 2.2}}& 62.7 $\pm$ 1.3& \textcolor{red}{\underline{62 $\pm$ 1.6}} $-$ \textcolor{red}{\underline{63.5 $\pm$ 0.7}} \\
HSA - $\Pi_{F} (\delta$) & \textcolor{red}{\underline{40.6 $\pm$ 1.7}} & \textcolor{red}{\underline{39.6 $\pm$ 3.6}} & \textcolor{red}{\underline{39.5 $\pm$ 9.2}} & \textcolor{red}{\textbf{15.5 $\pm$ 1.2}} & 52.8 $\pm$ 11 & \textcolor{red}{\underline{53.5 $\pm$ 17}} & 74.5 $\pm$ 4 $-$ 74.8 $\pm$ 3.3 \\
HSA - $\Pi_{F} (\delta,\delta^{max}_{FC} )$ & \textcolor{red}{\textbf{31.3 $\pm$ 2.7}} & \textcolor{red}{\textbf{33.9 $\pm$ 2.9}} & \textcolor{red}{\textbf{35.7 $\pm$ 4.5}} & \textcolor{red}{\underline{16.3 $\pm$ 2.1}} & \textcolor{red}{\textbf{42.7 $\pm$ 10}} & 
\textcolor{red}{\textbf{39.5 $\pm$ 8.41}}& \textcolor{red}{\textbf{59.5 $\pm$ 18}} $-$ \textcolor{red}{\textbf{52.5 $\pm$ 8.3}} \\ \hline
Min-Sum \cite{ndss2022_Attack} & \textcolor{blue}{\underline{41.1 $\pm$ 0.5}} & 55.5 $\pm$ 5.2 & \textcolor{blue}{\textbf{47.6 $\pm$ 0.4}} & \textcolor{blue}{\underline{55.1 $\pm$ 1.3}} & 49.4 $\pm$ 8.6 & 47.5 $\pm$ 7.4 & 55.1 $\pm$ 3 $-$ \textcolor{blue}{\textbf{49.9 $\pm$ 11.8}} \\
Min-Max \cite{ndss2022_Attack} & 67.2 $\pm$ 4.4 & \textcolor{blue}{\underline{40.3 $\pm$ 7.8}} & 48 $\pm$ 1.8 & 61.6 $\pm$ 12 & \textcolor{blue}{\underline{48.7 $\pm$ 2.8}} & \textcolor{blue}{\underline{42.9 $\pm$ 6.2}} & \textcolor{blue}{\textbf{47.1 $\pm$ 4}} $-$ 86.6 $\pm$ 0.2 \\
DHSA-$\Pi_{F} (\delta,\delta^{max}_{FC} )$ & \textcolor{blue}{\textbf{37.9 $\pm$ 5.6}} & \textcolor{blue}{\textbf{39.3 $\pm$ 3}} & \textcolor{blue}{\underline{47.7 $\pm$ 1}} & \textcolor{blue}{\textbf{14.5 $\pm$ 1.4}} & \textcolor{blue}{\textbf{39.5 $\pm$ 0.9}} & \textcolor{blue}{\textbf{33.9 $\pm$ 5.2}} & \textcolor{blue}{\underline{47.2 $\pm$ 10}} $-$ \textcolor{blue}{\underline{54.4 $\pm$ 1.2}} \\ \hline
\end{tabular}%
}
\end{table*}

\section{Numerical Results}
\label{sec:num-results}

In this section, we present comprehensive experimental evaluations of the proposed HSA framework against state-of-the-art Byzantine-robust aggregators. We begin by detailing our simulation setup, including the datasets, network architectures, threat models, and defence mechanisms considered in our experiments. Subsequently, we analyse the attack performance under both IID and non-IID data distributions. Our results demonstrate that HSA achieves superior attack effectiveness compared to existing Byzantine attacks, reducing test accuracy to as low as $15.5\%$ against robust aggregators in IID scenarios and $9.2\%$ in non-IID settings, significantly outperforming baseline attacks such as ALIE \cite{ALIE} and ROP \cite{rop}. Furthermore, we investigate the impact of sparsity constraints and pruning strategies on attack performance through comprehensive ablation studies in Appendix \ref{sec:ablation_on_sparsity}.

\begin{table*}[]
\centering
\normalsize
\caption{Performance comparison of eight attack methods 
against eight robust aggregators.  
We show the final test accuracy results for the ResNet-20 architecture with the CIFAR-10 dataset for {\em non-IID} split. 
}
\label{tab:nonIID-main}
\resizebox{\textwidth}{!}{%
\begin{tabular}{cccccccl}
\hline
\multicolumn{1}{l}{\textbf{Attack - Defence}} & \textbf{Bulyan} \cite{Bulyan} & \textbf{CC} \cite{CC} & \textbf{CM} \cite{Trimmed_mean} & \textbf{M-Krum} \cite{Krum} & \textbf{RFA} \cite{RFA} & \textbf{TM} \cite{Trimmed_mean} & \textbf{GAS (Krum-Bulyan)} \cite{GAS}   \\ \hline
\textbf{No ATK} & 86 $\pm$ 0.5 & 86.9 $\pm$ 0.3 & 79.9 $\pm$ 0.4 & 88.1 $\pm$ 0.4 & 86.4 $\pm$ 0.7 & 86.7 $\pm$ 0.5 & 86.7 $\pm$ 0.5 $-$ 86.3 $\pm$ 0.5 \\
ALIE \cite{ALIE} & 34.7 $\pm$ 0.7 & 59.7 $\pm$ 3.4 & 34.5 $\pm$ 1.9 & 49.7 $\pm$ 9.2 & \textcolor{red}{\underline{32.9 $\pm$ 3.4}} & \textcolor{red}{\underline{42.1 $\pm$ 7.2}} & \textcolor{red}{\underline{57 $\pm$ 11.4}} $-$ 63.6 $\pm$ 5.3 \\
Bit Flip & 79.8 $\pm$ 1 & 81.8 $\pm$ 0.5 & 73.5 $\pm$ 1.1 & 85.2 $\pm$ 0.5 & 80.9 $\pm$ 0.6 & 77 $\pm$ 1.3 & 84.2 $\pm$ 0.3 $-$ 84.5 $\pm$ 0.2 \\
IPM \cite{IPM} & 46.3 $\pm$ 0.9 & 83.3 $\pm$ 0.7 & 61.9 $\pm$ 4.6 & 73.2 $\pm$ 3.9 & 66.4 $\pm$ 0.7 & 77.9 $\pm$ 1.9 & 83.3 $\pm$ 0.6 $-$ 82.3 $\pm$ 0.7 \\
Label Flip \cite{Label_flip1} & 80.9 $\pm$ 1 & 85.2 $\pm$ 0.4 & 72.4 $\pm$ 0.2 & 87.3 $\pm$ 0.3 & 84.6 $\pm$ 0 & 73 $\pm$ 1.5 & 86 $\pm$ 0.6 $-$ 85.5 $\pm$ 0.3 \\
ROP \cite{rop} & \textcolor{red}{\underline{22.5 $\pm$ 5.4}} & 28.9 $\pm$ 1.1 & 31.2 $\pm$ 2.4 & 87.3 $\pm$ 0.7 & 34.4 $\pm$ 2 & 55.6 $\pm$ 1.6 & 64.6 $\pm$ 1.7 $-$\textcolor{red}{\underline{62.5 $\pm$ 1.2}} \\ 
HSA-$\Pi_{F} (\delta$) & 26.3 $\pm$ 3.4 & \textcolor{red}{\underline{24.3 $\pm$ 3}} & \textcolor{red}{\underline{26.4 $\pm$ 2.3}} & \textcolor{red}{\underline{10.9 $\pm$ 1.3}} & 39.6 $\pm$ 5.8 & \textcolor{red}{\textbf{10 $\pm$ 0}} & \textcolor{red}{\textbf{10 $\pm$ 0}} $-$ 64.3 $\pm$ 7.7 \\
HSA-$\Pi_{F} (\delta,\delta^{max}_{FC} )$ & \textcolor{red}{\textbf{9.9 $\pm$ 0.3}} & \textcolor{red}{\textbf{17.6 $\pm$ 4}} & \textcolor{red}{\textbf{22.9 $\pm$ 2.1}} & \textcolor{red}{\textbf{9.8 $\pm$ 0.1}} & \textcolor{red}{\textbf{10 $\pm$ 0}} & \textcolor{red}{\textbf{10 $\pm$ 0}} & \textcolor{red}{\textbf{10 $\pm$ 0}} $-$ \textcolor{red}{\textbf{9.2 $\pm$ 1.1}} \\ \hline
Min-Sum \cite{ndss2022_Attack} & \textcolor{blue}{\underline{21.6 $\pm$ 4.1}} & 38.1 $\pm$ 2.6 & \textcolor{blue}{\underline{26.1 $\pm$ 2.5}} & \textcolor{blue}{\underline{27.2 $\pm$ 5.9}} & \textcolor{blue}{\underline{26.8 $\pm$ 2.2}} & 32.5 $\pm$ 1.9 & 45.8 $\pm$ 4.7 $-$ 45.1 $\pm$ 5.8 \\
Min-Max \cite{ndss2022_Attack} & 82.2 $\pm$ 0.1 & \textcolor{blue}{\underline{34 $\pm$ 1.4}} & \textcolor{blue}{\textbf{26 $\pm$ 3}} & 86.9 $\pm$ 0.5 & 30.2 $\pm$ 2.2 & \textcolor{blue}{\underline{18 $\pm$ 1.5}} & \textcolor{blue}{\underline{30.1 $\pm$ 16}} $-$ 85.9 $\pm$ 0.4 \\ 
DHSA-$\Pi_{F} (\delta,\delta^{max}_{FC} )$ & \textcolor{blue}{\textbf{14.4 $\pm$ 3.1}} & \textcolor{blue}{\textbf{16.7 $\pm$ 1.5}} & 49 $\pm$ 13 & \textcolor{blue}{\textbf{10 $\pm$ 0}} & \textcolor{blue}{\textbf{10 $\pm$ 0}} & \textcolor{blue}{\textbf{10 $\pm$ 0}} & \textcolor{blue}{\textbf{10 $\pm$ 0}} $-$ \textcolor{blue}{\textbf{10 $\pm$ 0}} \\ \hline
\end{tabular}%
}
\end{table*}
\subsection{Simulation Setup}
\label{subsec:sim_setup}

\textbf{Datasets and Networks:} We consider an image classification task on the CIFAR-10 dataset \cite{cifar10} and train a ResNet-20 NN \cite{resnet}. In the appendices, we provide additional numerical results for alternative datasets and architectures.\footnote{We select ResNet-20 to emphasize our key findings, since it offers more reliable benchmarking. Our observations indicate that in the case of smaller NNs, the Byzantines can easily prevent convergence, making it difficult to draw clear conclusions about the effectiveness of different attacks.}

We consider IID and non-IID training data distributions among the clients. In the IID scenario, we distribute the training data homogeneously among the clients so that each client has an equal number of samples for each label. In the non-IID scenario, following the common practice \cite{B2DB,dirichlet1,dirichlet2,dirichlet_cited}, the dataset is divided among the clients according to the Dirichlet distribution \cite{dirichlet}, $\mathrm{Dir}(\alpha)$, where $\alpha$ controls the skewness of the probability distribution with respect to the labels. In our experiments, to mimic the practical scenarios, we follow the setup in \cite{B2DB} and set $\alpha=1$. 

Following the common approach in \cite{CC,rop}, we train ResNet-20 for 100 epochs with a local batch size of 32 and an initial learning rate of $\eta=0.1$, which is reduced at epoch 75 by a factor of $0.1$, and set the local momentum to $\beta=0.9$.

\textbf{Setup:} In all the experiments, we assume that Byzantines can collude to share their local data with each other. They can access, or predict, benign model updates shared with the PS, while remaining oblivious to the local datasets of benign clients. Further, following the prior works \cite{CC,rop,ndss2022_Attack},
we set Byzantine ratio $k_m/k=0.2$, and the number of clients is set to $k=25$, of which $k_{m}=5$ are malicious. In the appendices, we also provide results for different ratios, as well as results for cross-device settings where the number of participating clients is much higher. The reported results are obtained by averaging 3 independent trials.

\textbf{Defences:} From the defence perspective, we consider state-of-the-art robust aggregators: Bulyan \cite{Bulyan}, CC \cite{CC}, CM \cite{Trimmed_mean}, M-Krum \cite{Krum}, RFA \cite{RFA}, and TM \cite{Trimmed_mean}. We also employ a more sophisticated robust aggregation strategy called GAS with two variations, built on Bulyan and M-Krum, respectively \cite{GAS}. 

Next, we briefly summarize the parameter setup of the robust aggregators. For CC, we employ a fixed ball radius of $\tau=1$ since a higher radius is more susceptible to Byzantine attacks \cite{decentCC,rop}, and lower $\tau$ performs suboptimally in non-IID settings \cite{rop}. We set the number of clipping iterations to $1$. For SignSGD \cite{SignSGD}, we employ a starting learning rate of $0.01$ for more stable training. For the number of sub-vectors $p$ in GAS, we set $p=1000$ on ResNet-20 and 2-Layer CNN, while on 2-layer MLP, we set $p=100$ due to the smaller network size. Our $p$ values are proportional to the best $p$ parameters employed in \cite{GAS} for their NN size.

\textbf{Attacks:} For non-omniscient attacks, we consider the bit-flip \cite{RSA-def} and label-flip \cite{Label_flip1,FL_limitations,LMPA} attacks. In the bit-flip attack, Byzantine clients flip the signs of their own gradient values, whereas, in the label-flip attack, Byzantines compute the local updates according to the labels obtained by a circular shift, i.e., given true label $c$, the target label is set to $c+1 \mod(C)$, where $C$ is the total number of classes/labels.

For the omniscient model poisoning attacks, we consider three different attacks: ALIE \cite{ALIE}, IPM \cite{IPM}, and ROP \cite{rop}. In ROP, we employ the original parameter values given in \cite{rop}. For IPM, previous works have considered different $z$ values, i.e., in \cite{CC} $z=0.1$ and in \cite{rop} $z=0.2$. In our preliminary experiments, we have considered these values, but observed that IPM performs best with $z=0.4$. Hence, in our results, we report the performance with $z=0.4$. 

Apart from the aforementioned static attacks, we also include a dynamic attack framework, where the parameters are optimized adaptively over the iterations. In particular, we consider Min-Sum and Min-Max attacks \cite{ndss2022_Attack}. 

\textbf{Hybrid Sparse Attack (HSA) Implementation:} We consider 3 different implementations of HSA. First, we utilize a sparsity mask generated according to the FORCE \cite{prune_force} algorithm without any layer-wise sparsity constraint, denoted by $\Pi_{F}(\delta)$. 

Then, we follow the same strategy, but introduce a layer-wise sparsity constraint on the fully connected (FC) layer of the network, denoted by $\Pi_{F}(\delta,\delta^{max}_{FC})$, which ensures that the sparsity ratio of the FC layer cannot exceed $\delta^{max}_{FC}$. Although it is possible to consider different layer-wise sparsity constraints, we limit our focus to the FC layer for two reasons: First, after the network pruning, FC layer has the highest layer-wise density. Second, the FC layer contains larger weights and proportionally higher variance; therefore, attacking the FC layer makes the Byzantines more easily detectable under Euclidean distance–based outlier detection.\\
\indent In Appendix~\ref{sec:ablation_on_sparsity}, we conduct further experiments on layer-wise sparsity constraint to to better visualize its impact on the performance. Overall we find that ensuring certain layer-wise sparsity substantially increase the attack robustness. 

The first two implementations of HSA are static in the sense that the scaling coefficients are fixed throughout training. We also consider dynamic HSA as described in Algorithm 3 as option 2. For DHSA, and we employ the mask formation of $\Pi_{F}(\delta,\delta^{max}_{FC})$ and we set $\delta=5\times 10^{-3}$, with $\delta^{max}_{FC}=2.5\times10^{-1}$.


\begin{figure*}
    \centering    \includegraphics[width=\textwidth]{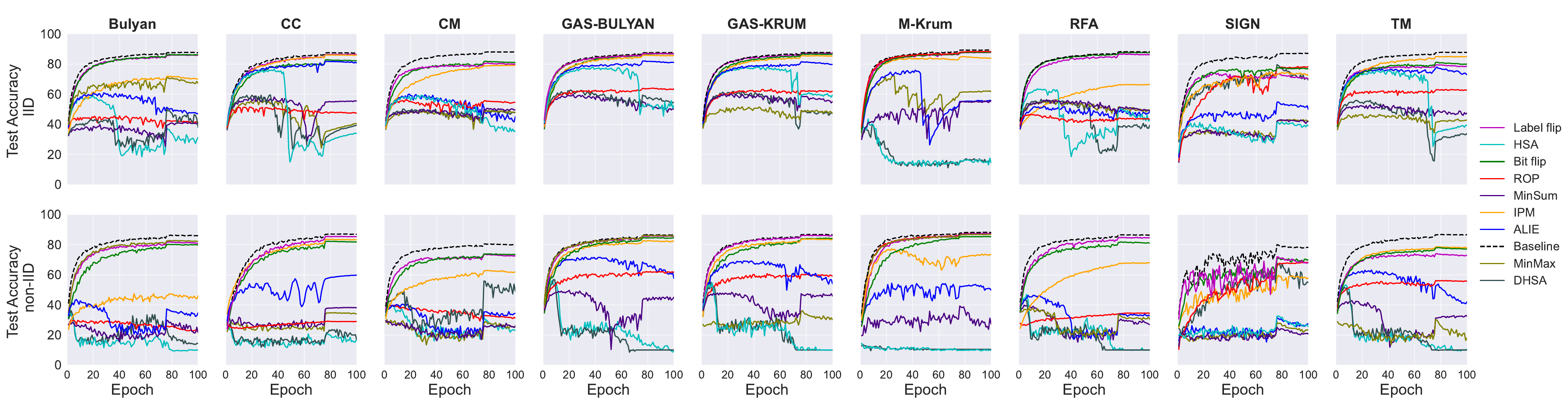}
    \caption{Test accuracy convergence results of training ResNet-20 architecture with CIFAR-10 dataset for IID and non-IID distributions. Training and testing are performed over 100 epochs with 9 robust aggregators, namely; Bulyan, CM, CC, M-Krum, RFA, SignSGD, TM, GAS-Bulyan, GAS-Krum against 8 different Byzantine attacks, namely; ALIE, Bit-Flip, Label-Flip, IPM, ROP, Min-Sum, Min-Max, HSA/DHSA).}
    \label{fig:cifar-10}
\end{figure*}

\subsection{Experimental Results} \label{sec:exp_results}
The experimental results for IID and non-IID settings are shown in Table~\ref{tab:Main-IID} and Table~\ref{tab:nonIID}, respectively. For fairness, we compare the performance of static and dynamic attacks separately, for which the best performing attacks are illustrated with {\color{red} red} and {\color{blue} blue}, respectively. We use \textbf{bold} and {\underline{underline}} to highlight the \textbf{best} (i.e., yielding the lowest test accuracy) and the \underline{runner-up} performance, respectively.

\begin{table}[]
\caption{Performance comparison of Byzantine attacks, illustrating average test accuracy over all eight robust aggregators and the test accuracy against the best-performing robust aggregator. 
}
\label{tab:attack_success}
\resizebox{\columnwidth}{!}{%
\begin{tabular}{@{}ccccc@{}}
\toprule
\textbf{} & \multicolumn{2}{c}{IID} & \multicolumn{2}{c}{non-IID} \\ \midrule
\textbf{ATTACK} & Average & Highest & Average & Highest \\
ALIE \cite{ALIE} & 59.2 & 81.1 & 48.5 & 63.6 \\
Bit Flip & 84.4 & 87.8 & 80.7 & 85.2 \\
IPM \cite{IPM} & 75.2 & 85.8 & 70.6 & 83.3 \\
Label Flip \cite{Label_flip1} & 84.5 & 87.7 & 81.3 & 87.3 \\
ROP \cite{rop} & 60.6 & 88.2 & 48.3 & 87.3 \\
HSA - $\Pi_{F} (\delta$) & 45.2 & 74.8 & 25.9 & 64.3 \\
HSA - $\Pi_{F} (\delta,\delta^{max}_{FC} )$ & \underline{42.1} & \underline{55.5} & \textbf{13.2} & \textbf{22.9} \\
Min-Sum \cite{ndss2022_Attack} & 50.8 & \underline{55.5} & 31.0 & \underline{45.8} \\
Min-Max \cite{ndss2022_Attack} & 51.4 & 86.6 & 41.6 & 86.9 \\
DHSA-$\Pi_{F} (\delta,\delta^{max}_{FC} )$ & \textbf{37.2} & \textbf{54.4} & \underline{15.7} & 49.0 \\ \bottomrule
\end{tabular}%
}
\end{table}

\subsubsection{Performance evaluation of HSA in the IID scenario:} \label{sec:perf_HSA_iid}
The results in Table~\ref{tab:Main-IID} clearly demonstrate that in the \textit{IID} scenario, the proposed HSA design is effective against all the considered robust aggregators. 
We remark that against all the listed robust aggregators, HSA consistently outperforms, or at least closely matches, the best-performing attack, showing a universal success.

Next, we further analyse the results from the lens of design principles introduced in Section~\ref{sec:revisiting_alie}. From the escape ratio analysis shown in Tables~\ref{tab:cm-tm} and \ref{tab:Krum-Bulyan}, we observe that ALIE can successfully evade robust aggregators Bulyan, M-Krum, TM, and CM. On the other hand, since ALIE utilizes a comparatively small scaling coefficient for the sake of imperceptibility, the impact of the adversarial perturbation on the final test accuracy is noticeable but limited. However, HSA trades some imperceptibility for greater strength, making the attack slightly more detectable but also more powerful, which can be easily noticed from the performance comparison of ALIE and HSA against M-Krum, with the test accuracies achieved after these attacks as $55.8\% $ and $15.5\% $, respectively. We remind that increasing the scaling coefficient $z$ alone does not strengthen ALIE. Indeed, Table~\ref{tab:Krum-Bulyan} shows that when $z=1.5$, M-Krum achieves a test accuracy of $87.4\%$. The performance comparison of HSA and ALIE against M-Krum, Bulyan, TM, and CM supports our arguments in Section \ref{sec:design} that sparse formation of HSA enables stronger attacks without compromising stealthiness. 

\textbf{On the importance of sparsity constraint:} In Table~\ref{tab:Main-IID}, one can observe that ALIE outperforms $\Pi_{F}(\delta)$ (HSA without layer-wise sparsity constraints) only for RFA. We recall that, RFA assigns a weight to each client update which is inversely proportional to its distance to the geometric median, making it more sensitive to the Euclidean distance between the geometric median and the Byzantine updates. Hence, ALIE, locating the Byzantines closer to the benign mean, performs better compared to $\Pi_{F}(\delta)$. However, by inducing a sparsity constraint on the FC layer, HSA $\Pi_{F}(\delta,\delta^{max}_{FC})$ gets closer to the benign mean and outperforms the ALIE.

\textbf{On the importance of sparsity distribution:}
Another important observation from Table \ref{tab:Main-IID} is that the distribution of sparsity plays a key role in the performance of HSA. In particular, when HSA is evaluated against GAS-Krum and GAS-Bulyan, both of which perform layer-wise aggregation, we observe that distributing non-sparse positions more uniformly by imposing a layer-wise sparsity constraint during network pruning enhances attack strength. We further conduct experiments to analyse the correlation between the sparsity mask and the attack performance, the results of which are in Appendix \ref{sec:ablation_on_sparsity}, Overall we find that with less sparsity, HSA can even get stronger if we keep the layer constraints between \%20-40.

\textbf{Dynamic attacks for better results:} One of the key shortcomings of ALIE and static HSA is the use of the fixed scaling coefficient $z$.
The results in Table \ref{tab:Main-IID} show that dynamic attacks, Min-Sum, and Min-Max can outperform their static counterparts especially against GAS-Krum and GAS-Bulyan, both of which perform layer-wise outlier investigation. Hence, inspired by {\cite{ndss2022_Attack}, we implement a dynamic version of HSA (DHSA) where the scaling coefficient $z_{1}$ is adjusted over the iterations. The performance summarization in Table \ref{tab:attack_success}, illustrates that DHSA is the best-performing attack in the IID setting by keeping the accuracy of the best-performing robust aggregator below $55\%$, and the average performance across all eight aggregators below $38\%$.

To better understand the impact of the dynamically adjusted scaling efficient,  we investigate the histogram of scaling coefficients under DHSA, Min-Sum, and Min-Max, as illustrated in Fig.~\ref{fig:z_scales}. Based on the histogram plots, we observe that dynamic attacks utilize, on average, considerably larger scaling coefficients than ALIE, which enhances their strength. However, thanks to their adaptive nature, where the $z$ parameter is adjusted according to the constraint on statistical visibility, they also remain imperceptible. Consequently, the proposed DHSA enjoys the advantage of scaling $z$ dynamically as well as identifying non-sparse positions for stronger perturbation.
\begin{figure}
    \centering
    \includegraphics[width=0.8\linewidth]{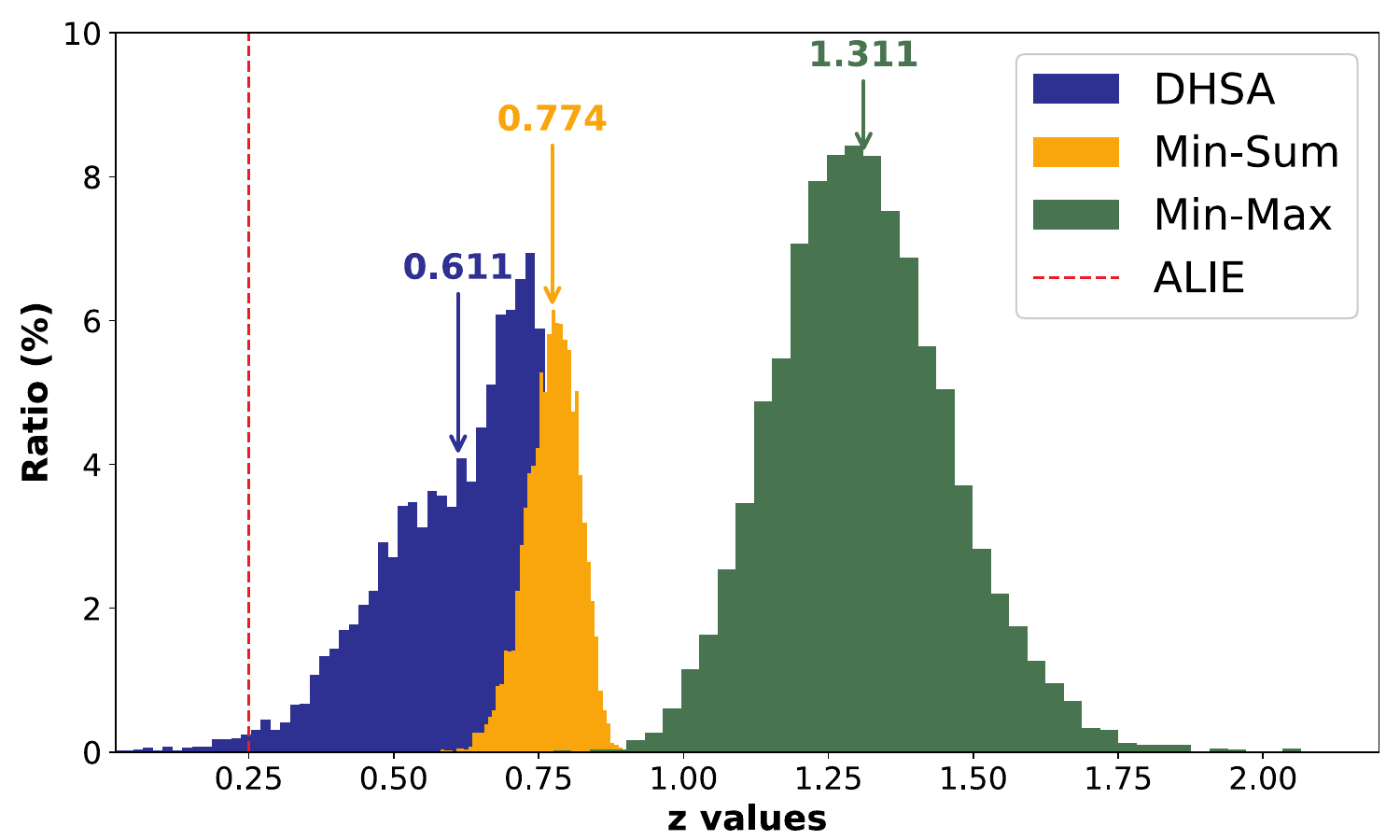}
    \caption{Histogram of $z$ values measured for DHSA, Min-Sum and Min-Max attacks, during training ResNet-20 in the \textit{IID} setting against TM aggregator.}
    \label{fig:z_scales}
    \vspace{-5mm}
\end{figure}

\subsubsection{Performance evaluation of HSA in the non-IID scenario} \label{sec:perf_HSA_noniid}
The overall performance analysis for the \textit{non-IID} scenario, shown in Table \ref{tab:nonIID-main}, demonstrates that our proposed HSA design is effective against all known fundamental robust aggregation methods. Notably, when a maximum sparsity constraint is applied to the FC layer, i.e., $\Pi_F(\delta, \delta^{max}_{FC})$, the accuracy of the best-performing robust aggregator drops below $23\%$, and the average performance across all eight aggregators falls below $14\%$, as also illustrated in the table \ref{tab:attack_success}.

When we compare the performance of two HSA variations, namely, $\Pi_F(\delta)$ and $\Pi_F(\delta, \delta^{max}_{FC})$, we observe that, in parallel to our initial design objectives, introducing a constraint on the sparsity ratio for certain layers, i.e., FC layer, enhances the impact of the attack against robust aggregators performing layer-wise investigation such as GAS. Another important observation, which underscores the significance of imposing a constraint on layer-wise sparsity and leveraging side information about the network topology, is the performance gap observed between $\Pi_F(\delta)$ and $\Pi_F(\delta, \delta^{\text{max}}_{\text{FC}})$ against RFA. We note that the FORCE framework, like other common network pruning strategies, tends to over-emphasize the FC layer. Moreover, the weights in the FC layer are often larger than those in other layers. As a result, even when a sparse mask is applied in the HSA, the perturbations may still remain detectable under RFA.

The numerical results in Table \ref{tab:nonIID-main}  also show that our proposed DHSA strategy outperforms two other existing dynamic attack formations, namely Min-Sum and Min-Max, especially when GAS is the underlying robust aggregator. However, interestingly, we have also observed that against CM, DHSA is outperformed by its counterparts, Min-Sum and Min-Max. To understand the underlying reason, we examine the histogram plots of the scaling coefficient which is illustrated in Fig.~\ref{fig:z_scalesDir}. Histogram plots indicate that DHSA utilizes, on average, a much smaller scaling coefficient $z$ compared to Min-Sum and Min-Max. Besides, compared to the IID scenario, we observe that for DHSA the average value of scaling coefficient $z_{1}$ is decreased, while for Min-Sum and Min-Max the average value of $z$ either remained constant or increased. When we take a closer look on the numerical values, we observe that the targeted locations in the network, where DHSA use a larger fixed scaling coefficient $z_{2}=1.5$, often exhibits larger index-wise variance estimation, hence it consumes the main portion of the perturbation budget. In order to make DHSA more effective, one can consider employing a more stable variance estimation strategy, which we further explain in the Appendix \ref{sec:IQR}. 

\begin{figure}
    \centering    \includegraphics[width=0.8\linewidth]{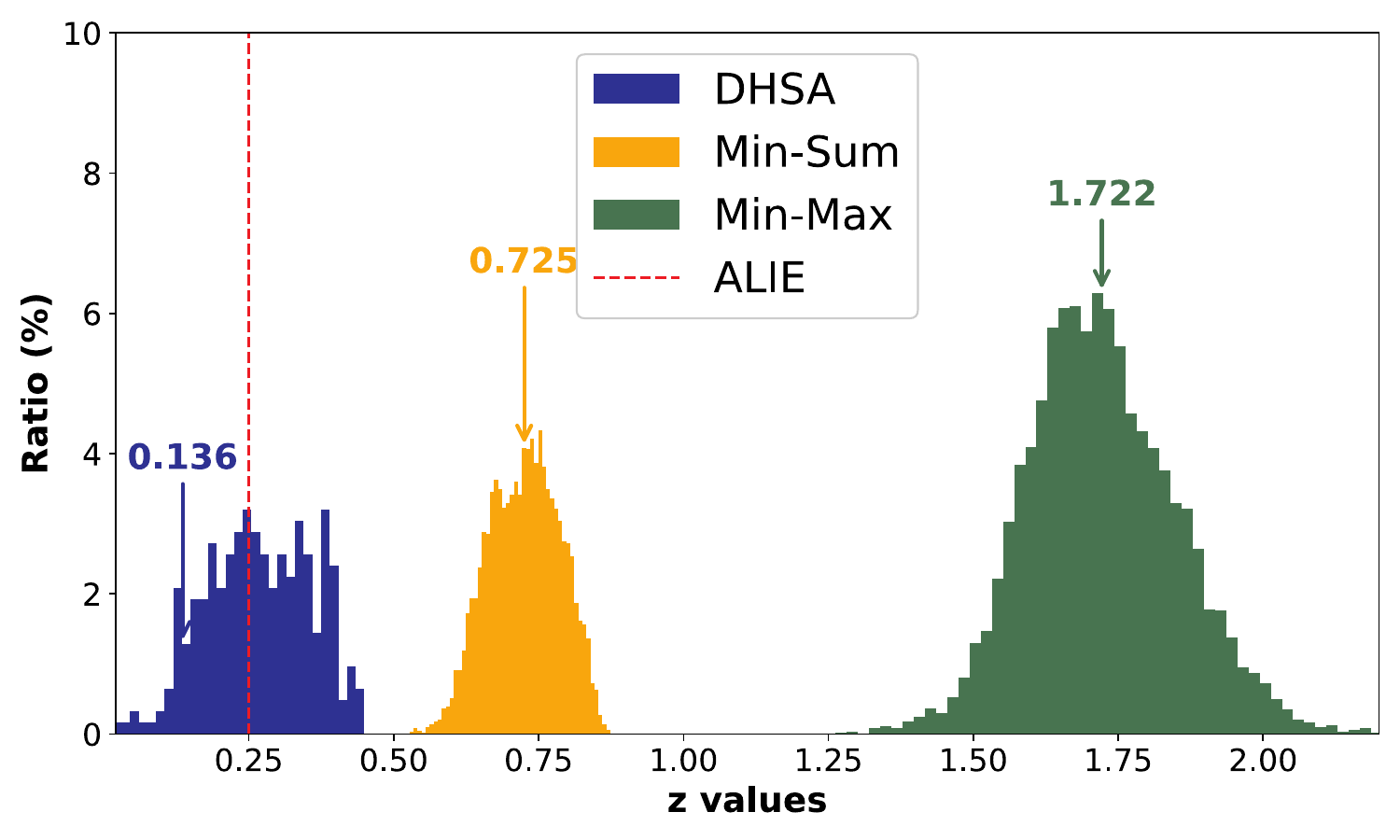}
    \caption{Histogram of $z$ values generated by the DHSA, Min-Sum and Min-Max attacks in the \textit{non-IID} setting with the TM aggregator.}
    \label{fig:z_scalesDir}
    \vspace{-5mm}
\end{figure}
Overall, we notice that our HSA is extremely effective in the non-IID setting, and our DHSA is the most effective attack in the IID setting, regardless of the defence strategy. Based on the malicious clients' data distribution, IID vs. non-IID is known by the adversary. Thus, we affirmatively answer our research question, and realize that there can be universal Byzantine attacks against FL defence mechanisms.

Figure~\ref{fig:cifar-10} illustrates the test accuracy convergence over 100 epochs for all 9 robust aggregators under 8 Byzantine attack strategies in both IID and non-IID settings. The results show that in non-IID scenarios, robust aggregators dramatically fail to prevent most attacks, with test accuracy remaining consistently low throughout training. In contrast, the IID setting demonstrates more varied defence performance across different attack-aggregator combinations.

\subsection{Complexity Analysis}
\begin{table}[h!]
\centering
\caption{Average attack generation times of the untargeted Byzantine attacks for total clients of $k=25$, $k=50$, and $k=100$ reported in milliseconds (ms). Bit-flip and Label-flip attacks also include gradient calculation times.}
\label{tab:ATK-runtime}
\resizebox{\columnwidth}{!}{%
\begin{tabular}{ccccc}
\hline
ATK | $k$ & $k=25$ & $k=50$ & $k=100$ & Coplexity \\ \hline
ALIE \cite{ALIE} & 0.47 & 0.91 & 2.64 & $O(dk)$ \\
Bit-Flip* \cite{RSA-def} & 1.24 - 13.3 & 1.14 - 13.3 & 1.69 - 13.2 & $O(d)$ \\
Label-Flip* \cite{Label_flip1} & 0.05 - 12.5 & 0.06 - 12-7 & 0.06 - 12.7 & $O(c)$ \\
IPM \cite{IPM} & 0.43 & 0.91 & 2.06 & $O(dk)$ \\
ROP \cite{rop} & 1.91 & 1.92 & 5.53 & $O(dk)$ \\
\textbf{HSA (ours)} & 0.47 & 1.00 & 2.68 & $O(dk)$ \\
\midrule
Min-Max \cite{ndss2022_Attack}& 31.75 & 63.97 & 213.78 & $O(dk^2)$ \\
Min-Sum \cite{ndss2022_Attack} & 33.24 & 68.09 & 205.15 & $O(dk^2)$ \\
\textbf{DHSA (ours)} & 31.40 & 61.35 & 190.28 & $O(dk^2)$ \\ \hline
\end{tabular}%
}%
\end{table}

In this section, we go over the runtime of the untargeted poisoning attacks in practical scenarios. We experimented with $k=[25,50,100]$ total clients and calculated the average attack generation time on ResNet-20 architecture, which has a total of $\sim7\times10^{5}$ trainable parameters ($d\cong 700K$). All simulations run on the NVIDIA A40 GPU and are averaged over 1000 aggregations. We report the running time (ms) of each attack in Table \ref{tab:ATK-runtime}. 

For our non-omniscient attacks, such as bit-flip or label-flip, we can achieve near-constant complexity if we do not include the gradient calculation. Furthermore, since these attacks do not employ updates from benign clients, their runtime is also not affected with respect to the total number of clients. Omniscient attacks such as IPM, ALIE, ROP, and proposed HSA can achieve asymptotic complexity of $O(dk)$ with very similar run times. However, ROP employs multiple operations from angle manipulation to vector rejection, and thus incurs slightly higher runtime than IPM and ALIE.

Optimized attacks, namely Min-Max, Min-Sum, and our proposed DHSA, have asymptotic complexity of $O(dk^2)$ due to computing pairwise client distances during their binary search optimization process. Consequently, their actual runtime is considerably slower than the non-optimized omniscient attacks. Among the optimized attacks, proposed DHSA achieves faster execution at larger client counts, particularly at $k=100$, due to faster convergence in its binary search as the sparse perturbation component reduces the search space.

\section{Conclusions}
In this work, we introduced a novel Byzantine attack framework, named Hybrid Sparse Attack (HSA), that is simultaneously effective against different types of defence mechanisms. The proposed framework generates an attack by combining two principles in a judicious manner: a sparse but aggressive and a dense but stealthy component, together forming a stronger but less perceptible attack. Using a network pruning framework, we identify the sensitive locations of the model architecture being trained, which are then targeted by the sparse component of the proposed attack strategy to increase its strength. Finally, through extensive simulations over different NNs, datasets, and defence mechanisms, we demonstrated the effectiveness of the proposed HSA approach. To the best of our knowledge, this is the first work in the literature to propose a Byzantine attack strategy that adapts opportunistically by exploiting the underlying architectural structure. We emphasize that the same approach can be applied to other existing attacks, revealing yet another vulnerability in current defence mechanisms and underscoring the need for further research in this area. 

In this work, our focus is limited to extracting NN architecture-based side information, specifically, identifying important or sensitive weights, to appropriately scale the base adversarial perturbation. 
As a future research direction, we plan to develop a framework where the direction of the perturbations is also determined with respect to the NN architecture. 
\bibliographystyle{IEEEtran}
\bibliography{main}

@inproceedings{backdoor_FL,
  title={How to backdoor federated learning},
  author={Bagdasaryan, Eugene and Veit, Andreas and Hua, Yiqing and Estrin, Deborah and Shmatikov, Vitaly},
  booktitle={AISTATS},
  year={2020}}

@inproceedings{RSA-def,
  title={RSA: Byzantine-robust stochastic aggregation methods for distributed learning from heterogeneous datasets},
  author={Li, Liping and Xu, Wei and Chen, Tianyi and Giannakis, Georgios B and Ling, Qing},
  booktitle={Proceedings of the AAAI Conference on Artificial Intelligence},
  volume={33},
  number={01},
  pages={1544--1551},
  year={2019}
}

@ARTICLE{signSGD_2,
  author={Jin, Richeng and Liu, Yuding and Huang, Yufan and He, Xiaofan and Wu, Tianfu and Dai, Huaiyu},
  journal={TNNLS}, 
  title={Sign-Based Gradient Descent With Heterogeneous Data: Convergence and Byzantine Resilience}, 
  year={2024}}

@article{mean-median,
  title={Distributed training with heterogeneous data: Bridging median-and mean-based algorithms},
  author={Chen, Xiangyi and Chen, Tiancong and Sun, Haoran and Wu, Steven Z and Hong, Mingyi},
  journal={Advances in Neural Information Processing Systems},
  volume={33},
  pages={21616--21626},
  year={2020}
}

@inproceedings{ALIE,
 author = {Baruch, Gilad and Baruch, Moran and Goldberg, Yoav},
 booktitle = {NeurIPS},
 title = {A Little Is Enough: Circumventing Defenses For Distributed Learning},
 year = {2019}
}

@InProceedings{IPM,
  title = 	 {Fall of Empires: Breaking Byzantine-tolerant SGD by Inner Product Manipulation},
  author = {Xie, Cong and Koyejo, Oluwasanmi and Gupta, Indranil},
  booktitle = {Uncertainty in Artificial Intelligence},
  year = 	 {2020}}

@inproceedings{LMPA,
author = {Fang, Minghong and Cao, Xiaoyu and Jia, Jinyuan and Gong, Neil Zhenqiang},
title = {Local Model Poisoning Attacks to Byzantine-Robust Federated Learning},
year = {2020},
booktitle = {USENIX Conference on Security Symposium}}

@inproceedings{ndss2022_Attack,
  title={Manipulating the Byzantine: Optimizing Model Poisoning Attacks and Defenses for Federated Learning},
  author={Virat Shejwalkar and Amir Houmansadr},
  booktitle={NDSS},
  year={2021}
}

@ARTICLE{rop,
  author={Özfatura, Kerem and Özfatura, Emre and Küpçü, Alptekin and Gunduz, Deniz},
  journal={IEEE Transactions on Information Forensics and Security}, 
  title={Byzantines Can Also Learn From History: Fall of Centered Clipping in Federated Learning}, 
  year={2024},
  volume={19},
  number={},
  pages={2010-2022},
  doi={10.1109/TIFS.2023.3345171}}

@inproceedings{
DBA,
title={DBA: Distributed Backdoor Attacks against Federated Learning},
author={Chulin Xie and Keli Huang and Pin-Yu Chen and Bo Li},
booktitle={ICLR},
year={2020}}

@article{SGD.nesterov,
  title={A method for solving the convex programming problem with convergence rate $O(1/k^2)$},
  author={Y. Nesterov},
  journal={Proceedings of the USSR Academy of Sciences},
  year={1983}}

@InProceedings{SGD.init,
  title = 	 {On the importance of initialization and momentum in deep learning},
  author = 	 {Sutskever, Ilya and Martens, James and Dahl, George and Hinton, Geoffrey},
  booktitle = 	 {ICML},
  year = 	 {2013}}

@article{SGD.polyak,
title = {Some methods of speeding up the convergence of iteration methods},
journal = {USSR Computational Mathematics and Mathematical Physics},
year = {1964},
author = {B.T. Polyak}}

@article{cosine_profiling,
  title={Byzantine-robust federated machine learning through adaptive model averaging},
  author={Mu{\~n}oz-Gonz{\'a}lez, Luis and Co, Kenneth T and Lupu, Emil C},
  journal={arXiv preprint arXiv:1909.05125},
  year={2019}
}

@inproceedings{Krum,
 author = {Blanchard, Peva and El Mhamdi, El Mahdi and Guerraoui, Rachid and Stainer, Julien},
 booktitle = {NIPS},
 title = {Machine Learning with Adversaries: Byzantine Tolerant Gradient Descent},
 year = {2017}
}

@inproceedings{Bulyan,
author = {El Mhamdi, El Mahdi and Guerraoui, Rachid and Rouault, Sébastien},
year = {2018},
booktitle = {ICML},
title = {The Hidden Vulnerability of Distributed Learning in Byzantium}
}

@ARTICLE{RFA,
  author={Pillutla, Krishna and Kakade, Sham M. and Harchaoui, Zaid},
  journal={IEEE Transactions on Signal Processing}, 
  title={Robust Aggregation for Federated Learning}, 
  year={2022},
  }

@InProceedings{Trimmed_mean,
  title = 	 {{B}yzantine-Robust Distributed Learning: Towards Optimal Statistical Rates},
  author =       {Yin, Dong and Chen, Yudong and Kannan, Ramchandran and Bartlett, Peter},
  booktitle = {ICML},
  year = 	 {2018}}

@inproceedings{SignSGD,
  title={signSGD with Majority Vote is Communication Efficient and Fault Tolerant},
  author={Jeremy Bernstein and Jiawei Zhao and Kamyar Azizzadenesheli and Anima Anandkumar},
  booktitle={ICLR},
  year={2019}
}

@inproceedings{B2DB,
  title={Back to the drawing board: A critical evaluation of poisoning attacks on production federated learning},
  author={Shejwalkar, Virat and Houmansadr, Amir and Kairouz, Peter and Ramage, Daniel},
  booktitle={IEEE SP},
  year={2022}}

@InProceedings{AFLAL,
  title = {Analyzing Federated Learning through an Adversarial Lens},
  author = {Bhagoji, Arjun Nitin and Chakraborty, Supriyo and Mittal, Prateek and Calo, Seraphin},
  booktitle = {ICML},
  year = 	 {2019}}

@article{cross-device,
  title={Advances and open problems in federated learning},
  author={Kairouz, Peter and McMahan, H Brendan and Avent, Brendan and Bellet, Aur{\'e}lien and Bennis, Mehdi and Bhagoji, Arjun Nitin and Bonawitz, Kallista and Charles, Zachary and Cormode, Graham and Cummings, Rachel and others},
  journal={Foundations and trends{\textregistered} in machine learning},
  volume={14},
  number={1--2},
  pages={1--210},
  year={2021},
  publisher={Now Publishers, Inc.}
}

@inproceedings{Byz.momentum,
  author    = {El-Mahdi El-Mhamdi and
               Rachid Guerraoui and
               S\'{e}bastien Rouault},
  title     = {Distributed Momentum for Byzantine-resilient Stochastic Gradient Descent},
  booktitle = {ICLR},
  year      = {2021}
}

@inproceedings{CC,
  title={Learning from history for Byzantine robust optimization},
  author={Karimireddy, Sai Praneeth and He, Lie and Jaggi, Martin},
    year={2021},
  booktitle={ICML}}

@article{SCC,
  author    = {Lie He andfan
               Sai Praneeth Karimireddy and
               Martin Jaggi},
  title     = {Byzantine-Robust Decentralized Learning via Self-Centered Clipping},
  journal   = {ArXiv},
  year      = {2022}
}

@inproceedings{
Bucketing,
title={Byzantine-Robust Learning on Heterogeneous Datasets via Bucketing},
author={Sai Praneeth Karimireddy and Lie He and Martin Jaggi},
booktitle={ICLR},
year={2022}}

@inproceedings{momentum_def,
  title={Byzantine machine learning made easy by resilient averaging of momentums},
  author={Farhadkhani, Sadegh and Guerraoui, Rachid and Gupta, Nirupam and Pinot, Rafael and Stephan, John},
  booktitle={ICML},
  year={2022}}

@article{cifar10,
title= {CIFAR-10 (Canadian Institute for Advanced Research)},
author= {Alex Krizhevsky and Vinod Nair and Geoffrey Hinton}}

@inproceedings{resnet,
  title={Deep residual learning for image recognition},
  author={He, Kaiming and Zhang, Xiangyu and Ren, Shaoqing and Sun, Jian},
  booktitle={Proceedings of the IEEE conference on computer vision and pattern recognition},
  year={2016}
}

@misc{dirichlet,
  title={Estimating a Dirichlet distribution},
  author={Minka, Thomas},
  year={2000},
  publisher={Technical report, MIT}
}

@inproceedings{FedAVG1,
  title={Communication-efficient learning of deep networks from decentralized data},
  author={McMahan, Brendan and Moore, Eider and Ramage, Daniel and Hampson, Seth and y Arcas, Blaise Aguera},
  booktitle={AISTATS},
  year={2017}}

@article{fedAVG_2,
  title={Federated learning: Strategies for improving communication efficiency},
  author={Kone{\v{c}}n{\`y}, Jakub and McMahan, H Brendan and Yu, Felix X and Richt{\'a}rik, Peter and Suresh, Ananda Theertha and Bacon, Dave},
  journal={NIPS workshop on Private Multiparty Machine Learning},
  year={2016}
}

@article{byzantine_orig,
author = {Lamport, Leslie and Shostak, Robert and Pease, Marshall},
title = {The Byzantine Generals Problem},
year = {1982},
journal = {ACM Trans. Program. Lang. Syst.}}

@inproceedings{FL_limitations,
  title={The Limitations of Federated Learning in Sybil Settings.},
  author={Fung, Clement and Yoon, Chris JM and Beschastnikh, Ivan},
  booktitle={Usenix RAID},
  year={2020}
}

@inproceedings{ByzSGD_def,
author = {El-Mhamdi, El-Mahdi and Guerraoui, Rachid and Guirguis, Arsany and Hoang, L\^{e} Nguy\^{e}n and Rouault, S\'{e}bastien},
title = {Genuinely Distributed Byzantine Machine Learning},
year = {2020},
publisher = {Association for Computing Machinery},
series = {PODC '20}
}

@article{backdoor_norm,
  title={Can you really backdoor federated learning?},
  author={Sun, Ziteng and Kairouz, Peter and Suresh, Ananda Theertha and McMahan, H Brendan},
  journal={arXiv preprint arXiv:1911.07963},
  year={2019}
}

@inproceedings{FLtrust,
  title={FLTrust: Byzantine-robust Federated Learning via Trust Bootstrapping},
  author={Cao, Xiaoyu and Fang, Minghong and Liu, Jia and Gong, Neil Zhenqiang},
  booktitle={NDSS},
  year={2021}
}

@inproceedings{zeno_def,
  title={Zeno: Distributed stochastic gradient descent with suspicion-based fault-tolerance},
  author={Xie, Cong and Koyejo, Sanmi and Gupta, Indranil},
  booktitle={ICML},
  year={2019}}

@inproceedings{GAS,
author = {Liu, Yuchen and Chen, Chen and Lyu, Lingjuan and Wu, Fangzhao and Wu, Sai and Chen, Gang},
title = {Byzantine-robust learning on heterogeneous data via gradient splitting},
year = {2023},
booktitle = {ICML}}

@inproceedings{Label_flip1,
author = {Biggio, Battista and Nelson, Blaine and Laskov, Pavel},
title = {Poisoning Attacks against Support Vector Machines},
year = {2012},
booktitle = {ICML}}

@article{prune_fund1,
  title={Scalable training of artificial neural networks with adaptive sparse connectivity inspired by network science},
  author={Mocanu, Decebal Constantin and Mocanu, Elena and Stone, Peter and Nguyen, Phuong H and Gibescu, Madeleine and Liotta, Antonio},
  journal={Nature communications},
  volume={9},
  number={1},
  pages={2383},
  year={2018},
  publisher={Nature Publishing Group UK London}
}

@inproceedings{prune_fund2,
  title={Parameter efficient training of deep convolutional neural networks by dynamic sparse reparameterization},
  author={Mostafa, Hesham and Wang, Xin},
  booktitle={International Conference on Machine Learning},
  pages={4646--4655},
  year={2019},
  organization={PMLR}
}

@inproceedings{
prune0,
title={The Lottery Ticket Hypothesis: Finding Sparse, Trainable Neural Networks},
author={Jonathan Frankle and Michael Carbin},
booktitle={ICLR},
year={2019}}

@inproceedings{
prune_force,
title={Progressive Skeletonization: Trimming more fat from a network at initialization},
author={Pau de e and Amartya Sanyal and Harkirat Behl and Philip Torr and Gr{\'e}gory Rogez and Puneet K. Dokania},
booktitle={ICLR},
year={2021}
}

@inproceedings{prune_nodata,
 author = {Tanaka, Hidenori and Kunin, Daniel and Yamins, Daniel L and Ganguli, Surya},
 booktitle = {Neurips},
 title = {Pruning neural networks without any data by iteratively conserving synaptic flow},
 year = {2020}
}

@InProceedings{prune3,
  title = 	 {Efficient Lottery Ticket Finding: Less Data is More},
  author =       {Zhang, Zhenyu and Chen, Xuxi and Chen, Tianlong and Wang, Zhangyang},
  booktitle = 	 {ICML},
  year = 	 {2021}
}

@InProceedings{prune6,
  title = 	 {Rigging the Lottery: Making All Tickets Winners},
  author =       {Evci, Utku and Gale, Trevor and Menick, Jacob and Castro, Pablo Samuel and Elsen, Erich},
  booktitle = 	 {ICML},
  year = 	 {2020}
}

@article{prune7,
  title={Pruning convolutional neural networks for resource efficient inference},
  author={Molchanov, Pavlo and Tyree, Stephen and Karras, Tero and Aila, Timo and Kautz, Jan},
  journal={arXiv preprint arXiv:1611.06440},
  year={2016}
}

@inproceedings{
prune_layerwise,
title={Layer-adaptive Sparsity for the Magnitude-based Pruning},
author={Jaeho Lee and Sejun Park and Sangwoo Mo and Sungsoo Ahn and Jinwoo Shin},
booktitle={ICLR},
year={2021}
}

@inproceedings{prune_connectivity,
  title={Group fisher pruning for practical network compression},
  author={Liu, Liyang and Zhang, Shilong and Kuang, Zhanghui and Zhou, Aojun and Xue, Jing-Hao and Wang, Xinjiang and Chen, Yimin and Yang, Wenming and Liao, Qingmin and Zhang, Wayne},
  booktitle={ICML},
  year={2021}}

@inproceedings{
prune_connectivity2,
title={The Unreasonable Effectiveness of Random Pruning: Return of the Most Naive Baseline for Sparse Training},
author={Shiwei Liu and Tianlong Chen and Xiaohan Chen and Li Shen and Decebal Constantin Mocanu and Zhangyang Wang and Mykola Pechenizkiy},
booktitle={International Conference on Learning Representations},
year={2022}}

@inproceedings{
snip,
title={{SNIP}: {SINGLE}-{SHOT} {NETWORK} {PRUNING} {BASED} {ON} {CONNECTION} {SENSITIVITY}},
author={Namhoon Lee and Thalaiyasingam Ajanthan and Philip Torr},
booktitle={ICLR},
year={2019}}

@inproceedings{
prune_at_init2,
title={Rare Gems: Finding Lottery Tickets at Initialization},
author={Kartik Sreenivasan and Jy-yong Sohn and Liu Yang and Matthew Grinde and Alliot Nagle and Hongyi Wang and Eric Xing and Kangwook Lee and Dimitris Papailiopoulos},
booktitle={Neurips},
year={2022}}

@inproceedings{
prune_at_init3,
title={Dual Lottery Ticket Hypothesis},
author={Yue Bai and Huan Wang and ZHIQIANG TAO and Kunpeng Li and Yun Fu},
booktitle={ICLR},
year={2022}}

@article{powerSGD,
  title={PowerSGD: Practical low-rank gradient compression for distributed optimization},
  author={Vogels, Thijs and Karimireddy, Sai Praneeth and Jaggi, Martin},
  journal={Neurips},
  volume={32},
  year={2019}
}

@inproceedings{sec.vr1,
title={Variance Reduction is an Antidote to Byzantines: Better Rates, Weaker Assumptions and Communication Compression as a Cherry on the Top},
author={Eduard Gorbunov and Samuel Horv{\'a}th and Peter Richt{\'a}rik and Gauthier Gidel},
booktitle={ICLR},
year={2023}}

@article{sec.vr2,
title = {Byzantine-robust variance-reduced federated learning over distributed non-i.i.d. data},
journal = {Information Sciences},
year = {2022},
author = {Jie Peng and Zhaoxian Wu and Qing Ling and Tianyi Chen}}

@ARTICLE{sec.vr3,

  author={Wu, Zhaoxian and Ling, Qing and Chen, Tianyi and Giannakis, Georgios B.},

  journal={IEEE Transactions on Signal Processing}, 

  title={Federated Variance-Reduced Stochastic Gradient Descent With Robustness to Byzantine Attacks}, 

  year={2020}}

@INPROCEEDINGS{sec.vr4,

  author={Zhu, Heng and Ling, Qing},

  booktitle={ICASSP}, 

  title={Byzantine-Robust Aggregation with Gradient Difference Compression and Stochastic Variance Reduction for Federated Learning}, 

  year={2022}}

@INPROCEEDINGS{sec.vr5,

  author={Peng, Jie and Li, Weiyu and Ling, Qing},

  booktitle={ICASSP}, 

  title={Variance Reduction-Boosted Byzantine Robustness in Decentralized Stochastic Optimization}, 

  year={2022}}

@article{decentCC,
  title={Can Decentralized Learning be more robust than Federated Learning?},
  author={Raynal, Mathilde and Pasquini, Dario and Troncoso, Carmela},
  journal={arXiv},
  year={2023}
}

@inproceedings{PODC-BYZ_DSGD,
author = {Gupta, Nirupam and Vaidya, Nitin H.},
title = {Fault-Tolerance in Distributed Optimization: The Case of Redundancy},
year = {2020},
isbn = {9781450375825},
publisher = {Association for Computing Machinery},
address = {New York, NY, USA},
url = {https://doi.org/10.1145/3382734.3405748},
doi = {10.1145/3382734.3405748},
booktitle = {Proceedings of the 39th Symposium on Principles of Distributed Computing},
pages = {365–374},
numpages = {10},
location = {Virtual Event, Italy},
series = {PODC '20}
}

@article{dirichlet1,
  title={Robust federated learning with attack-adaptive aggregation},
  author={Wan, Ching Pui and Chen, Qifeng},
  journal={arXiv},
  year={2021}
}

@InProceedings{dirichlet2,
author="Gupta, Ashish
and Luo, Tie
and Ngo, Mao V.
and Das, Sajal K.",
title="Long-Short History of Gradients Is All You Need: Detecting Malicious and Unreliable Clients in Federated Learning",
booktitle="ESORICS",
year="2022"}

@ARTICLE{dirichlet_cited,
  author={Yin, Chunyong and Zeng, Qingkui},
  journal={IEEE Transactions on Computational Social Systems}, 
  title={Defending Against Data Poisoning Attack in Federated Learning With Non-IID Data}, 
  year={2023}}

@article{game_theory,
  title={Mixed Nash for Robust Federated Learning},
  author={Xie, Wanyun and Pethick, Thomas and Ramezani-Kebrya, Ali and Cevher, Volkan},
  journal={TMLR},
  year={2023}
}

@article{fedprox,
  title={Federated optimization in heterogeneous networks},
  author={Li, Tian and Sahu, Anit Kumar and Zaheer, Manzil and Sanjabi, Maziar and Talwalkar, Ameet and Smith, Virginia},
  journal={Proceedings of Machine learning and systems},
  volume={2},
  pages={429--450},
  year={2020}
}

@INPROCEEDINGS{FedADC,
  author={Ozfatura, Emre and Ozfatura, Kerem and Gündüz, Deniz},
  booktitle={ISIT}, 
  title={FedADC: Accelerated Federated Learning with Drift Control}, 
  year={2021},
  volume={},
  number={},
  pages={467-472},
  doi={10.1109/ISIT45174.2021.9517850}}

@inproceedings{
fedDyn,
title={Federated Learning Based on Dynamic Regularization},
author={Durmus Alp Emre Acar and Yue Zhao and Ramon Matas and Matthew Mattina and Paul Whatmough and Venkatesh Saligrama},
booktitle={ICLR},
year={2021}
}

@article{adamOPT,
  title={Adam: A method for stochastic optimization},
  author={Kingma, Diederik P and Ba, Jimmy},
  journal={ICLR},
  year={2015}
}

@inproceedings{scaffold,
  title={Scaffold: Stochastic controlled averaging for federated learning},
  author={Karimireddy, Sai Praneeth and Kale, Satyen and Mohri, Mehryar and Reddi, Sashank and Stich, Sebastian and Suresh, Ananda Theertha},
  booktitle={ICML},
  year={2020}}

@InProceedings{FedDC,
    author    = {Gao, Liang and Fu, Huazhu and Li, Li and Chen, Yingwen and Xu, Ming and Xu, Cheng-Zhong},
    title     = {FedDC: Federated Learning With Non-IID Data via Local Drift Decoupling and Correction},
    booktitle = {CVPR},
    year      = {2022}}

@article{SlowMo,
  author       = {Jianyu Wang and
                  Vinayak Tantia and
                  Nicolas Ballas and
                  Michael G. Rabbat},
  title        = {SlowMo: Improving Communication-Efficient Distributed {SGD} with Slow
                  Momentum},
  journal      = {ICLR},
  year         = {2020}}

@inproceedings{buyukates_fedsecurity, author = {Han, Shanshan and Buyukates, Baturalp and Hu, Zijian and Jin, Han and Jin, Weizhao and Sun, Lichao and Wang, Xiaoyang and Wu, Wenxuan and Xie, Chulin and Yao, Yuhang and Zhang, Kai and Zhang, Qifan and Zhang, Yuhui and Joe-Wong, Carlee and Avestimehr, Salman and He, Chaoyang}, title = {{FedSecurity}: A Benchmark for Attacks and Defenses in Federated Learning and Federated {LLMs}}, year = {2024}, booktitle = {Proceedings of the 30th ACM SIGKDD Conference on Knowledge Discovery and Data Mining}, pages = {5070–5081} }

@article{han2023kick,
  title={Kick Bad Guys Out! Conditionally Activated Anomaly Detection in Federated Learning with Zero-Knowledge Proof Verification},
  author={Han, Shanshan and Wu, Wenxuan and Buyukates, Baturalp and Jin, Weizhao and Zhang, Qifan and Yao, Yuhang and Avestimehr, Salman and He, Chaoyang},
  journal={arXiv preprint arXiv:2310.04055},
  year={2023}
}

@article{IQR-stats,
  title={Robust statistics for outlier detection},
  author={Rousseeuw, Peter J and Hubert, Mia},
  journal={Wiley interdisciplinary reviews: Data mining and knowledge discovery},
  volume={1},
  number={1},
  pages={73--79},
  year={2011},
  publisher={Wiley Online Library}
}

@INPROCEEDINGS{L-method,
  author={Salvador, S. and Chan, P.},
  booktitle={16th IEEE International Conference on Tools with Artificial Intelligence}, 
  title={Determining the number of clusters/segments in hierarchical clustering/segmentation algorithms}, 
  year={2004},
  volume={},
  number={},
  pages={576-584},
  doi={10.1109/ICTAI.2004.50}}

@INPROCEEDINGS{L-method2,
  author={Satopaa, Ville and Albrecht, Jeannie and Irwin, David and Raghavan, Barath},
  booktitle={2011 31st International Conference on Distributed Computing Systems Workshops}, 
  title={Finding a "Kneedle" in a Haystack: Detecting Knee Points in System Behavior}, 
  year={2011},
  volume={},
  number={},
  pages={166-171},
  doi={10.1109/ICDCSW.2011.20}}

@INPROCEEDINGS{FoundationFL,
  title={Do We Really Need to Design New Byzantine-robust Aggregation Rules?},
  author={Fang, Minghong and Nabavirazavi, Seyedsina and Liu, Zhuqing and Sun, Wei and Iyengar, Sundaraja Sitharama and Yang, Haibo},
  booktitle={NDSS},
  year={2025}
}

@inproceedings{lasa,
  title={Achieving byzantine-resilient federated learning via layer-adaptive sparsified model aggregation},
  author={Xu, Jiahao and Zhang, Zikai and Hu, Rui},
  booktitle={2025 IEEE/CVF Winter Conference on Applications of Computer Vision (WACV)},
  pages={1508--1517},
  year={2025},
  organization={IEEE}
}

@inproceedings{shejwalkar2021manipulating,
  title={Manipulating the byzantine: Optimizing model poisoning attacks and defenses for federated learning},
  author={Shejwalkar, Virat and Houmansadr, Amir},
  booktitle={NDSS},
  year={2021}
}

@inproceedings{nguyen2022flame,
  title={$\{$FLAME$\}$: Taming backdoors in federated learning},
  author={Nguyen, Thien Duc and Rieger, Phillip and Chen, Huili and Yalame, Hossein and M{\"o}llering, Helen and Fereidooni, Hossein and Marchal, Samuel and Miettinen, Markus and Mirhoseini, Azalia and Zeitouni, Shaza and others},
  booktitle={31st USENIX security symposium (USENIX Security 22)},
  pages={1415--1432},
  year={2022}
}

@inproceedings{zhang2022fldetector,
  title={Fldetector: Defending federated learning against model poisoning attacks via detecting malicious clients},
  author={Zhang, Zaixi and Cao, Xiaoyu and Jia, Jinyuan and Gong, Neil Zhenqiang},
  booktitle={Proceedings of the 28th ACM SIGKDD conference on knowledge discovery and data mining},
  pages={2545--2555},
  year={2022}
}

@inproceedings{xie2024fedredefense,
  title={FedREDefense: Defending against model poisoning attacks for federated learning using model update reconstruction error},
  author={Xie, Yueqi and Fang, Minghong and Gong, Neil Zhenqiang},
  year={2024},
  organization={International Conference on Machine Learning}
}

@inproceedings{yan2024skymask,
  title={SkyMask: Attack-agnostic robust federated learning with fine-grained learnable masks},
  author={Yan, Peishen and Wang, Hao and Song, Tao and Hua, Yang and Ma, Ruhui and Hu, Ningxin and Haghighat, Mohammad Reza and Guan, Haibing},
  booktitle={European Conference on Computer Vision},
  pages={291--308},
  year={2024},
  organization={Springer}
}

@article{jebreel2023fl,
  title={Fl-defender: Combating targeted attacks in federated learning},
  author={Jebreel, Najeeb Moharram and Domingo-Ferrer, Josep},
  journal={Knowledge-Based Systems},
  volume={260},
  pages={110178},
  year={2023},
  publisher={Elsevier}
}
\newpage
\appendices

\section{Summary of appendices}
This appendix provides comprehensive supplementary material, including technical foundations, extended experimental results, and detailed ablation studies to support the main findings of this work. 

Appendix~\ref{sec:network_pruning} presents the technical foundations of iterative network pruning methodologies employed in HSA, including the Erdos-Renyi Kernel (ERK) for layer-wise sparsity distribution, pruning with saliency measures (FORCE and SNIP algorithms), layer-constrained mask generation, and sparsity scheduling strategies (Appendix~\ref{sch}) used to construct topology-aware sparse attack masks. 

Appendix~\ref{ans} extends the experimental validation across diverse settings, presenting additional numerical results on multiple datasets (F-MNIST, MNIST) and network architectures (CNN, MLP), cross-device federated learning scenarios with varying client populations, and partial knowledge settings (Appendix~\ref{sec:no_known_grad}) where Byzantine clients have limited information about benign client gradients. 

Appendix~\ref{sec:ablation_on_sparsity} conducts extensive ablation studies on network sparsity, systematically comparing random sparsity patterns, ERK-based sparsity distributions, and FORCE pruning methods across various sparsity levels and layer-wise constraints, demonstrating the critical importance of network topology awareness for maximizing attack effectiveness. Further illustrations of various sparsity regimes are also provided in Appendix~\ref{sec:figsForce}.

Appendix~\ref{app:AggregatorCombs} investigates robust aggregation methods through multiple perspectives, examining sequential combination strategies that apply multiple aggregation rules in succession, analyzing hybrid defence mechanisms that combine multiple robust aggregators, and providing a detailed analysis of index-wise elimination versus coordinate sanitization mechanisms, including our proposed TM-Oracle, CliM, and TM-History approaches.

Appendix~\ref{app:Complexity} presents a complexity and running time analysis of Byzantine-robust aggregators, providing empirical runtime measurements and asymptotic complexity bounds for various defence mechanisms under different client population sizes.

Appendix~\ref{sec:IQR} addresses numerical instability challenges in non-IID simulations, introducing variance stabilization techniques including IQR-based thresholding and gradient-based elbow methods for HSA and DHSA attacks.

Finally, in Appendix~\ref{sec:ablations on Robust Aggregators}, we analyze state-of-the-art defence methods beyond classical robust aggregators, examining recent approaches such as LASA \cite{lasa} and FoundationFL \cite{FoundationFL}, while demonstrating counter-attack strategies that expose vulnerabilities in these advanced defence mechanisms.

\section{network Pruning} \label{sec:network_pruning}
\subsection{Erdos-Renyi Kernel (ERK)}
Authors of \cite{prune_fund2} model the connection between any two neurons that belong to two successive layers of the NN architecture as {\em Erdos-Renyi random graph}:
\begin{equation}
p(w^{\ell}_{i,j}=1) = \epsilon \frac{n^{\ell}+n^{\ell-1}}{n^{\ell} n^{\ell-1}},
\end{equation}
where $w^{\ell}_{i,j}$ denotes the connection between the $i$th neuron from the $\ell$th layer and $j$th neuron from the $\ell-1$th layer, $n^{\ell}$ denotes the number of neurons in $\ell$th layer, and $\epsilon$ is the scaling parameter controlling the sparsity layer. Hence, the Erdos-Renyi approach in network pruning states the sparsity of each layer should be correlated with input/output neuron numbers.

ERK approach, introduced in \cite{prune_fund1}, extends the Erdos-Renyi approach to convolutional layers, including kernel dimensions. That is, the sparsity of the convolutional layers are proportional to:
\begin{equation}
\frac{n^{\ell}+n^{\ell-1}+w^{\ell}+h^{\ell}}{n^{\ell} n^{\ell-1} w^{\ell} h^{\ell}},
\end{equation}
where $w^{\ell}$ and $ h^{\ell}$ are the kernel dimensions (width and height, respectively) in the $\ell$th layer.

\subsection{Pruning with Saliency Measure}
In SNIP \cite{snip}, for a given $\boldsymbol{\theta}$, the additional loss incurred due to the pruning of the $i^{th}$ weight is given by:
\begin{equation}
\Delta F_{i} = F(\mathbf{1} \odot \boldsymbol{\theta}; \mathcal{D} ) - F((\mathbf{1}- \mathbf{e}_{i}) \odot \boldsymbol{\theta}; \mathcal{D}, )
\end{equation}
where $\mathbf{e}_{i}$ is a one-hot vector with all zeros except the $i^{th}$ index which is equal to one. This additionally incurred loss can be approximated by the gradient with respect to the $\mathbf{c}$:

\begin{align} \label{eq:iter_prune}
\Delta F_{i} \approx & \frac{\partial F(\mathbf{c} \odot \boldsymbol{\theta}; \mathcal{D} )}{\partial c_{i} }\bigg|_{\mathbf{c}= \mathbf{1}}\\=
&lim_{\delta \xrightarrow{}0} \frac{F(\mathbf{c} \odot \boldsymbol{\theta}; \mathcal{D} )-F((\mathbf{c}-\delta\mathbf{e}_{i}) \odot \boldsymbol{\theta}; \mathcal{D} )}{\delta}\bigg|_{\mathbf{c}= \mathbf{1}}. \nonumber
\end{align}
The term above represents a directional derivative along a sparse direction, and can therefore be written in the following form:
\begin{equation}
\boldsymbol{\theta}_{i} \frac{\partial F(\boldsymbol{\theta}; \mathcal{D} )}{\delta \boldsymbol{\theta}_{i}}.
\end{equation}
We can conclude that $s_{i} \defeq \vert \boldsymbol{\theta}_{i} \frac{\partial F(\boldsymbol{\theta}; \mathcal{D} )}{\boldsymbol{\theta}_{i}} \vert $ can be used as a saliency measure for the connection sensitivity \cite{snip}. 

In this work, we mainly leverage the FORCE pruning strategy \cite{prune_force}, which building on the same idea, measures connection sensitivity after pruning rather than before. Hence, FORCE evaluates the gradient for the foresight sparse model:
\begin{equation}
\frac{\partial F(\mathbf{c} \odot \boldsymbol{\theta}; \mathcal{D} )}{\partial \mathbf{c} }\bigg|_{\mathbf{c}= \hat{\mathbf{c}}}, ~\vert\vert\hat{\mathbf{c}}\vert\vert_{0}=\delta, \hat{\mathbf{c}}\in\left\{0,1 \right\}^d .
\end{equation}
Accordingly, FORCE searches for the binary sparse mask $\mathbf{c}$ that maximizes the following saliency measure:

\begin{equation} \label{eq:saliency_force}
S(\boldsymbol{\theta}, \mathbf{c} ) = \sum_{i \in [d]: \mathbf{c}[i]=1} \underbrace{\vert \boldsymbol{\theta} \odot \nabla F(\mathbf{c} \odot \boldsymbol{\theta}; \mathcal{D} )\vert}_{\triangleq \  \mathbf{s}(\boldsymbol{\theta}, \mathbf{c} )}[i] \ ,
\end{equation}
where $ \nabla F(\mathbf{c} \odot \boldsymbol{\theta}; \mathcal{D} )=\left(\frac{\partial F(\bar{\boldsymbol{\theta}}; \mathcal{D} )}{\partial \bar{\boldsymbol{\theta}} }\big|_{\mathbf{c}}\right)$, ~$\bar{\boldsymbol{\theta}}=\mathbf{c} ~\odot ~\boldsymbol{\theta}$, from equation (5) in  \cite{prune_force}. 

In practice, a strategy that measures connection sensitivity after pruning is not feasible due to its combinatorial complexity. Ideally, we need to compute the saliency of all possible $d \choose \lfloor d\delta \rfloor$ sparsity patterns. To mitigate combinatorial complexity, FORCE performs iterative pruning under the assumption that the change in the gradient between consecutive iterations can be neglected.

\textbf{Progressive Sparsification (FORCE):}
The iterative network pruning framework gradually prunes the NN weights over $T$ steps. Formally speaking, the sparsity mask $\mathbf{c}_{t+1}$ is chosen as the one that maximizes the saliency measure:
\begin{equation}
\mathbf{c}_{t+1}= \argmax_{\mathbf{c}} S(\bar{\boldsymbol{\theta}}, ~ \mathbf{c} ) ~ \text{s.t.} ~ \mathbf{c}\in \mathbb{R}^{d}, ~ \vert\vert \mathbf{c}\vert\vert_{0} = \lfloor\delta_{t}d\rfloor,
\end{equation}
which is conditioned on the previous mask $\mathbf{c}_{t}$, i.e.,  $\bar{\boldsymbol{\theta}}=\mathbf{c}_{t} ~\odot ~\boldsymbol{\theta}$. Here, we note that the FORCE framework differentiates from the iterative SNIP, which introduces an extra constraint on the sparsity mask, i.e., $\mathbf{c}_{t} \odot \mathbf{c} = \mathbf{c} $. That is, SNIP does not
allow previously pruned parameters to resurrect. However, the FORCE algorithm \cite{prune_force} assigns zero values to those pruned weights rather than removing them from the network, enabling the resurrection of the pruned weights later based on the saliency measure. The overall FORCE framework implemented in this work is illustrated in Algorithm \ref{code:force}. Different from the original FORCE implementation  \cite{prune_force}, we introduce a layer-wise sparsity constraint, which we next explain further.
\begin{algorithm}[t]
    \small
    \caption{FORCE with layer-wise constraint}
    \label{code:force}
    \begin{algorithmic}[1]
     \State \textbf{Inputs:} Global target sparsity $\delta$, Layer-wise sparsity constraints $\Delta=\left\{\delta^{\ell}\right\}_{\ell\in\mathcal{L}}$, Initialized weights $\boldsymbol{\theta}\in \mathbb{R}^{d}$, Number of pruning steps $T$, Colluded dataset $D_{\text{mal}.}$
     \State \textbf{Outputs:} $\mathbf{c}$
     \State Initialize $\mathbf{c}_{0} \leftarrow{} \boldsymbol{1}_{d}$
     \For{$t=1,\ldots,T$ }
      \State Schedule sparsity: $\delta_{t}=f_{sc}(t,T,\delta)$
      \State Adjust layer-wise constraint: $\Delta_{t}=f_{lc}(\Delta,\delta,\delta_{t})$
      \State Sample mini-batch: $\zeta \in D_{mal}$
      \State Compute saliency vector: $\mathbf{s}(\boldsymbol{\theta}, \mathbf{c}_{t} ,\zeta)$
     \State \textbf{Update binary sparse  mask:}
     \State$\mathbf{c}_{t} \gets \mathrm{SLC}(\mathbf{s}_t, \Delta_{t})$
     \EndFor
     \end{algorithmic}
    \label{code:PruningGeneral}
\end{algorithm}

\textbf{Layer-wise Sparsity Constraint:}
One of the main limitations of using a pruning framework to identify target positions for the Byzantine attack is that pruning can overemphasize certain layers, often those closer to the input \cite{prune0,prune7}, or the final fully connected layer that produces the logits. However, such an accumulation of non-zero positions in certain layers may make the attack visible. Hence, we consider a network pruning strategy with sparsity constraints introduced for each layer in order to avoid the accumulation of non-zero positions while still identifying the target location based on their importance in the network topology.

Hence, when employing the FORCE algorithm to generate binary sparse masks for HSA, we revise the network pruning process by introducing a minimum sparsity constraint for each layer. That is, while targeting a global sparsity ratio $\delta$, the sparsity ratio for $\ell$th layer can not exceed $\delta_{\ell}$. The overall procedure for generating a sparsity mask, based on given list of saliency measure vectors $\mathcal{S}= \left\{\mathbf{s}_{1}, \ldots, \mathbf{s}_{L}\right\}$, and layer-wise sparsity constraint $\Delta= \left\{\delta_{1}, \ldots, \delta_{L}\right\}$ is illustrated in Algorithm \ref{code:SLC}.

The algorithm first initializes a set of indices $\mathcal{A}=[L]$, which is used to keep track of layers where the corresponding binary sparse mask is not finalized yet. In each iteration, until a binary sparse mask is assigned to each layer, i.e., $\mathcal{A}=\emptyset$, the algorithm flattens the saliency vectors of the unassigned layers  (line 7), and then computes the top-k threshold for the given $n_{target}$ value (line 8). Once the threshold $u_{th}$ is computed, it is used to generate binary sparse masks for the unassigned layers (line 10). Following these initial masks, the algorithm checks if each mask satisfies the layer-wise sparsity constraint (line 12). If a mask violates the constraint, then {\em local mask generator}, illustrated in Algorithm \ref{code:lmg}, is executed to generate a binary sparse mask with layer-wise thresholding, and the corresponding layer index is removed from list $\mathcal{A}$ (line 14). The algorithm terminates when there is no binary sparse mask violating the constraint.

\begin{algorithm}[t]
    \small
    \caption{Sparsification with Layer-wise Constraint (SLC)}
    \label{code:SLC}
    \begin{algorithmic}[1]
     \State \textbf{Inputs:} Target sparsity $\delta$, Saliency measure $\mathcal{S}= \left\{\mathbf{s}_{1}, \ldots, \mathbf{s}_{L}\right\}$, 
      $\Delta= \left\{\delta_{1}, \ldots, \delta_{L}\right\}$, 
     \State \textbf{Outputs:} Binary masks $\mathcal{C}= \left\{\mathbf{c}_{1}, \ldots, \mathbf{c}_{L}\right\}$
    \State Layer-wise target: $n^{(\ell)}_{target} = \lfloor d_{\ell} \delta_{\ell} \rfloor ~~ \forall \ell\in\mathcal{L}$
    \State Global target: $n_{target}= \lfloor d \delta \rfloor$
    \State List of unassigned layers $\mathcal{A}=[L]$
    \While{$\mathcal{A}\neq \emptyset$}
    \State $\mathbf{S}= f_{flatten}(\mathcal{S},\mathcal{A})$
    \State $u_{th} = f_{top-k}(\mathbf{S},n_{target})$
    \State $\mathcal{M}= f_{gmg}(\mathcal{S},\mathcal{A}, u_{th})$
    \State $STOP = True $
          \For{$i \in \mathcal{A}$}
          \If{$\vert\vert \mathbf{m}_{i}\vert\vert > n_{target}^{(i)}$}
          \State $\mathbf{c}_{i}= f_{lmg} (\mathbf{s}_{i},n_{target}^{(i)} )$
          \State \textbf{Update unassigned layers list:} $\mathcal{A} \leftarrow{} \mathcal{A}\setminus\left\{ i \right\}$
          \State $n_{target} \leftarrow{} n_{target} - n_{target}^{(i)}$
          \State $STOP=False$
          \EndIf
          \EndFor
          \If{$STOP==True$}
          \For{$i \in \mathcal{A}$}
          \State $\mathbf{c}_{i} \leftarrow{} \mathbf{m}_{i}$
          \EndFor
          \EndIf
    \EndWhile
     
 \end{algorithmic}
\end{algorithm}

\begin{algorithm}[t]
    \small
    \caption{Local Mask Generator (LMG)}
    \label{code:lmg}
    \begin{algorithmic}[1]
     \State \textbf{Inputs:} Saliency tensor $\mathbf{s}_{i}$, target parameters $n_{target}^{(i)}$
     \State \textbf{Outputs:} Binary mask $\mathbf{c}_{i}$
     \State $u_{th} \leftarrow f_{top-k}(\mathbf{s}_{flat}, n_{target}^{(i)})$
     \State $\mathbf{c}_{i} \leftarrow (\mathbf{s} \geq u_{th})$
     \State \Return $\mathbf{c}_{i}$
 \end{algorithmic}
\end{algorithm}

\begin{algorithm}[t]
    \small
    \caption{Global Mask Generater (GMG)}
\label{code:GlobalMask}
    \begin{algorithmic}[1]
     \State \textbf{Inputs:}  $\mathcal{S} = \{\mathbf{s}_1, \ldots, \mathbf{s}_L\}$, $\mathcal{A}$, $u_{th}$
     \State \textbf{Outputs:} $\mathcal{M} = \{\mathbf{m}_1, \ldots, \mathbf{m}_L\}$ 
     
     \For{$i = 1$ to $L$}
      \If{$i \in \mathcal{A}$}
       \State $\mathbf{m}_i \gets (\mathbf{s}_i \geq u_{th})$ \Comment{Apply threshold}
      \Else
       \State $\mathbf{m}_i \gets \mathbf{0}$ \Comment{Zero mask for already assigned layers}
      \EndIf
       \State $\mathcal{M} \leftarrow \mathcal{M} \cup \{\mathbf{m}_{i}\}$

     \EndFor
     
     \State \textbf{return} $\mathcal{M}$
     \end{algorithmic}
    \label{alg:GlobalMask}
\end{algorithm}

\subsection{Sparsity Scheduler} \label{sch}
For scheduling the sparsity, we employ the exponential decay scheduling \cite{prune_force} such that, over $T$ pruning steps, the number of non-sparse positions is gradually decreased from $\delta_{1}=1$ to $\delta_{T}=\delta$:
\begin{equation}
\delta_t = f_{sc}(t,T,\delta)=\frac{\exp\left\{\alpha(t) \log(\lfloor \delta d \rfloor ) + (1-\alpha(t)) \log(d)\right\}}{d},
\end{equation}
where $\alpha(t) = \frac{t}{T}$. Since the pruning ratio is gradually increased, we consider a similar gradual process for the layer-wise sparsity constraint such that:
\begin{equation}\label{layer-const}
\delta^{\ell}_{t} = \min(\frac{\delta_{t}}{\delta}\delta^{\ell},1),
\end{equation}
where the objective is to keep the ratio between the local and global sparsity constant during the pruning steps.
\section{Additional Numerical Results}\label{ans}

\subsection{Experiments with Various Networks and Datasets}

In this work, we consider federated training of the ResNet-20 architecture for a CIFAR-10 classification task to demonstrate our main results, since smaller NN architectures are often more vulnerable to Byzantine attacks and may not be the optimal choice for benchmarking. Another important reason for conducting experiments with the ResNet-20 architecture is to highlight the marginal importance of different network layers, such as batch normalization, convolutional layers, and fully connected layers, within the context of Byzantine attacks.

For the sake of completeness, in this section, we repeat our experiments for different NN architectures. First, we consider a smaller CNN architecture without batch normalization layers. Then we consider a NN without any convolutional layers, e.g., MLP. For these architectures, we perform training on F-MNIST and MNIST datasets, respectively.

\textbf{CNN without BatchNorm:}
In this scenario, we use the F-MNIST dataset and evaluate the performance of HSA on a smaller NN architecture. Our CNN architecture consists of 2 convolutional layers and 1 fully connected layer, with additional max pooling and dropout operations, totalling $431\times10^3$ parameters. For these experiments, HSA utilizes the $\Pi_{F}^{+}(\delta=5 \times 10^{-3}, \delta^{max}_{FC}=0.25)$ configuration, which proved effective for CIFAR-10.

In Fig. \ref{fig:fmnist_new}, we present the test accuracies for both IID (top row) and non-IID (bottom row) data distributions across various attack strategies and defence mechanisms. The first observation is that despite the absence of BatchNorm parameters, the 2-layer CNN exhibits lower robustness compared to the ResNet-20 architecture employed in Section \ref{sec:num-results}. Overall, we observe that HSA is capable of diverging the PS model in both data distributions with the exception of SignSGD. Similar to the CIFAR-10 results, we find that GAS offers relatively higher robustness, with GAS-Bulyan achieving near baseline accuracy against all attacks but HSA. Other than HSA, we observe that Min-Max attack is relatively successful especially against SignSGD. Ultimately, we find that all robust aggregators fail to defend against 3-4 attacks on non-IID simulations. 

\textbf{MLP:}
In this scenario, we use the MNIST dataset and examine the effectiveness of HSA against a compact MLP architecture that lacks both BatchNorm and convolutional layers, representing a minimalist network design. Our MLP network consists of 2 fully connected layers with $80\times10^3$ trainable parameters. Due to the MLP architecture consisting solely of fully connected layers, we employ the $\Pi_{F}^{+}(\delta=1 \times 10^{-2})$ configuration without layer-wise constraints.

In Fig. \ref{fig:mnist_new}, we highlight the IID (top row) and non-IID (bottom row) accuracies for attacks and benchmark defences. Once again, we observe that HSA is by far the most successful attack for both data distributions, with only the CC aggregator onthe  IID distribution capable of defending against it. Only Min-Max and ROP are able to marginally reduce the test accuracy against CC on IID distributions. This is mostly because ROP is designed around countering the CC aggregator \cite{rop}, and Min-Max has larger perturbation than HSA to accumulate more error to reduce the performance. Similar to other classification tasks, we see that GAS-Bulyan offers the most robustness, yet it still diverges against the proposed HSA attack in non-IID simulations, while also losing 50\% test accuracy on IID simulations. We also observe that TM is relatively successful against most attacks, with the exception of HSA and Min-Max, which are able to cause TM to diverge. Overall, we find that Min-Max and HSA are the most successful attacks where Min-Max has better performance against CM, while HSA dominates every aggregator but CC on IID.

\begin{figure*}
    \centering
    \includegraphics[width=\textwidth]{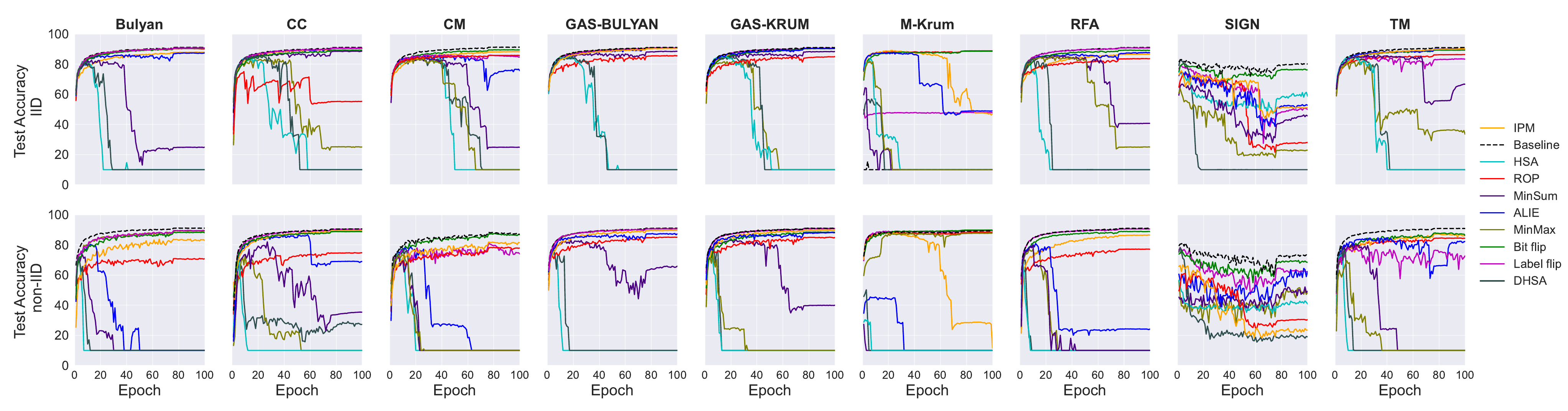}
    \caption{Test accuracy of 2-layer CNN architecture on F-MNIST dataset under IID and non-IID distributions with $k=25$ clients, of which $k_{m}=5$ are malicious. Training is conducted over 100 epochs with 9 different robust aggregation mechanisms evaluated against 8 Byzantine attack strategies. Results represent the average of 3 independent trials.}
    \label{fig:fmnist_new}
\end{figure*}

\begin{figure*}
    \centering
    \includegraphics[width=\textwidth]{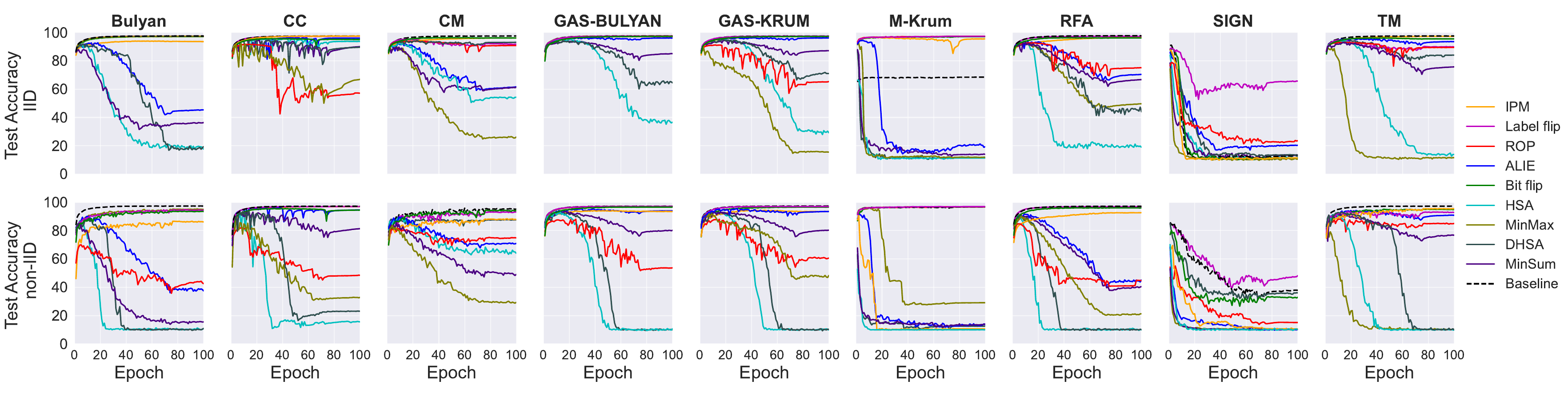}
    \caption{Test accuracy of 2-layer MLP architecture on MNIST dataset under IID and non-IID distributions with $k=25$ clients, of which $k_{m}=5$ are malicious. Training is conducted over 100 epochs with 9 different robust aggregation mechanisms evaluated against 8 Byzantine attack strategies. Results represent the average of 3 independent trials.}
    \label{fig:mnist_new}
\end{figure*}

For the F-MNIST classification objective, HSA is capable of diverging the global model on both IID and non-IID simulations except on the SignSGD aggregator, in which case it reduces the test accuracy by 35\% on IID and 55\% on non-IID simulations as illustrated in Fig \ref{fig:fmnist_new}.
For the MNIST classification task, HSA fails only with the CC aggregator in IID simulations and fails to achieve the best attack performance with the CM aggregator in non-IID simulations, while successfully diverging the PS model in all other simulation settings as illustrated in Fig. \ref{fig:mnist_new}.

\subsection{Cross-Device FL Scenario}
We use the term {\em cross-device FL} \cite{cross-device} to refer to a FL setup where 
the number of clients $k$ is large, i.e., $k>>10^3$, and clients have a limited number of training samples. In this scenario, due to communication overheads, at each communication round only a small fraction of the clients are chosen to participate the model update.

To mimic a cross-device scenario, we set $k=1000$, and allow $k_c=25$ clients to participate in the model update at each communication round, which corresponds to a $2.5\%$ participation rate. The number of malicious clients is set to $k_m=100$, such that we have $10\%$ Byzantine ratio. This ratio might be pessimistic with compared to the setup considered in \cite{B2DB}. However, it ensures a statistically significant presence of malicious clients in each round to test Byzantine robustness. This is in line with the existing literature which often tests Byzantine attacks with $20\%$ Byzantine ratio \cite{CC,rop,ndss2022_Attack}.

\begin{figure}
    \centering
    \includegraphics[width=1\linewidth]{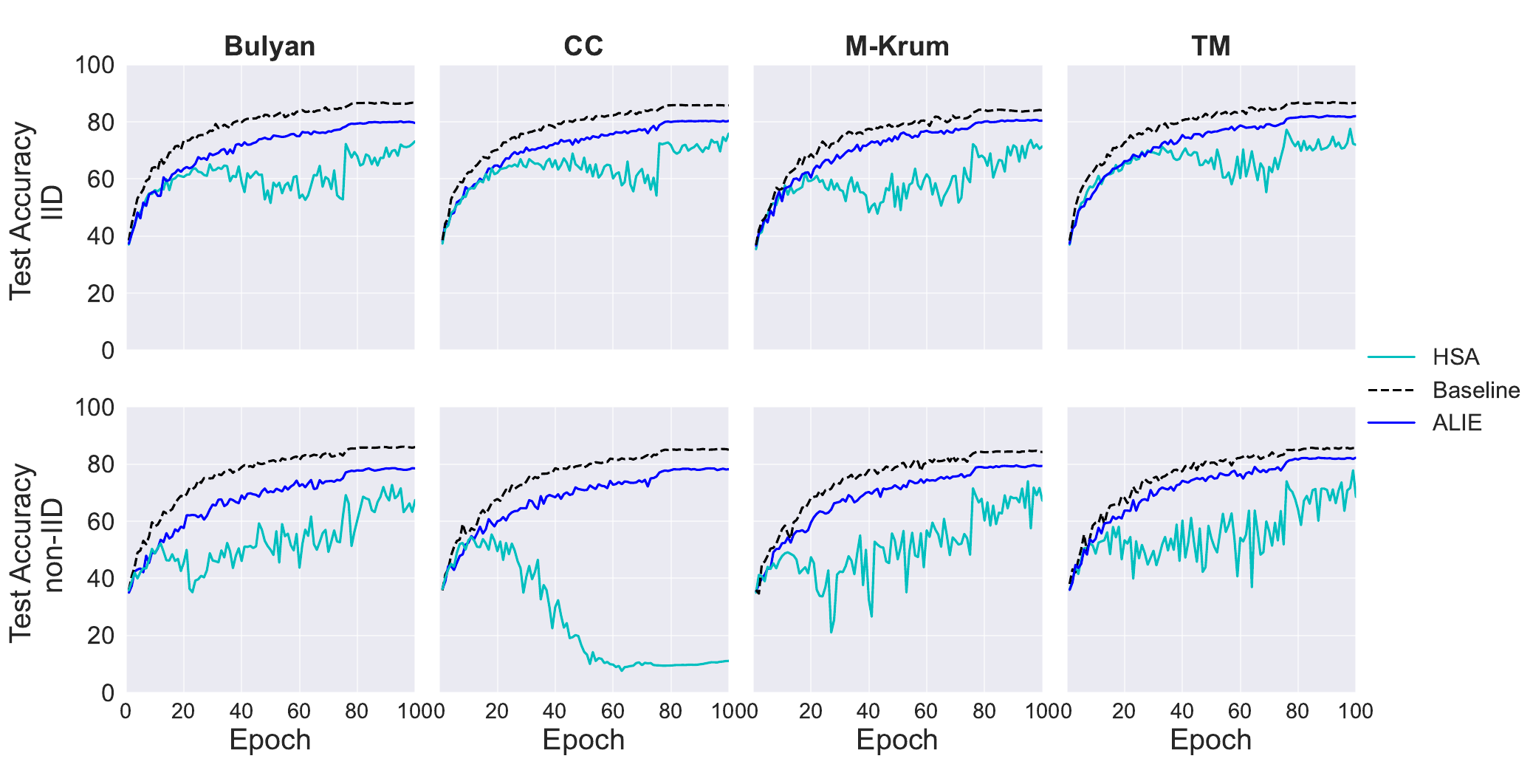}
    \caption{Test accuracy on CIFAR-10 classification task using ResNet-20 architecture in a cross-device FL scenario. The setup consists of 1000 total clients, including 100 malicious clients ($10\%$ Byzantine ratio). In each communication round, 25 clients are selected ($2.5\%$ participation rate). Results compare HSA and ALIE attacks against various robust aggregation mechanisms under both IID (top row) and non-IID (bottom row) data distributions over 100 epochs.}

    \label{fig:cross-device}
\end{figure}

Fig.~\ref{fig:cross-device} compares the performance of the proposed HSA with ALIE against a baseline in the CIFAR-10 classification task with a ResNet-20 architecture for this cross-device setting.

\subsection{Partial Knowledge on Benign Clients} \label{sec:no_known_grad}
In this section, we consider the scenario where the adversary is not able to observe the model updates of the benign clients, but estimates the benign mean ($\Bar{\mathbf{m}}_{t}$) and variance (${\boldsymbol{\sigma}}_{t}$) using the local datasets of the malicious clients. For this experiment, we use the CIFAR-10 dataset with ResNet-20 architecture and evaluate three attack strategies: ALIE, HSA, and DHSA, all utilizing the estimated statistics.

While this estimation is less precise than having direct knowledge of the benign gradients, it still enables the creation of potent attacks that can significantly degrade the global model's test accuracy. Fig.~\ref{fig:non-omni} presents our results. Our findings demonstrate that HSA achieves a significant reduction in test accuracy compared to ALIE across all aggregators and data distributions. Specifically, HSA reduces test accuracy by approximately 50\% on IID and 60\% on non-IID settings, with the only exception being Bulyan on non-IID distribution, where both attacks show similar performance. 

\begin{figure}[t!]
    \centering
    \includegraphics[width=1\linewidth]{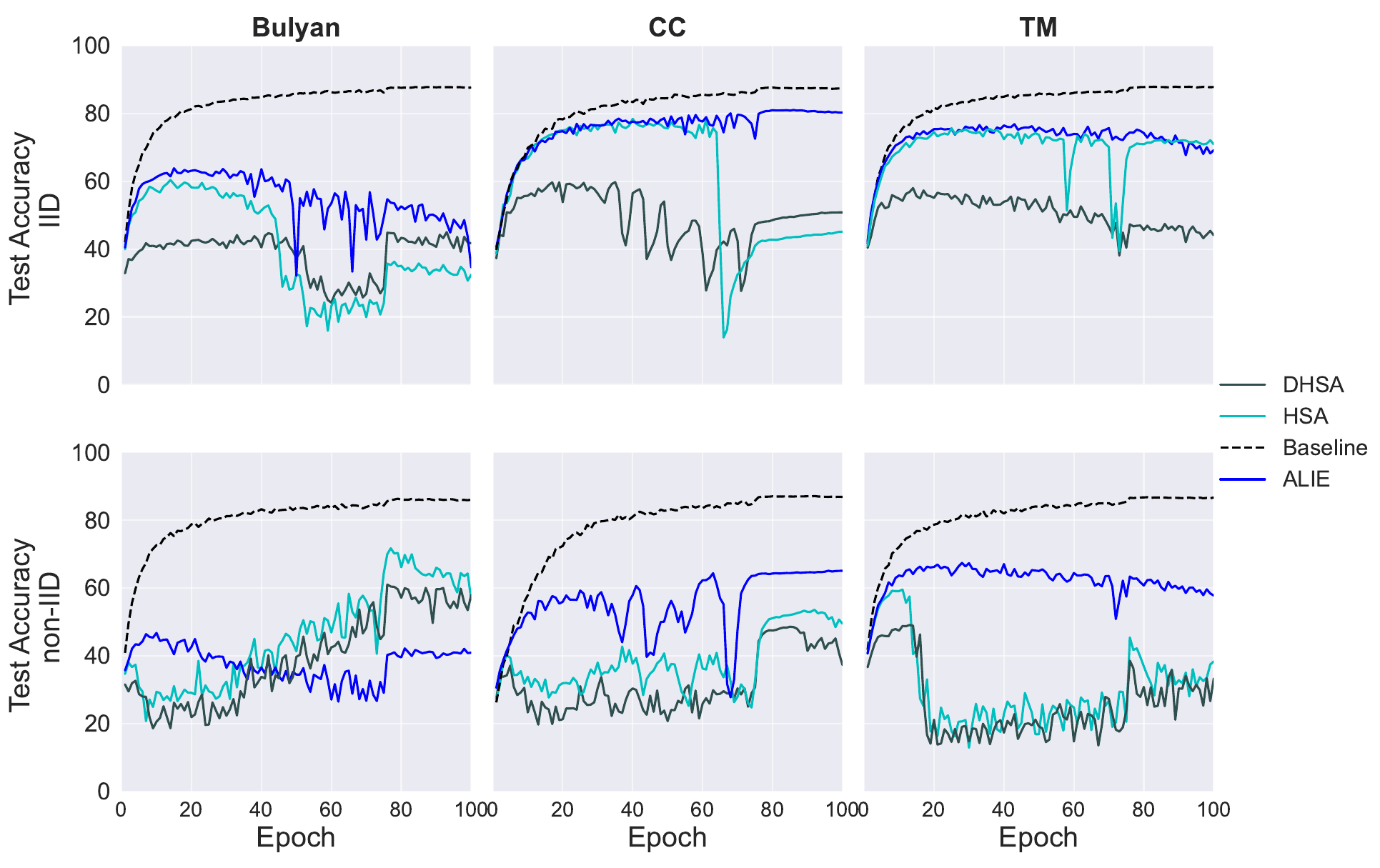}
    \caption{Test accuracy on CIFAR-10 classification task using ResNet-20 architecture when adversaries have partial knowledge of benign clients. The setup consists of 25 total clients with 5 malicious clients ($20\%$ Byzantine ratio). Attacks (ALIE, HSA, and DHSA) estimate benign mean and variance using malicious clients' local datasets rather than directly observing benign gradients. Results demonstrate attack performance against various robust aggregation mechanisms under both IID (top row) and non-IID (bottom row) data distributions over 100 epochs }
    \label{fig:non-omni}
\end{figure}

\begin{table*}[t!]
\centering
\normalsize
\caption{Accuracy of HSA under different sparsity mask formations; random sparsity mask, ERK-based random sparsity, and FORCE. In our experiments we investigate, for the IID scenario, 6 different sparsity levels; $\delta_{1}=5\times 10^{-3}$, $\delta_{2}=1\times 10^{-2}$, $\delta_{3}=5\times 10^{-2}$,$\delta_{4}=1\times 10^{-1}$, $\delta_{5}=2\times 10^{-1}$, and $\delta_{6}=5\times 10^{-1}$.
The best-performing mask formations (i.e., the one yielding the lowest test accuracy) are highlighted with \textbf{bold}, and the second-best ones are \underline{underlined}.}
\label{tab:IID-otherPrunes}
\resizebox{\textwidth}{!}{%
\begin{tabular}{@{}clllllll@{}}
\toprule
\textbf{Method} &
\textbf{Bulyan} \cite{Bulyan} &
\textbf{CC} \cite{CC} &
\textbf{CM} \cite{Trimmed_mean} &
\textbf{M-Krum} \cite{Krum} &
\textbf{RFA} \cite{RFA} &
\textbf{TM} \cite{Trimmed_mean} &
\textbf{GAS-Krum-Bulyan} \cite{GAS} \\ 
\midrule

\textbf{No ATK} & 87.6 $\pm$ 1 & 87.4 $\pm$ 0.5 & 88 $\pm$ 0.1 & 89.2 $\pm$ 0.7 & 88.2 $\pm$ 0.3 & 87.8 $\pm$ 0.4 & 87.7 $\pm$ 0.3 $-$ 87.6 $\pm$ 0.4 \\

$\Pi_{r} (\delta_1$) & 44.5 $\pm$ 17 & 79.2 $\pm$ 0.3 & \textbf{36.7 $\pm$ 1.8} & 81.28 $\pm$ 0.59 & 53 $\pm$ 3.8 & 71.7 $\pm$ 3.5 & 81.2 $\pm$ 1 $-$ 80.2 $\pm$ 0.9 \\

$\Pi_{r}^{+} (\delta_1$) & \underline{32.8 $\pm$ 4.4} & 74.7 $\pm$ 5.2 & 45.3 $\pm$ 6.3 & 81.21 $\pm$ 0.46 & \underline{45.6 $\pm$ 4.3} & \underline{56.6 $\pm$ 19} & 78.9 $\pm$ 1.2 $-$ 77.3 $\pm$ 1.8 \\

$\Pi_{ERK}^{+} (\delta_1$) & \textbf{26.7 $\pm$ 1.1} & \underline{41 $\pm$ 5.2} & 44.4 $\pm$ 14 & \underline{75.31 $\pm$ 2.29} & \textbf{45.3 $\pm$ 13} & 72.1 $\pm$ 0.6 & \underline{78 $\pm$ 0.6} $-$ \underline{73.3 $\pm$ 1.6} \\

$\Pi_{F}^{+} (\delta_1$) & 40.6 $\pm$ 1.7 & \textbf{39.6 $\pm$ 3.6} & \underline{39.5 $\pm$ 9.2} & \textbf{\textcolor{red}{15.5 $\pm$ 1.2}} & 52.8 $\pm$ 11 & 53.5 $\pm$ 17 & \textbf{74.5 $\pm$ 4} $-$ \textbf{74.8 $\pm$ 3.3} \\ 

\midrule

$\Pi_{r} (\delta_2$) & \textbf{\textcolor{red}{25.30 $\pm$ 16.52}} & 79.08 $\pm$ 0.28 & 51.64 $\pm$ 7.05 & 80.59 $\pm$ 0.52 & \textbf{\textcolor{red}{23.47 $\pm$ 9.58}} & 67.15 $\pm$ 3.19 & 79.84 $\pm$ 0.25 $-$ 79.73 $\pm$ 0.49 \\

$\Pi_{r}^{+} (\delta_2$) & \underline{40.71 $\pm$ 3.95} & 78.96 $\pm$ 1.71 & \textbf{\textcolor{red}{33.22 $\pm$ 15.04}} & 81.12 $\pm$ 0.90 & 51.35 $\pm$ 7.81 & 70.85 $\pm$ 2.96 & 79.20 $\pm$ 0.62 $-$  79.79 $\pm$ 0.37 \\

$\Pi_{ERK}^{+} (\delta_2$) & 53.42 $\pm$ 2.69 & \underline{74.83 $\pm$ 0.06} & \underline{50.12 $\pm$ 3.67} & \underline{74.46 $\pm$ 1.15} & 68.33 $\pm$ 0.29 & \underline{60.29 $\pm$ 6.07} & \underline{73.67 $\pm$ 4.48} $-$  \underline{75.42 $\pm$ 1.87}\\

$\Pi_{F}^{+} (\delta_2$) & 43.56 $\pm$ 7.99 & \textbf{69.27 $\pm$ 3.34} & 63.01 $\pm$ 1.17 & \textbf{68.52 $\pm$ 1.94} & \underline{45.64 $\pm$ 12.87} & \textbf{60.80 $\pm$ 3.32} & \textbf{75.35 $\pm$ 2.52} $-$ \textbf{75.29 $\pm$ 2.07} \\

\midrule

$\Pi_{r} (\delta_3$) & 53.1 $\pm$ 3.9 & 75.4 $\pm$ 1.9 & \underline{36.3 $\pm$ 5.6} & 78.54 $\pm$ 0.66 & \textbf{53.9 $\pm$ 5.9} & 67.9 $\pm$ 2.9 & 74.2 $\pm$ 2.6 $-$ 74.7 $\pm$ 0.7 \\

$\Pi_{r}^{+} (\delta_3$) & \textbf{32 $\pm$ 5.4} & \textbf{45.1 $\pm$ 2} & \textbf{35.9 $\pm$ 7.8} &  78.41 $\pm$ 0.22 & \underline{56.8 $\pm$ 2.8} & \textbf{59 $\pm$ 7.1} & 73.2 $\pm$ 0.6 $-$ \textbf{69.9 $\pm$ 1.9} \\

$\Pi_{ERK}^{+} (\delta_3$) & 57.4 $\pm$ 0.4 & \underline{65.2 $\pm$ 9.3} & 47.1 $\pm$ 4.8 & \textbf{68.19 $\pm$ 0.86} & 69.9 $\pm$ 0.7 & \underline{62.9 $\pm$ 11} & \textbf{70.8 $\pm$ 3.5} $-$ 73.6 $\pm$ 1\\

$\Pi_{F}^{+} (\delta_3$) & \underline{52.02 $\pm$ 1.98} & 70.40 $\pm$ 1.04 & 61.70 $\pm$ 1.60 & \underline{69.60 $\pm$ 2.42} & 65.58 $\pm$ 1.88 & 69.84 $\pm$ 1.08 & \underline{72.55 $\pm$ 1.52} $-$  \underline{73.04 $\pm$ 0.80} \\
\midrule

$\Pi_{r} (\delta_4$) & 49.96 $\pm$ 2.99 & \textbf{68.19 $\pm$ 6.79} & \textbf{39.10 $\pm$ 16.37} & 76.84 $\pm$ 0.63 & \textbf{51.49 $\pm$ 4.01} & 63.74 $\pm$ 2.96 & \underline{70.12 $\pm$ 0.96} $-$  \underline{70.60 $\pm$ 3.26} \\

$\Pi_{r}^{+} (\delta_4$) & \textbf{48.61 $\pm$ 3.03} & \underline{70.42 $\pm$ 1.93} & \underline{43.59 $\pm$ 12.36} & 76.44  $\pm$ 1.10 & \underline{57.80 $\pm$ 3.71} & 64.43 $\pm$ 5.28 & 70.95 $\pm$ 2.41 $-$  72.10 $\pm$ 1.20 \\

$\Pi_{ERK}^{+} (\delta_4$) & 58.02 $\pm$ 0.90 & 72.54 $\pm$ 0.61 & 46.53 $\pm$ 3.15 & \textbf{63.48 $\pm$ 2.51} & 67.15 $\pm$ 3.42 & \textbf{54.75 $\pm$ 12.32} & 70.29 $\pm$ 1.54 $-$  71.81 $\pm$ 1.50 \\

$\Pi_{F}^{+} (\delta_4$) & \underline{48.91 $\pm$ 0.55} & 70.50 $\pm$ 3.34 & 63.87 $\pm$ 1.14 & \underline{66.35 $\pm$ 1.49} & 62.85 $\pm$ 0.53 & \underline{61.86 $\pm$ 6.23} & \textbf{63.84 $\pm$ 9.33 $-$ 65.72 $\pm$ 2.71} \\

\midrule

$\Pi_{r} ( \delta_5$) & \underline{40.2 $\pm$ 0.5} & \textbf{41.6 $\pm$ 2.5} & \underline{49.3 $\pm$ 8.5} & 73.26$\pm$ 0.58 & \underline{59.4 $\pm$ 0.6} & 61.9 $\pm$ 2.4 & 65 $\pm$ 1.3 $-$ \underline{66.3 $\pm$ 2.2}\\

$\Pi_{r}^{+} (\delta_5$) & \textbf{26.8 $\pm$ 1.2} & \underline{43 $\pm$ 1.3} & \textbf{45.5 $\pm$ 5.2} & 73.64 $\pm$ 1.12 & \textbf{39.3 $\pm$ 10} & \textbf{46.2 $\pm$ 2.4} & \underline{61.5 $\pm$ 6} $-$ \textbf{60.7 $\pm$ 5} \\

$\Pi_{ERK}^{+} (\delta_5$) & 58.5 $\pm$ 0.9 & 50.7 $\pm$ 11 & 55.4 $\pm$ 1.7 & \textbf{65.53 $\pm$ 1.39} & 65.6 $\pm$ 1.3 & \underline{58.3 $\pm$ 4.5} & \textbf{61.1 $\pm$ 4.5} $-$ 71.5 $\pm$ 2.1 \\

$\Pi_{F}^{+} (\delta_5$) & 49.29 $\pm$ 3.89 & 69.21 $\pm$ 1.52 & 63.98 $\pm$ 1.87 & \underline{66.02 $\pm$ 1.77} & 60.51 $\pm$ 1.54 & 64.23 $\pm$ 0.69 & 67.92 $\pm$ 0.86 $-$68.64 $\pm$ 1.47 \\

\midrule

$\Pi_{r} (\delta_6$) & \underline{28.2 $\pm$ 1.9} & \underline{46.8 $\pm$ 2.8} & \textbf{42.1 $\pm$ 6.5} &  \underline{61.08 $\pm$ 1} & \underline{47.8 $\pm$ 5.9} & \underline{40 $\pm$ 5.5} & \underline{48.3 $\pm$ 0.8} $-$ 86.8 $\pm$ 0.5\\ 

$\Pi_{r}^{+} (\delta_6$) & \textbf{27.6 $\pm$ 2.6} & \textbf{\textcolor{red}{38.9 $\pm$ 0.6}} & \underline{43.1 $\pm$ 11} & 61.28 $\pm$ 1.44 &  \textbf{36.8 $\pm$ 2.9} & \textbf{\textcolor{red}{31.3 $\pm$ 9.7}} & \textbf{\textcolor{red}{39.4 $\pm$ 7.3}}$-$ \underline{86.2 $\pm$ 0.1} \\

$\Pi_{ERK}^{+} (\delta_6$) & 85.3 $\pm$ 0.7 & 52.6 $\pm$ 10 & 50.1 $\pm$ 2.7 & 79.80 $\pm$ 1.11 & 59.7 $\pm$ 1.6 & 49.2 $\pm$ 3 & 51.9 $\pm$ 2.7 $-$ 86.3 $\pm$ 0.2 \\ 

$\Pi_{F}^{+} (\delta_6$) & 44.91 $\pm$ 3.58 & 68.71 $\pm$ 0.15 & 61.05 $\pm$ 2.56 & \textbf{58.86 $\pm$ 3.36} & 63.57 $\pm$ 1.22 & 62.49 $\pm$ 2.18 & 61.18 $\pm$ 2.80 $-$ \textbf{66.52 $\pm$ 2.10}\\

\bottomrule

\end{tabular}%
}
\end{table*}

\begin{table*}[t]
\centering
\normalsize
\caption{ Accuracy of HSA under different sparsity mask formations; random sparsity mask, ERK-based random sparsity, and FORCE. In our experiments we investigate, for the non-IID scenario, 6 different sparsity levels; $\delta_{1}=5\times 10^{-3}$, $\delta_{2}=1\times 10^{-2}$, $\delta_{3}=5\times 10^{-2}$,$\delta_{4}=1\times 10^{-1}$, $\delta_{5}=2\times 10^{-1}$, and $\delta_{6}=5\times 10^{-1}$.
The best-performing mask formations (i.e., the one yielding the lowest test accuracy) are highlighted with \textbf{bold}, and the second-best ones are \underline{underlined}.}
\label{tab:nonIID}
\resizebox{\textwidth}{!}{%
\begin{tabular}{@{}clllllll@{}}
\toprule
\textbf{Method} &
\textbf{Bulyan} \cite{Bulyan} &
\textbf{CC} \cite{CC} &
\textbf{CM} \cite{Trimmed_mean} &
\textbf{M-Krum} \cite{Krum} &
\textbf{RFA} \cite{RFA} &
\textbf{TM} \cite{Trimmed_mean} &
\textbf{GAS-Krum-Bulyan} \cite{GAS} \\ 
\midrule

\textbf{No ATK} & 86 $\pm$ 0.5 & 86.9 $\pm$ 0.3 & 79.9 $\pm$ 0.4 & 88.1 $\pm$ 0.4 & 86.4 $\pm$ 0.7 & 86.7 $\pm$ 0.5 & 86.7 $\pm$ 0.5 $-$ 86.3 $\pm$ 0.5 \\

$\Pi_{r} (\delta_1$) & 24.1 $\pm$ 1.5 & 56.1 $\pm$ 1.8 & 32.8 $\pm$ 3 & 45.4 $\pm$ 6 & 28.9 $\pm$ 7.3 & 46.6 $\pm$ 1.4 & 45.3 $\pm$ 3.2 $-$ 43.9 $\pm$ 11 \\
$\Pi_{r}^{+} (\delta_1$) & \textbf{\textcolor{red}{10.1 $\pm$ 0}} & \underline{25.1 $\pm$ 1.1} & 28 $\pm$ 2.5 & 14 $\pm$ 2.1 & \underline{10 $\pm$ 0} & \underline{23.3 $\pm$ 5} &  \underline{34.7 $\pm$ 3.6} $-$ \underline{21.7 $\pm$ 4.2} \\

$\Pi_{ERK}^{+} (\delta_1$) & \underline{15.5 $\pm$ 6} & \textbf{15.8 $\pm$ 3.4} & \textbf{22.1 $\pm$ 1.4} & \textbf{10 $\pm$ 0.7} &  \textbf{\textcolor{red}{9.4 $\pm$ 0.8}} & \textbf{\textcolor{red}{10 $\pm$ 0}}& \textbf{\textcolor{red}{10 $\pm$ 0 }}$-$ \textbf{\textcolor{red}{10 $\pm$ 0}} \\
$\Pi_{F}^{+} (\delta_1$) & 26.3 $\pm$ 3.4 & 24.3 $\pm$ 3 & \underline{26.4 $\pm$ 2.3} & \underline{10.9 $\pm$ 1.3} & 39.6 $\pm$ 5.8 & \textbf{\textcolor{red}{10 $\pm$ 0}} & \textbf{\textcolor{red}{10 $\pm$ 0}} $-$ 64.3 $\pm$ 7.7 \\ 

\midrule

$\Pi_{r} (\delta_2)$ & \underline{27.44 $\pm$ 6.72} & 51.06 $\pm$ 4.18 & 33.87 $\pm$ 2.74 & 58.26 $\pm$ 5.09 & \underline{32.79 $\pm$ 2.64} & 58.78 $\pm$ 5.78 & \underline{41.44 $\pm$ 17.08} $-$ 51.65 $\pm$ 5.69 \\
$\Pi_{r}^{+} (\delta_2)$ & \textbf{20.21 $\pm$ 0.79} & 47.99 $\pm$ 2.12 & \underline{36.35 $\pm$ 3.17} & 56.03 $\pm$ 9.66 & \textbf{26.96 $\pm$ 8.38} & \underline{45.54 $\pm$ 5.64} & 49.32 $\pm$ 14.17 $-$ \underline{45.25 $\pm$ 16.81} \\

$\Pi_{ERK}^{+} (\delta_2)$ & 28.18 $\pm$ 1.39 & \textbf{31.19 $\pm$ 3.29} & \textbf{32.79 $\pm$ 1.40} & \underline{51.12 $\pm$ 2.80} & 51.89 $\pm$ 0.30 & 48.13 $\pm$ 1.48 & 56.51 $\pm$ 2.11 $-$ 62.34 $\pm$ 4.71 \\
$\Pi_{F}^{+} (\delta_2)$ & 32.08 $\pm$ 2.14 & \underline{39.56 $\pm$ 3.86} & 36.63 $\pm$ 4.21 & \textbf{37.88 $\pm$ 2.88} & 33.31 $\pm$ 13.52 & \textbf{44.24 $\pm$ 6.47} & \textbf{39.52 $\pm$ 11.47} $-$ \textbf{49.76 $\pm$ 0.80} \\
\midrule

$\Pi_{r} (\delta_3$) & \underline{22.8 $\pm$ 4} & \underline{25.7 $\pm$ 4} & \underline{25.9 $\pm$ 5.2} & \textbf{9.4 $\pm$ 1.5} & \underline{44.2 $\pm$ 7.5} & 43.2 $\pm$ 8.6 & 61.1 $\pm$ 4.6 $-$ 57.2 $\pm$ 6.9 \\
$\Pi_{r}^{+} (\delta_3$) & \underline{10.3 $\pm$ 1.3} & \textbf{16.7 $\pm$ 3.7} & \textbf{23.3 $\pm$ 3.5} & \textbf{10.3 $\pm$ 0.4} & \textbf{10 $\pm$ 0} & \textbf{27.4 $\pm$ 9.9} & \textbf{22.2 $\pm$ 9} $-$ \textbf{11.3 $\pm$ 1.8}\\
$\Pi_{ERK}^{+} (\delta_3$) & 64 $\pm$ 8.4 & 61.6 $\pm$ 2.6 & 27.9 $\pm$ 3.6 & 48.9 $\pm$ 8.5 & 54.2 $\pm$ 4.7 & 41.3 $\pm$ 3.7 & 48 $\pm$ 10.8 $-$ 71.4 $\pm$ 4.6 \\
$\Pi_{F}^{+} (\delta_3)$ & 29.12 $\pm$ 0.99 & 33.65 $\pm$ 3.19 & 35.07 $\pm$ 2.94 & 51.10 $\pm$ 1.03 & 47.97 $\pm$ 0.98 & \underline{39.61 $\pm$ 3.06} & \underline{46.25 $\pm$ 4.78} $-$ \underline{50.91 $\pm$ 4.08} \\
\midrule

$\Pi_{r} (\delta_4)$ & \underline{24.98 $\pm$ 2.46} & \underline{35.74 $\pm$ 2.37} & \textbf{29.38 $\pm$ 7.35} & \textbf{31.43 $\pm$ 6.02} & \underline{32.38 $\pm$ 4.43} & \underline{35.39 $\pm$ 9.74} & \textbf{39.99 $\pm$ 11.69} $-$ \underline{40.18 $\pm$ 10.55} \\

$\Pi_{r}^{+} (\delta_4)$ & \textbf{23.13 $\pm$ 8.90} & \textbf{31.96 $\pm$ 1.23} & 35.96 $\pm$ 0.64 & \underline{32.88 $\pm$ 6.29} & \textbf{29.64 $\pm$ 5.45} & \textbf{25.58 $\pm$ 4.84} & 48.73 $\pm$ 6.31 $-$ \textbf{29.97 $\pm$ 12.09} \\

$\Pi_{ERK}^{+} (\delta_4)$ & 73.65 $\pm$ 0.80 & 58.92 $\pm$ 1.91 & \underline{33.93 $\pm$ 1.31} & 71.22 $\pm$ 1.58 & 51.91 $\pm$ 8.77 & 51.35 $\pm$ 1.53 & 69.79 $\pm$ 2.37 $-$ 74.24 $\pm$ 0.49 \\

$\Pi_{F}^{+} (\delta_4)$ & 26.89 $\pm$ 2.31 & 30.13 $\pm$ 2.51 & 35.61 $\pm$ 2.09 & 44.30 $\pm$ 3.90 & 43.00 $\pm$ 3.54 & 37.90 $\pm$ 9.75 & \underline{44.87 $\pm$ 4.68} $-$ 46.33 $\pm$ 1.42 \\

\midrule

$\Pi_{r} ( \delta_5$) & \textbf{11.7 $\pm$ 1.4} & \underline{19.6 $\pm$ 2.4} & \underline{27.3 $\pm$ 1.1} & \textbf{10 $\pm$ 0.1} & 39.8 $\pm$ 6.3 & \underline{23.9 $\pm$ 8.6} & 55.6 $\pm$ 8.8 $-$ \underline{23.4 $\pm$ 8} \\

$\Pi_{r}^{+} (\delta_5$) & \underline{18 $\pm$ 6.5} & \textbf{18.3 $\pm$ 1.3} & \textbf{27.3 $\pm$ 1} & \underline{11.1 $\pm$ 1.6} & \textbf{ 10 $\pm$ 0} & \textbf{\textcolor{red}{10 $\pm$ 0}} & \underline{41.3 $\pm$ 6.2} $-$ \textbf{15.6 $\pm$ 8.5} \\

$\Pi_{ERK}^{+} (\delta_5$) & 70 $\pm$ 3.2 & 52.7 $\pm$ 4.5 & 27.8 $\pm$ 2.1 & 46.8 $\pm$ 9.6 & 49.9 $\pm$ 4.4 & 26.2 $\pm$ 2.8 & \textbf{32.5 $\pm$ 2.2} $-$ 71 $\pm$ 2.3 \\

$\Pi_{F}^{+} (\delta_5)$ & 27.29 $\pm$ 0.41 & 33.97 $\pm$ 2.92 & 29.79 $\pm$ 4.82 & 39.99 $\pm$ 2.94 & \underline{37.66 $\pm$ 4.90} & 31.68 $\pm$ 2.81 & 44.42 $\pm$ 3.72 $-$ 39.14 $\pm$ 2.90 \\

\midrule

$\Pi_{r} (\delta_6$) & \underline{25.5 $\pm$ 4.2} & \textbf{\textcolor{red}{15.4 $\pm$ 3.5}} & \textbf{\textcolor{red}{20.9 $\pm$ 2.9}} & \underline{9.9 $\pm$ 0.6} & 28.9 $\pm$ 8.9 & \textbf{\textcolor{red}{10 $\pm$ 0}} & 57.3 $\pm$ 5 $-$ \underline{53.3 $\pm$ 8.7} \\

$\Pi_{r}^{+} (\delta_6$) & \textbf{16.4 $\pm$ 2.4} & \underline{16.6 $\pm$ 3.2} & \underline{23.2 $\pm$ 1.1} & \textbf{\textcolor{red}{9.7 $\pm$ 0.3}} & \textbf{14.5 $\pm$ 3.3} & \textbf{\textcolor{red}{10 $\pm$ 0}} & \textbf{19.2 $\pm$ 3.4} $-$ 60.4 $\pm$ 5 \\ 

$\Pi_{ERK}^{+} (\delta_6$) & 78.7 $\pm$ 3.5 & 41.6 $\pm$ 1.3 & 23.2 $\pm$ 2.4 & 47.9 $\pm$ 2.2 & \underline{26.6 $\pm$ 6.5} &  \underline{20.7 $\pm$ 1.5} & \underline{22.3 $\pm$ 3.5} $-$ 72.3 $\pm$ 19 \\ 

$\Pi_{F}^{+} (\delta_6)$ & 26.17 $\pm$ 1.26 & 26.39 $\pm$ 1.63 & 29.63 $\pm$ 1.56 & 26.93 $\pm$ 3.95 & 30.20 $\pm$ 7.50 & 31.18 $\pm$ 1.16 & 31.57 $\pm$ 2.54  $-$\textbf{ 23.30 $\pm$ 6.41 }\\ 
\bottomrule

\end{tabular}%
}
\end{table*}

\begin{figure*}[t!]
    \centering
    \includegraphics[width=\textwidth]{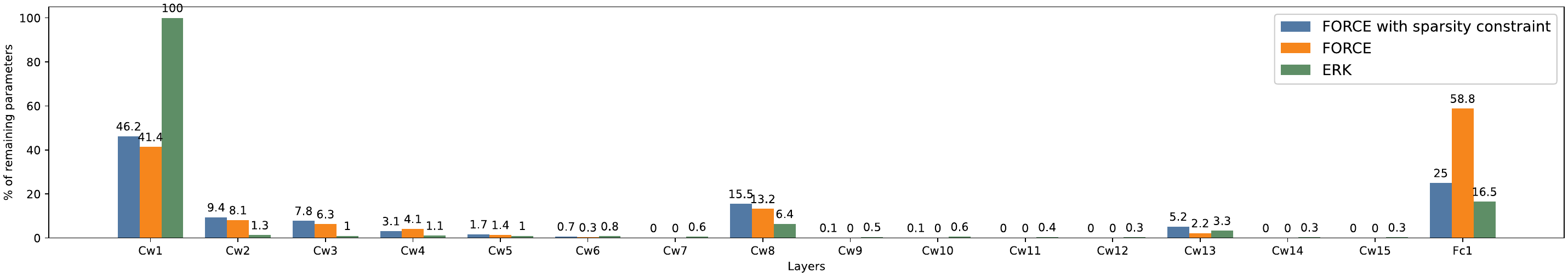}
    \caption{Distribution of the non-sparse locations (remaining weights) in ResNet-20 architecture after pruning with ERK, Vanilla FORCE, and FORCE with a sparsity constraint on the fully-connected penultimate layer, $\delta^{max}_{FC}=0.25$, for $\delta=0.005$. The x-axis denotes the NN layers, where  Cw$i$ denotes the weights of the $i$th convolutional layer, and Fc1 denotes the weights of the FC layer at the end.}
    \label{fig:layers_prune_comperison}
\end{figure*}

\section{Ablations on Network Sparsity Formation} \label{sec:ablation_on_sparsity}
In this section, we conduct comprehensive ablation studies investigating the impact of sparsity patterns employed in HSA on attack performance. Specifically, our objective is to address the following research questions through extensive experiments:

\begin{enumerate}
    \item Is it necessary to consider the underlying network topology, or can randomly generated sparse masks yield similar attack effectiveness?
    \item How critical is the layer-wise distribution of non-zero positions for evading robust aggregation methods?
    \item What is the impact of sparsity ratio on HSA performance in both IID and non-IID scenarios?   
\end{enumerate}

\subsection{Structure of the Sparsity Mask} \label{sec:sparsity_structure}

\begin{table*}[h!]
\caption{Impact of layer-wise sparsity constraints on HSA performance in the IID setting. Masks are generated using the FORCE algorithm with varying global sparsity ratios $\delta \in \{5\times10^{-3}, 1\times10^{-2}, 2.5\times10^{-2}, 5\times10^{-2}, 1\times10^{-1}\}$ and maximum layer constraints $\delta^{max} \in \{0.05, 0.1, 0.25, 1\}$. Best attack performances (lowest test accuracy) are highlighted in \textbf{bold}, second-best values are \underline{underlined}.}
\label{tab:ForceAblationsIID}
\begin{tabular}{@{}clllllll@{}}
\toprule
\textbf{Sparsity Pattern$\delta / \delta^{max}$} & 
\textbf{Bulyan} \cite{Bulyan} &
\textbf{CC} \cite{CC} &
\textbf{CM} \cite{Trimmed_mean} &
\textbf{M-Krum} \cite{Krum} &
\textbf{RFA} \cite{RFA} &
\textbf{TM} \cite{Trimmed_mean} &
\textbf{GAS-Krum-Bulyan} \cite{GAS} \\ \midrule

$5 \times 10^{-3}$ / $1$ & 40.6 $\pm$ 1.7 & 39.6 $\pm$ 3.6 & 39.5 $\pm$ 9.2 & \textbf{15.5 $\pm$ 1.2} & 52.8 $\pm$ 11 & \textbf{53.5 $\pm$ 17} & 74.5 $\pm$ 4 - 74.8 $\pm$ 3.3 \\

$5 \times 10^{-3}$ / $0.05$ & 41.9 $\pm$ 6 & 76.4 $\pm$ 1.4 & \underline{38.2 $\pm$ 8.5} & 70.8 $\pm$ 0.8 & 53.4 $\pm$ 7.2 & 66.7 $\pm$ 2.5 & 73.6 $\pm$ 2.7 - 73.9 $\pm$ 2 \\

$5 \times 10^{-3}$ / $0.1$ & 39.2 $\pm$ 4.3 & 76.2 $\pm$ 0.5 & 50.1 $\pm$ 4.6 & 67.2 $\pm$ 3.9 & 56.1 $\pm$ 1.1 & 66.3 $\pm$ 4.7 & 75.7 $\pm$ 1.7 - 75 $\pm$ 0.4 \\

$5 \times 10^{-3}$ / $0.25$ & 31.3 $\pm$ 2.7 & \textbf{33.9 $\pm$ 2.9} & \textbf{35.7 $\pm$ 4.5} & \underline{16.3 $\pm$ 2.1} & \underline{42.7 $\pm$ 10} & 63.4 $\pm$ 3.3 & \underline{59.5 $\pm$ 18} - \textbf{52.5 $\pm$ 8.3} \\

$1 \times 10^{-2}$ / $0.05$ & 46.1 $\pm$ 3.6 & 75.9 $\pm$ 0.6 & 55.4 $\pm$ 0.4 & 68.7 $\pm$ 4.2 & 50 $\pm$ 4 & 69.4 $\pm$ 1.5 & 75.8 $\pm$ 0.9 - 74.8 $\pm$ 2.1 \\

$1 \times 10^{-2}$ / $0.1$ & 46.5 $\pm$ 2.7 & 75.9 $\pm$ 0.4 & 52.7 $\pm$ 2.5 & 71.8 $\pm$ 2.3 & 55.2 $\pm$ 4.1 & 66.3 $\pm$ 2.5 & 69.4 $\pm$ 4.6 - 74.8 $\pm$ 0.9 \\ 

$1 \times 10^{-2}$ / $0.25$ & 28.4 $\pm$ 1.9 & \underline{38.7 $\pm$ 1} & 42.3 $\pm$ 18 & 16.5 $\pm$ 1.5 & \textbf{39.9 $\pm$ 9.4} & \underline{47.5 $\pm$ 6.7} & 67.8 $\pm$ 8.2 $-$ \underline{ 59.3 $\pm$ 9.5 } \\ 

$2.5 \times 10^{-2}$ / $1$ & 44.2 $\pm$ 4.2 & 43.4 $\pm$ 2.2 & 47.2 $\pm$ 3.3 & 73.9 $\pm$ 0.9 & 66.9 $\pm$ 1 & 67.6 $\pm$ 1.8 & 73.4 $\pm$ 1 - 70.3 $\pm$ 1.5 \\

$2.5 \times 10^{-2}$ / $0.25$ & 34.1 $\pm$ 3.5 & 39.7 $\pm$ 4.9 & 42 $\pm$ 3.8 & 70.6 $\pm$ 1.8 & 53.5 $\pm$ 6.3 & 67.6 $\pm$ 4 & 74 $\pm$ 1.8 - 69.5 $\pm$ 1.8 \\

$5 \times 10^{-2}$ / $1$ & 34.9 $\pm$ 0.8 & 52.2 $\pm$ 4 & 47 $\pm$ 2.9 & 67.5 $\pm$ 0.9 & 60.1 $\pm$ 3.2 & 54.3 $\pm$ 8.3 & 73.2 $\pm$ 0.7 - 72 $\pm$ 0.7 \\

$5 \times 10^{-2}$ / $0.25$ & \underline{26.7 $\pm$ 2.3} & 43.3 $\pm$ 5.2 & 51.9 $\pm$ 3.9 & 64.6 $\pm$ 8.6 & 48.8 $\pm$ 0.6 & 54 $\pm$ 9.5 & 71.1 $\pm$ 2.1 - 70.6 $\pm$ 1.6 \\

$1 \times 10^{-1}$ / $1$ & 55.4 $\pm$ 1 & 52.2 $\pm$ 5 & 53 $\pm$ 1.6 & 63.9 $\pm$ 2.2 & 58.4 $\pm$ 2.8 & 58.2 $\pm$ 4.5 & 68.7 $\pm$ 1.3 - 69.1 $\pm$ 2.6 \\

$1 \times 10^{-1}$ / $0.25$ & \textbf{25.8 $\pm$ 2.9} & 44.8 $\pm$ 5 & 49.4 $\pm$ 0.7 & 53.2 $\pm$ 7.5 & 47.7 $\pm$ 2.5 & \textbf{47.1 $\pm$ 9.5} & \textbf{59 $\pm$ 13.1} - 65 $\pm$ 2.5 \\ 

\bottomrule
\end{tabular}
\end{table*}

\begin{table*}[h!]
\caption{Impact of layer-wise sparsity constraints on HSA performance in non-IID setting. Masks are generated using the FORCE algorithm with varying global sparsity ratios $\delta \in \{5\times10^{-3}, 1\times10^{-2}, 2.5\times10^{-2}, 5\times10^{-2}, 1\times10^{-1}\}$ and maximum layer constraints $\delta^{max} \in \{0.05, 0.1, 0.25, 1\}$. Best attack performances (lowest test accuracy) are highlighted in \textbf{bold}, second-best values are \underline{underlined}.}
\label{tab:forceAblationsDir}
\begin{tabular}{@{}clllllll@{}}
\toprule
\textbf{Sparsity Pattern$\delta / \delta^{max}$} & 
\textbf{Bulyan} \cite{Bulyan} &
\textbf{CC} \cite{CC} &
\textbf{CM} \cite{Trimmed_mean} &
\textbf{M-Krum} \cite{Krum} &
\textbf{RFA} \cite{RFA} &
\textbf{TM} \cite{Trimmed_mean} &
\textbf{GAS-Krum-Bulyan} \cite{GAS} \\ \midrule
$5 \times 10^{-3}$ / $0.05$ & 25.6 $\pm$ 3.6 & 31.4 $\pm$ 3.8 & 37.8 $\pm$ 5.8 & 45.2 $\pm$ 3.9 & 33.8 $\pm$ 7.8 & 25.6 $\pm$ 4.6 & 33.2 $\pm$ 8.2 - 29.3 $\pm$ 2.7 \\

$5 \times 10^{-3}$ / $0.1$ & 26.5 $\pm$ 3.3 & 38.7 $\pm$ 4.3 & 35 $\pm$ 5.7 & 43.9 $\pm$ 8 & 36.4 $\pm$ 6.2 & 24.3 $\pm$ 5.6 & 36.5 $\pm$ 10.6 - 39.2 $\pm$ 6.8 \\

$1 \times 10^{-2}$ / $0.05$ & 21.8 $\pm$ 6.5 & 32.4 $\pm$ 2.7 & 38.1 $\pm$ 4.2 & \textbf{21.4 $\pm$ 7.2} & 29.5 $\pm$ 10.9 & \underline{23.3 $\pm$ 6.4} & 26.1 $\pm$ 9.6 - 30.2 $\pm$ 14.1 \\

$1 \times 10^{-2}$ / $0.1$ & 22.5 $\pm$ 5.1 & 31 $\pm$ 2.1 & 33.7 $\pm$ 3.5 & 34 $\pm$ 6.7 & 33 $\pm$ 2.7 & 32 $\pm$ 7.7 & 26 $\pm$ 3.2 - 36.8 $\pm$ 3.6 \\ 

$2.5\times 10^{-2}$ / $1$ & 32.4 $\pm$ 4.4 & 41.2 $\pm$ 5.6 & 27 $\pm$ 6.2 & 50.6 $\pm$ 3.4 & 54.6 $\pm$ 1.8 & 42.4 $\pm$ 5 & 42.6 $\pm$ 11.2 - 69.9 $\pm$ 2.7 \\

$2.5 \times 10^{-2}$ / $0.25$ & 30.2 $\pm$ 1.9 & 39 $\pm$ 3.7 & 32.9 $\pm$ 5.3 & 39.4 $\pm$ 15.1 & 52.4 $\pm$ 1.5 & 41.7 $\pm$ 2.9 & 52.3 $\pm$ 1 - 58 $\pm$ 7 \\

$5\times 10^{-2}$ / $1$& 34.2 $\pm$ 25 & 28 $\pm$ 2.2 & 23.6 $\pm$ 2  & 43 $\pm$ 2.5 & 45.2 $\pm$ 1.6 & \textbf{10 $\pm$ 0} & 14.6 $\pm$ 3.3 - 65 $\pm$ 4 \\

$5 \times 10^{-2}$ / $0.25$ & \underline{14 $\pm$ 5.7} & \underline{20.8 $\pm$ 1.2} & \underline{21.6 $\pm$ 2.3} & \underline{21.6 $\pm$ 2.3} & \textbf{10 $\pm$ 0} & \textbf{10 $\pm$ 0} & 10 $\pm$ 0 - \underline{12.7 $\pm$ 3.7} \\

$1 \times 10^{-1}$ / $1$& 60.9 $\pm$ 9.4 & 28.1 $\pm$ 5.3 & 22.4 $\pm$ 3.7 & 50.2 $\pm$ 8.7 & 40.7 $\pm$ 3.3 & \textbf{10 $\pm$ 0} & \textbf{9.7 $\pm$ 2.2} - 68.3 $\pm$ 2.7 \\

$1 \times 10^{-1}$ / $0.25$ & \textbf{10.4 $\pm$ 5.2} & \textbf{19.7 $\pm$ 1.8} & \textbf{19.4 $\pm$ 4.4} & 28 $\pm$ 5.1 & \underline{11.5 $\pm$ 1.1} & \textbf{10 $\pm$ 0} & \underline{9.7 $\pm$ 0.4} - \textbf{10 $\pm$ 0} \\ \bottomrule

\end{tabular}
\end{table*}

\begin{table*}[t!]
\caption{Impact of layer-wise sparsity constraints on HSA performance in non-IID setting. $\delta$ denotes global sparsity, $\delta^{max}_{cw1}$ denotes maximum sparsity for the first Conv layer, and $\delta^{max}_{fc}$ denotes maximum sparsity for the last FC layer. Constraint value of 1 indicates no constraint. Best attack performances (lowest test accuracy) are highlighted in \textbf{bold}, second-best values are \underline{underlined}.}
\label{tab:LayerConstraintsDIR}
\begin{tabular}{@{}clllllll@{}}
\toprule
\textbf{Sparsity Pattern $\delta / \delta^{max}_{cw1} / \delta^{max}_{fc}$} & \textbf{Bulyan} & \textbf{CC} & \textbf{M-Krum} & \textbf{RFA} & \textbf{TM} & \textbf{GAS-Bulyan} & \textbf{GAS-Krum} \\ \midrule
$5 \times 10^{-3}$ / $0.2$ / $0.2$ & 18.2 $\pm$ 4.2 & 26.9 $\pm$ 2.2 & 38.8 $\pm$ 10.2 & 43.9 $\pm$ 7.6 & \textbf{10.0 $\pm$ 0.0} & 35.9 $\pm$ 22.8 & \textbf{\textcolor{red}{10.0 $\pm$ 0.0}} \\
$5 \times 10^{-3}$ / $0.4$ / $0.2$ & \textbf{13.6 $\pm$ 3.8} & 22.9 $\pm$ 4.0 & 36.9 $\pm$ 5.8 & 35.9 $\pm$ 1.2 & 14.8 $\pm$ 5.9 & 33.4 $\pm$ 14.5 & 13.0 $\pm$ 2.3 \\
$5 \times 10^{-3}$ / $1$ / $0.2$ & \underline{14.7 $\pm$ 5.7} & 27.1 $\pm$ 2.0 & 37.5 $\pm$ 5.0 & 32.4 $\pm$ 15.6 & \underline{10.0 $\pm$ 0.0} & \textbf{23.0 $\pm$ 16.2} & \underline{10.0 $\pm$ 0.0} \\
$5 \times 10^{-3}$ / $0.2$ / $0.4$ & 16.3 $\pm$ 1.0 & 22.2 $\pm$ 2.9 & \underline{33.2 $\pm$ 12.7} & 27.6 $\pm$ 7.2 & 10.0 $\pm$ 0.0 & 41.1 $\pm$ 19.4 & 11.2 $\pm$ 1.2 \\
$5 \times 10^{-3}$ / $0.4$ / $0.4$ & 26.1 $\pm$ 6.0 & \textbf{20.3 $\pm$ 3.3} & 40.6 $\pm$ 7.3 & 40.4 $\pm$ 8.4 & 10.0 $\pm$ 0.0 & 29.2 $\pm$ 7.1 & 10.0 $\pm$ 0.0 \\
$5 \times 10^{-3}$ / $1$ / $0.4$ & 20.4 $\pm$ 10.8 & \underline{20.4 $\pm$ 2.7} & \textbf{28.3 $\pm$ 16.3} & 26.5 $\pm$ 12.8 & 10.6 $\pm$ 0.8 & 37.1 $\pm$ 19.2 & 12.3 $\pm$ 3.2 \\
$5 \times 10^{-3}$ / $0.2$ / $1$ & 31.3 $\pm$ 23.5 & 29.5 $\pm$ 6.4 & 41.2 $\pm$ 12.4 & \textbf{15.8 $\pm$ 7.5} & 22.9 $\pm$ 13.6 & 35.6 $\pm$ 20.0 & 10.0 $\pm$ 0.0 \\
$5 \times 10^{-3}$ / $0.4$ / $1$ & 23.0 $\pm$ 1.2 & 23.7 $\pm$ 2.2 & 39.9 $\pm$ 3.7 & \underline{20.4 $\pm$ 14.2} & 11.7 $\pm$ 1.3 & \underline{28.4 $\pm$ 18.8} & 12.2 $\pm$ 3.1 \\
$5 \times 10^{-3}$ / $1$ / $1$ & 23.2 $\pm$ 7.3 & 24.5 $\pm$ 1.7 & 39.6 $\pm$ 13.5 & 23.5 $\pm$ 13.9 & 10.0 $\pm$ 0.0 & 34.4 $\pm$ 12.5 & 10.0 $\pm$ 0.0 \\
\midrule
$1 \times 10^{-2}$ / $0.2$ / $0.2$ & 16.7 $\pm$ 7.1 & 24.0 $\pm$ 0.9 & 36.5 $\pm$ 6.5 & \textbf{12.5 $\pm$ 2.4} & \textbf{10.0 $\pm$ 0.0} & \textbf{14.1 $\pm$ 4.1} & \textbf{10.0 $\pm$ 0.0} \\
$1 \times 10^{-2}$ / $0.4$ / $0.2$ & 21.1 $\pm$ 10.2 & 23.1 $\pm$ 1.5 & \underline{26.2 $\pm$ 9.1} & 37.9 $\pm$ 19.7 & \underline{10.0 $\pm$ 0.0} & 27.0 $\pm$ 21.2 & \underline{10.0 $\pm$ 0.0} \\
$1 \times 10^{-2}$ / $1$ / $0.2$ & \underline{16.5 $\pm$ 6.0} & \underline{23.0 $\pm$ 1.8} & 35.6 $\pm$ 13.9 & \underline{20.1 $\pm$ 7.2} & 14.8 $\pm$ 6.8 & \underline{15.6 $\pm$ 0.5} & 10.0 $\pm$ 0.0 \\
$1 \times 10^{-2}$ / $0.2$ / $0.4$ & 28.0 $\pm$ 15.3 & \textbf{22.8 $\pm$ 2.0} & 45.3 $\pm$ 12.4 & 45.4 $\pm$ 5.5 & 10.0 $\pm$ 0.0 & 57.8 $\pm$ 11.5 & 10.0 $\pm$ 0.0 \\
$1 \times 10^{-2}$ / $0.4$ / $0.4$ & 25.2 $\pm$ 9.9 & 24.1 $\pm$ 1.9 & 40.4 $\pm$ 11.0 & 40.5 $\pm$ 13.0 & 11.8 $\pm$ 2.5 & 39.4 $\pm$ 23.9 & 10.0 $\pm$ 0.0 \\
$1 \times 10^{-2}$ / $1$ / $0.4$ & 25.9 $\pm$ 10.2 & 23.2 $\pm$ 2.6 & \textbf{19.6 $\pm$ 1.3} & 26.0 $\pm$ 16.0 & 10.0 $\pm$ 0.0 & 44.9 $\pm$ 4.5 & 10.0 $\pm$ 0.0 \\
$1 \times 10^{-2}$ / $0.2$ / $1$ & 24.7 $\pm$ 9.1 & 29.2 $\pm$ 2.6 & 41.8 $\pm$ 7.3 & 48.2 $\pm$ 4.6 & 10.0 $\pm$ 0.0 & 53.9 $\pm$ 12.7 & 10.0 $\pm$ 0.0 \\
$1 \times 10^{-2}$ / $0.4$ / $1$ & \textbf{12.7 $\pm$ 8.6} & 24.7 $\pm$ 1.3 & 34.8 $\pm$ 14.4 & 25.4 $\pm$ 14.0 & 11.0 $\pm$ 1.4 & 37.2 $\pm$ 17.2 & 10.0 $\pm$ 0.0 \\
$1 \times 10^{-2}$ / $1$ / $1$ & 24.8 $\pm$ 10.9 & 24.0 $\pm$ 0.8 & 49.1 $\pm$ 6.3 & 39.1 $\pm$ 13.5 & 11.8 $\pm$ 2.6 & 39.1 $\pm$ 17.1 & 10.0 $\pm$ 0.0 \\
\midrule
$5 \times 10^{-2}$ / $0.2$ / $0.2$ & \textbf{\textcolor{red}{10.8 $\pm$ 2.2}} & \underline{19.8 $\pm$ 3.9} & 32.7 $\pm$ 4.3 & \textbf{10.0 $\pm$ 0.0} & \underline{10.0 $\pm$ 0.0} & \textbf{\textcolor{red}{9.9 $\pm$ 2.4}} & \textbf{10.0 $\pm$ 0.0} \\
$5 \times 10^{-2}$ / $0.4$ / $0.2$ & 18.6 $\pm$ 4.5 & 23.1 $\pm$ 1.6 & \underline{21.6 $\pm$ 8.0} & \underline{11.5 $\pm$ 1.7} & 13.4 $\pm$ 1.6 & \underline{10.0 $\pm$ 0.0} & \underline{10.0 $\pm$ 0.0} \\
$5 \times 10^{-2}$ / $1$ / $0.2$ & \underline{10.8 $\pm$ 1.3} & \textbf{\textcolor{red}{16.5 $\pm$ 2.4}} & 35.1 $\pm$ 11.1 & 12.0 $\pm$ 2.8 & \textbf{\textcolor{red}{9.9 $\pm$ 0.1}} & 10.3 $\pm$ 0.3 & 10.0 $\pm$ 0.0 \\
$5 \times 10^{-2}$ / $0.2$ / $0.4$ & 16.8 $\pm$ 7.8 & 23.2 $\pm$ 1.2 & 29.8 $\pm$ 5.3 & 28.9 $\pm$ 14.6 & 12.1 $\pm$ 3.0 & 41.0 $\pm$ 5.6 & 10.0 $\pm$ 0.0 \\
$5 \times 10^{-2}$ / $0.4$ / $0.4$ & 17.4 $\pm$ 3.8 & 22.5 $\pm$ 3.2 & 25.1 $\pm$ 6.1 & 34.8 $\pm$ 14.6 & 10.0 $\pm$ 0.0 & 43.0 $\pm$ 1.9 & 10.0 $\pm$ 0.0 \\
$5 \times 10^{-2}$ / $1$ / $0.4$ & 19.0 $\pm$ 4.9 & 22.8 $\pm$ 2.4 & \textbf{\textcolor{red}{17.6 $\pm$ 10.6}} & 12.0 $\pm$ 2.6 & 10.0 $\pm$ 0.0 & 43.6 $\pm$ 23.4 & 10.0 $\pm$ 0.0 \\
$5 \times 10^{-2}$ / $0.2$ / $1$ & 40.0 $\pm$ 10.8 & 30.1 $\pm$ 1.4 & 39.2 $\pm$ 6.8 & 50.4 $\pm$ 3.9 & 10.0 $\pm$ 0.0 & 67.3 $\pm$ 7.8 & 10.0 $\pm$ 0.0 \\
$5 \times 10^{-2}$ / $0.4$ / $1$ & 40.4 $\pm$ 9.0 & 30.1 $\pm$ 2.8 & 54.8 $\pm$ 5.2 & 38.3 $\pm$ 15.6 & 10.0 $\pm$ 0.0 & 64.2 $\pm$ 6.0 & 10.0 $\pm$ 0.0 \\
$5 \times 10^{-2}$ / $1$ / $1$ & 50.0 $\pm$ 20.9 & 26.4 $\pm$ 7.7 & 39.4 $\pm$ 15.5 & 33.4 $\pm$ 16.8 & 10.0 $\pm$ 0.0 & 71.0 $\pm$ 8.0 & 10.0 $\pm$ 0.0 \\
\midrule
$1 \times 10^{-1}$ / $0.2$ / $0.2$ & \underline{13.4 $\pm$ 3.5} & \underline{19.7 $\pm$ 2.2} & 24.6 $\pm$ 2.0 & \textbf{\textcolor{red}{9.9 $\pm$ 0.1}} & 14.8 $\pm$ 6.7 & \textbf{10.0 $\pm$ 0.0} & \textbf{10.0 $\pm$ 0.0} \\
$1 \times 10^{-1}$ / $0.4$ / $0.2$ & 19.3 $\pm$ 10.4 & 22.5 $\pm$ 3.4 & 24.0 $\pm$ 8.3 & 11.5 $\pm$ 2.1 & 12.2 $\pm$ 3.1 & 10.9 $\pm$ 1.3 & \underline{10.0 $\pm$ 0.0} \\
$1 \times 10^{-1}$ / $1$ / $0.2$ & \textbf{11.5 $\pm$ 1.8} & \textbf{18.5 $\pm$ 2.1} & \textbf{21.4 $\pm$ 5.1} & \underline{10.0 $\pm$ 0.0} & \textbf{10.0 $\pm$ 0.0} & \underline{10.2 $\pm$ 0.3} & 10.0 $\pm$ 0.0 \\
$1 \times 10^{-1}$ / $0.2$ / $0.4$ & 24.6 $\pm$ 11.6 & 21.1 $\pm$ 2.0 & \underline{23.4 $\pm$ 7.8} & 12.5 $\pm$ 3.6 & \underline{10.0 $\pm$ 0.0} & 34.9 $\pm$ 2.7 & 10.0 $\pm$ 0.0 \\
$1 \times 10^{-1}$ / $0.4$ / $0.4$ & 16.5 $\pm$ 6.0 & 23.5 $\pm$ 1.7 & 26.7 $\pm$ 7.7 & 10.0 $\pm$ 0.0 & 10.0 $\pm$ 0.0 & 20.9 $\pm$ 14.8 & 10.0 $\pm$ 0.0 \\
$1 \times 10^{-1}$ / $1$ / $0.4$ & 18.9 $\pm$ 3.3 & 20.0 $\pm$ 2.1 & 26.2 $\pm$ 4.6 & 17.3 $\pm$ 5.8 & 10.0 $\pm$ 0.0 & 39.5 $\pm$ 25.1 & 10.0 $\pm$ 0.0 \\
$1 \times 10^{-1}$ / $0.2$ / $1$ & 65.9 $\pm$ 11.8 & 28.5 $\pm$ 1.1 & 39.4 $\pm$ 5.9 & 33.6 $\pm$ 9.2 & 10.0 $\pm$ 0.0 & 70.2 $\pm$ 7.3 & 15.2 $\pm$ 7.3 \\
$1 \times 10^{-1}$ / $0.4$ / $1$ & 54.7 $\pm$ 18.9 & 25.7 $\pm$ 1.1 & 46.3 $\pm$ 14.1 & 37.2 $\pm$ 9.3 & 15.6 $\pm$ 7.9 & 62.6 $\pm$ 10.3 & 22.5 $\pm$ 13.8 \\
$1 \times 10^{-1}$ / $1$ / $1$ & 56.5 $\pm$ 26.9 & 27.6 $\pm$ 3.9 & 51.3 $\pm$ 8.7 & 42.1 $\pm$ 7.0 & 10.0 $\pm$ 0.0 & 71.6 $\pm$ 3.1 & 10.0 $\pm$ 0.0 \\
\bottomrule
\end{tabular}
\end{table*}

\begin{table*}[]
\caption{Performance of HSA with sparsity constraint on the overemphasized network layers; last fully connected layer and initial convolutional layer. For the numerical experiments, we consider IID data split on CIFAR-10 datas. Here, we examine ResNet-20 architecture and $\delta$ denotes global sparsity, $\delta^{max}_{cw1}$ denotes maximum sparsity for the first Conv layer, and $\delta^{max}_{fc}$ denotes maximum sparsity for the last FC layer. Best attack performances (lowest test accuracy) are highlighted in \textbf{bold}, second-best values are \underline{underlined}.}
\label{tab:LayerConstraintsIID}
\begin{tabular}{@{}clllllll@{}}
\toprule
\textbf{Sparsity Pattern $\delta / \delta^{max}_{cw1} / \delta^{max}_{fc}$} & \textbf{Bulyan} & \textbf{CC} & \textbf{M-Krum} & \textbf{RFA} & \textbf{TM} & \textbf{GAS-Bulyan} & \textbf{GAS-Krum} \\ \midrule
$5 \times 10^{-3}$ / $0.2$ / $0.2$ & 33.2 $\pm$ 1.9 & 42.2 $\pm$ 1.0 & 73.8 $\pm$ 0.9 & 52.0 $\pm$ 8.0 & 60.2 $\pm$ 0.7 & 76.5 $\pm$ 0.3 & 75.3 $\pm$ 0.6 \\
$5 \times 10^{-3}$ / $0.4$ / $0.2$ & \textbf{25.1 $\pm$ 5.2} & 46.9 $\pm$ 1.9 & 74.6 $\pm$ 1.8 & 57.5 $\pm$ 2.2 & 66.6 $\pm$ 0.6 & \underline{73.1 $\pm$ 2.6} & 77.1 $\pm$ 0.3 \\
$5 \times 10^{-3}$ / $1$ / $0.2$ & 30.9 $\pm$ 2.1 & \textbf{38.5 $\pm$ 3.4} & \underline{65.1 $\pm$ 1.4} & \textbf{44.1 $\pm$ 6.1} & \underline{55.0 $\pm$ 13.0} & 73.3 $\pm$ 0.3 & 73.5 $\pm$ 1.1 \\
$5 \times 10^{-3}$ / $0.2$ / $0.4$ & \underline{28.3 $\pm$ 2.3} & 54.4 $\pm$ 9.3 & 75.2 $\pm$ 1.4 & 56.8 $\pm$ 4.8 & 70.8 $\pm$ 1.0 & 73.3 $\pm$ 0.6 & 75.2 $\pm$ 0.3 \\
$5 \times 10^{-3}$ / $0.4$ / $0.4$ & 37.5 $\pm$ 6.0 & 47.2 $\pm$ 2.7 & 74.2 $\pm$ 1.6 & 55.8 $\pm$ 2.0 & 67.4 $\pm$ 0.1 & \textbf{71.5 $\pm$ 3.8} & \textbf{70.2 $\pm$ 2.3} \\
$5 \times 10^{-3}$ / $1$ / $0.4$ & 30.8 $\pm$ 4.8 & \underline{40.5 $\pm$ 4.9} & \textbf{64.8 $\pm$ 4.7} & 58.6 $\pm$ 2.3 & \textbf{50.1 $\pm$ 3.3} & 73.3 $\pm$ 0.6 & \underline{72.3 $\pm$ 0.6} \\
$5 \times 10^{-3}$ / $0.2$ / $1$ & 31.6 $\pm$ 5.5 & 49.6 $\pm$ 6.6 & 74.7 $\pm$ 0.9 & 53.0 $\pm$ 7.9 & 67.0 $\pm$ 2.0 & 74.5 $\pm$ 1.0 & 77.3 $\pm$ 1.4 \\
$5 \times 10^{-3}$ / $0.4$ / $1$ & 29.5 $\pm$ 7.0 & 44.3 $\pm$ 6.3 & 73.8 $\pm$ 1.4 & 57.5 $\pm$ 1.6 & 70.4 $\pm$ 0.5 & 74.4 $\pm$ 0.2 & 72.3 $\pm$ 0.6 \\
$5 \times 10^{-3}$ / $1$ / $1$ & 29.1 $\pm$ 4.8 & 43.0 $\pm$ 3.4 & 72.4 $\pm$ 1.6 & \underline{50.4 $\pm$ 4.0} & 56.4 $\pm$ 15.3 & 75.7 $\pm$ 0.2 & 75.4 $\pm$ 1.8 \\
\midrule
$1 \times 10^{-2}$ / $0.2$ / $0.2$ & 32.7 $\pm$ 7.3 & 47.5 $\pm$ 3.2 & 73.0 $\pm$ 0.2 & \underline{51.5 $\pm$ 2.2} & 67.5 $\pm$ 0.4 & \underline{73.4 $\pm$ 0.8} & \underline{73.8 $\pm$ 0.1} \\
$1 \times 10^{-2}$ / $0.4$ / $0.2$ & 36.5 $\pm$ 1.2 & 45.3 $\pm$ 3.9 & 75.6 $\pm$ 0.4 & 58.0 $\pm$ 0.6 & 53.9 $\pm$ 17.8 & 74.4 $\pm$ 1.1 & 75.3 $\pm$ 1.3 \\
$1 \times 10^{-2}$ / $1$ / $0.2$ & 32.9 $\pm$ 2.0 & 39.4 $\pm$ 2.6 & \underline{59.0 $\pm$ 12.4} & \textbf{46.4 $\pm$ 8.6} & \textbf{40.8 $\pm$ 3.7} & 74.7 $\pm$ 0.0 & 74.2 $\pm$ 1.0 \\
$1 \times 10^{-2}$ / $0.2$ / $0.4$ & 34.0 $\pm$ 2.9 & 45.3 $\pm$ 0.5 & 75.7 $\pm$ 0.6 & 56.6 $\pm$ 0.4 & 68.6 $\pm$ 2.6 & 75.8 $\pm$ 0.1 & 75.0 $\pm$ 1.6 \\
$1 \times 10^{-2}$ / $0.4$ / $0.4$ & 33.1 $\pm$ 7.9 & \textbf{37.3 $\pm$ 6.1} & 74.8 $\pm$ 0.6 & 61.0 $\pm$ 0.0 & 70.6 $\pm$ 1.2 & 76.0 $\pm$ 1.3 & 75.6 $\pm$ 0.7 \\
$1 \times 10^{-2}$ / $1$ / $0.4$ & \underline{31.3 $\pm$ 6.3} & 41.9 $\pm$ 3.9 & \textbf{58.0 $\pm$ 3.2} & 53.0 $\pm$ 11.1 & \underline{46.8 $\pm$ 3.8} & 74.5 $\pm$ 1.6 & 75.5 $\pm$ 1.4 \\
$1 \times 10^{-2}$ / $0.2$ / $1$ & 33.7 $\pm$ 5.2 & 44.2 $\pm$ 1.9 & 74.3 $\pm$ 0.8 & 62.2 $\pm$ 2.8 & 67.9 $\pm$ 2.8 & 76.0 $\pm$ 1.6 & \textbf{71.8 $\pm$ 2.7} \\
$1 \times 10^{-2}$ / $0.4$ / $1$ & \textbf{31.2 $\pm$ 1.5} & 47.2 $\pm$ 3.3 & 74.9 $\pm$ 1.6 & 55.6 $\pm$ 0.6 & 66.6 $\pm$ 2.6 & 74.9 $\pm$ 0.1 & 74.6 $\pm$ 1.2 \\
$1 \times 10^{-2}$ / $1$ / $1$ & 31.7 $\pm$ 7.6 & \underline{39.2 $\pm$ 2.7} & 68.9 $\pm$ 1.2 & 52.8 $\pm$ 5.6 & 53.8 $\pm$ 10.9 & \textbf{72.9 $\pm$ 1.3} & 74.4 $\pm$ 1.9 \\
\midrule
$5 \times 10^{-2}$ / $0.2$ / $0.2$ & 31.0 $\pm$ 0.8 & 46.3 $\pm$ 1.6 & 72.2 $\pm$ 1.4 & 48.9 $\pm$ 1.5 & 63.6 $\pm$ 1.2 & \textbf{69.5 $\pm$ 2.3} & 72.7 $\pm$ 1.2 \\
$5 \times 10^{-2}$ / $0.4$ / $0.2$ & 29.8 $\pm$ 2.3 & \underline{42.4 $\pm$ 5.1} & 68.9 $\pm$ 1.1 & \underline{48.5 $\pm$ 6.5} & 67.0 $\pm$ 1.2 & 72.6 $\pm$ 1.6 & 72.4 $\pm$ 2.1 \\
$5 \times 10^{-2}$ / $1$ / $0.2$ & \underline{29.5 $\pm$ 1.8} & 42.6 $\pm$ 1.8 & \textbf{50.6 $\pm$ 19.9} & 57.5 $\pm$ 4.1 & 53.2 $\pm$ 12.0 & \underline{70.4 $\pm$ 2.3} & 72.7 $\pm$ 0.7 \\
$5 \times 10^{-2}$ / $0.2$ / $0.4$ & \textbf{29.4 $\pm$ 7.5} & 45.9 $\pm$ 3.7 & 62.6 $\pm$ 2.2 & \textbf{\textcolor{red}{41.7 $\pm$ 4.8}} & 54.6 $\pm$ 14.7 & 73.8 $\pm$ 1.5 & \textbf{\textcolor{red}{57.6 $\pm$ 17.0}} \\
$5 \times 10^{-2}$ / $0.4$ / $0.4$ & 30.9 $\pm$ 5.0 & 51.7 $\pm$ 1.2 & 65.2 $\pm$ 1.1 & 52.9 $\pm$ 6.6 & 64.4 $\pm$ 0.6 & 73.3 $\pm$ 0.4 & 75.4 $\pm$ 1.1 \\
$5 \times 10^{-2}$ / $1$ / $0.4$ & 32.1 $\pm$ 1.4 & 44.9 $\pm$ 9.3 & \underline{53.0 $\pm$ 13.4} & 57.4 $\pm$ 6.0 & \underline{49.7 $\pm$ 0.6} & 73.7 $\pm$ 0.0 & 72.7 $\pm$ 0.5 \\
$5 \times 10^{-2}$ / $0.2$ / $1$ & 41.2 $\pm$ 2.3 & 43.9 $\pm$ 7.6 & 60.0 $\pm$ 13.5 & 56.9 $\pm$ 2.5 & 55.0 $\pm$ 12.4 & 73.2 $\pm$ 0.1 & 71.3 $\pm$ 1.6 \\
$5 \times 10^{-2}$ / $0.4$ / $1$ & 37.1 $\pm$ 3.8 & 43.6 $\pm$ 6.2 & 68.4 $\pm$ 1.3 & 58.5 $\pm$ 5.0 & \textbf{44.3 $\pm$ 4.1} & 73.1 $\pm$ 2.0 & \underline{71.0 $\pm$ 1.0} \\
$5 \times 10^{-2}$ / $1$ / $1$ & 33.1 $\pm$ 11.1 & \textbf{\textcolor{red}{34.6 $\pm$ 4.2}} & 66.2 $\pm$ 6.9 & 54.1 $\pm$ 7.4 & 55.4 $\pm$ 11.5 & 72.8 $\pm$ 1.6 & 71.44 $\pm$ 2.3 \\
\midrule
$1 \times 10^{-1}$ / $0.2$ / $0.2$ & \underline{26.7 $\pm$ 0.9} & \textbf{41.6 $\pm$ 0.8} & 60.9 $\pm$ 0.5 & 50.0 $\pm$ 0.3 & 43.9 $\pm$ 9.0 & 70.0 $\pm$ 0.0 & 70.2 $\pm$ 0.5 \\
$1 \times 10^{-1}$ / $0.4$ / $0.2$ & 30.3 $\pm$ 0.5 & \underline{43.6 $\pm$ 1.2} & 62.7 $\pm$ 3.1 & \underline{48.7 $\pm$ 6.1} & 54.0 $\pm$ 11.9 & 68.9 $\pm$ 0.6 & 70.5 $\pm$ 0.3 \\
$1 \times 10^{-1}$ / $1$ / $0.2$ & \textbf{\textcolor{red}{24.1 $\pm$ 0.7}} & 43.6 $\pm$ 2.4 & \underline{59.4 $\pm$ 13.8} & 52.4 $\pm$ 1.2 & 40.9 $\pm$ 3.3 & 71.0 $\pm$ 1.9 & 69.5 $\pm$ 0.2 \\
$1 \times 10^{-1}$ / $0.2$ / $0.4$ & 31.3 $\pm$ 7.0 & 46.4 $\pm$ 0.5 & 61.0 $\pm$ 1.5 & \textbf{46.5 $\pm$ 6.9} & 49.6 $\pm$ 11.1 & \textbf{\textcolor{red}{66.5 $\pm$ 0.3}} & 70.3 $\pm$ 0.3 \\
$1 \times 10^{-1}$ / $0.4$ / $0.4$ & 30.1 $\pm$ 6.1 & 43.7 $\pm$ 4.9 & 59.4 $\pm$ 2.4 & 53.3 $\pm$ 1.6 & 56.1 $\pm$ 7.0 & 70.5 $\pm$ 1.3 & \underline{66.5 $\pm$ 4.0} \\
$1 \times 10^{-1}$ / $1$ / $0.4$ & 28.3 $\pm$ 5.9 & 46.1 $\pm$ 6.6 & \textbf{\textcolor{red}{50.2 $\pm$ 8.5}} & 53.4 $\pm$ 5.6 & \textbf{\textcolor{red}{36.6 $\pm$ 1.9}} & \underline{67.7 $\pm$ 2.5} & \textbf{65.1 $\pm$ 4.1} \\
$1 \times 10^{-1}$ / $0.2$ / $1$ & 49.6 $\pm$ 1.7 & 50.1 $\pm$ 6.1 & 67.2 $\pm$ 1.6 & 58.7 $\pm$ 1.9 & 62.8 $\pm$ 7.7 & 70.3 $\pm$ 0.2 & 70.5 $\pm$ 2.2 \\
$1 \times 10^{-1}$ / $0.4$ / $1$ & 49.7 $\pm$ 3.1 & 49.6 $\pm$ 3.1 & 63.8 $\pm$ 2.2 & 62.7 $\pm$ 2.4 & 48.2 $\pm$ 8.8 & 69.7 $\pm$ 0.7 & 71.0 $\pm$ 1.5 \\
$1 \times 10^{-1}$ / $1$ / $1$ & 43.4 $\pm$ 6.3 & 50.6 $\pm$ 3.6 & 66.8 $\pm$ 3.3 & 59.1 $\pm$ 3.9 & \underline{40.8 $\pm$ 8.7} & 70.6 $\pm$ 0.2 & 69.2 $\pm$ 0.9 \\
\bottomrule
\end{tabular}
\end{table*}

\begin{table*}[h!]
\caption{Impact of layer-wise sparsity constraints on DHSA performance in IID setting. $\delta$ denotes global sparsity, $\delta^{max}_{cw1}$ denotes maximum sparsity for the first Conv layer, and $\delta^{max}_{fc}$ denotes maximum sparsity for the last FC layer. denotes maximum sparsity for the last layer. Constraint value of 1 indicates no constraint. Best attack performances (lowest test accuracy) are highlighted in \textbf{bold}, second-best values are \underline{underlined}.}
\label{tab:LayerConstraintsIID-DHSA}
\begin{tabular}{@{}clllllll@{}}
\toprule
\textbf{Sparsity Pattern $\delta / \delta^{max}_{cw1} / \delta^{max}_{fc}$} & \textbf{Bulyan} & \textbf{CC} & \textbf{M-Krum} & \textbf{RFA} & \textbf{TM} & \textbf{GAS-Bulyan} & \textbf{GAS-Krum} \\ \midrule
$5 \times 10^{-3}$ / $0.2$ / $0.2$ & 41.7 $\pm$ 2.3 & \underline{42.1 $\pm$ 5.8} & 61.1 $\pm$ 3.2 & 54.0 $\pm$ 2.9 & 54.5 $\pm$ 2.1 & 48.4 $\pm$ 3.3 & 58.0 $\pm$ 2.5 \\
$5 \times 10^{-3}$ / $0.4$ / $0.2$ & \underline{38.4 $\pm$ 3.3} & 43.2 $\pm$ 4.1 & 59.1 $\pm$ 1.5 & 48.5 $\pm$ 0.8 & \textbf{47.4 $\pm$ 11.8} & 56.0 $\pm$ 3.8 & 57.8 $\pm$ 1.9 \\
$5 \times 10^{-3}$ / $1$ / $0.2$ & 38.5 $\pm$ 6.4 & 43.4 $\pm$ 11.7 & 58.7 $\pm$ 1.5 & 47.3 $\pm$ 4.5 & 54.0 $\pm$ 2.3 & 59.3 $\pm$ 1.7 & 57.4 $\pm$ 0.7 \\
$5 \times 10^{-3}$ / $0.2$ / $0.4$ & 38.9 $\pm$ 2.6 & 47.9 $\pm$ 9.2 & 58.7 $\pm$ 3.2 & 47.6 $\pm$ 5.7 & 53.0 $\pm$ 1.3 & \textbf{45.6 $\pm$ 3.2} & 59.0 $\pm$ 2.0 \\
$5 \times 10^{-3}$ / $0.4$ / $0.4$ & 38.8 $\pm$ 4.6 & 55.4 $\pm$ 6.8 & \underline{56.8 $\pm$ 2.2} & \underline{42.6 $\pm$ 7.3} & \underline{51.0 $\pm$ 6.6} & 48.1 $\pm$ 2.3 & \textbf{56.7 $\pm$ 3.0} \\
$5 \times 10^{-3}$ / $1$ / $0.4$ & 39.8 $\pm$ 2.9 & \textbf{40.5 $\pm$ 6.7} & 60.0 $\pm$ 1.7 & \textbf{42.0 $\pm$ 9.9} & 51.8 $\pm$ 3.4 & 51.8 $\pm$ 5.6 & 59.5 $\pm$ 2.1 \\
$5 \times 10^{-3}$ / $0.2$ / $1$ & \textbf{37.0 $\pm$ 3.5} & 47.0 $\pm$ 12.0 & \textbf{52.5 $\pm$ 9.1} & 44.5 $\pm$ 4.6 & 54.9 $\pm$ 2.9 & \underline{44.5 $\pm$ 13.6} & 59.7 $\pm$ 0.4 \\
$5 \times 10^{-3}$ / $0.4$ / $1$ & 45.2 $\pm$ 5.5 & 49.0 $\pm$ 6.1 & 59.6 $\pm$ 4.0 & 45.7 $\pm$ 8.8 & 52.1 $\pm$ 3.7 & 54.0 $\pm$ 7.4 & \underline{57.3 $\pm$ 1.2} \\
$5 \times 10^{-3}$ / $1$ / $1$ & 40.7 $\pm$ 3.6 & 48.9 $\pm$ 3.1 & 60.2 $\pm$ 1.1 & 50.2 $\pm$ 2.9 & 56.8 $\pm$ 1.0 & 54.9 $\pm$ 7.7 & 58.6 $\pm$ 3.5 \\
\midrule
$1 \times 10^{-2}$ / $0.4$ / $0.2$ & \textbf{39.1 $\pm$ 3.3} & \underline{42.1 $\pm$ 5.4} & 55.2 $\pm$ 2.5 & 50.9 $\pm$ 2.1 & 55.1 $\pm$ 0.5 & 59.7 $\pm$ 0.8 & 59.7 $\pm$ 1.5 \\
$1 \times 10^{-2}$ / $1$ / $0.2$ & 44.0 $\pm$ 0.8 & 44.6 $\pm$ 1.9 & 56.8 $\pm$ 3.2 & \textbf{45.8 $\pm$ 1.4} & \textbf{44.4 $\pm$ 9.7} & 56.7 $\pm$ 3.9 & \textbf{58.5 $\pm$ 2.4} \\
$1 \times 10^{-2}$ / $0.2$ / $0.4$ & 50.2 $\pm$ 3.2 & 51.4 $\pm$ 4.6 & 59.7 $\pm$ 3.3 & 50.9 $\pm$ 2.4 & 56.3 $\pm$ 2.6 & 63.3 $\pm$ 0.3 & 64.1 $\pm$ 1.3 \\
$1 \times 10^{-2}$ / $0.4$ / $0.4$ & \underline{42.0 $\pm$ 5.7} & 52.0 $\pm$ 5.8 & \textbf{50.2 $\pm$ 7.9} & 53.3 $\pm$ 1.1 & \underline{52.8 $\pm$ 2.4} & 60.5 $\pm$ 2.5 & 62.8 $\pm$ 1.8 \\
$1 \times 10^{-2}$ / $1$ / $0.4$ & 46.1 $\pm$ 4.4 & \textbf{41.2 $\pm$ 8.1} & 55.0 $\pm$ 3.4 & \underline{45.9 $\pm$ 5.2} & 54.7 $\pm$ 2.4 & 58.5 $\pm$ 2.7 & 60.1 $\pm$ 2.1 \\
$1 \times 10^{-2}$ / $0.2$ / $1$ & 49.6 $\pm$ 3.3 & 52.7 $\pm$ 3.4 & 60.3 $\pm$ 3.1 & 52.2 $\pm$ 0.2 & 57.3 $\pm$ 1.1 & 62.1 $\pm$ 0.8 & \underline{59.5 $\pm$ 2.9} \\
$1 \times 10^{-2}$ / $0.4$ / $1$ & 42.9 $\pm$ 1.1 & \underline{48.7 $\pm$ 3.1} & 56.5 $\pm$ 4.4 & 56.3 $\pm$ 2.4 & 56.0 $\pm$ 3.5 & 59.2 $\pm$ 0.7 & 62.7 $\pm$ 2.2 \\
$1 \times 10^{-2}$ / $1$ / $1$ & \underline{47.7 $\pm$ 4.5} & 47.0 $\pm$ 4.3 & \underline{53.8 $\pm$ 9.3} & 54.5 $\pm$ 1.8 & 57.0 $\pm$ 2.5 & 59.8 $\pm$ 1.6 & 62.3 $\pm$ 1.1 \\
\midrule
$5 \times 10^{-2}$ / $0.2$ / $0.2$ & \underline{34.5 $\pm$ 2.2} & 43.6 $\pm$ 9.1 & 56.0 $\pm$ 0.3 & \textbf{\textcolor{red}{38.4 $\pm$ 3.6}} & \textbf{\textcolor{red}{34.1 $\pm$ 8.6}} & \underline{57.3 $\pm$ 3.3} & \textbf{53.9 $\pm$ 1.2} \\
$5 \times 10^{-2}$ / $0.4$ / $0.2$ & 41.3 $\pm$ 3.6 & \underline{41.9 $\pm$ 11.7} & \textbf{\textcolor{red}{42.2 $\pm$ 9.9}} & 43.4 $\pm$ 2.6 & \underline{41.7 $\pm$ 4.6} & \textbf{49.8 $\pm$ 13.9} & \underline{56.3 $\pm$ 1.6} \\
$5 \times 10^{-2}$ / $1$ / $0.2$ & \textbf{33.7 $\pm$ 3.2} & 42.5 $\pm$ 5.5 & \underline{52.0 $\pm$ 7.0} & 48.4 $\pm$ 0.9 & 45.5 $\pm$ 12.1 & 58.1 $\pm$ 4.6 & 56.3 $\pm$ 3.2 \\
$5 \times 10^{-2}$ / $0.2$ / $0.4$ & 48.4 $\pm$ 4.3 & 45.4 $\pm$ 4.5 & 54.7 $\pm$ 7.0 & 46.5 $\pm$ 4.5 & 43.6 $\pm$ 3.4 & 60.5 $\pm$ 1.8 & 66.1 $\pm$ 0.6 \\
$5 \times 10^{-2}$ / $0.4$ / $0.4$ & 46.3 $\pm$ 4.8 & 45.4 $\pm$ 6.0 & 57.2 $\pm$ 1.8 & \textbf{43.8 $\pm$ 9.9} & 48.8 $\pm$ 3.0 & 59.6 $\pm$ 0.6 & 63.3 $\pm$ 3.6 \\
$5 \times 10^{-2}$ / $1$ / $0.4$ & 52.1 $\pm$ 1.0 & 46.4 $\pm$ 4.7 & 53.9 $\pm$ 5.5 & 51.6 $\pm$ 4.2 & 47.6 $\pm$ 2.6 & 58.9 $\pm$ 2.2 & 59.7 $\pm$ 2.2 \\
$5 \times 10^{-2}$ / $0.2$ / $1$ & 50.9 $\pm$ 2.2 & 45.2 $\pm$ 5.2 & 61.1 $\pm$ 3.6 & 57.7 $\pm$ 2.7 & 43.9 $\pm$ 9.1 & 64.6 $\pm$ 4.8 & 65.2 $\pm$ 2.0 \\
$5 \times 10^{-2}$ / $0.4$ / $1$ & 55.0 $\pm$ 3.9 & 42.3 $\pm$ 1.9 & 57.5 $\pm$ 4.4 & 54.5 $\pm$ 6.0 & 49.2 $\pm$ 4.7 & 67.5 $\pm$ 1.8 & 69.6 $\pm$ 1.1 \\
$5 \times 10^{-2}$ / $1$ / $1$ & 55.3 $\pm$ 3.8 & \textbf{41.4 $\pm$ 2.8} & 64.8 $\pm$ 2.3 & 51.7 $\pm$ 0.7 & 57.6 $\pm$ 0.9 & 68.2 $\pm$ 1.2 & 62.0 $\pm$ 6.4 \\
\midrule
$1 \times 10^{-1}$ / $0.2$ / $0.2$ & \textbf{\textcolor{red}{31.4 $\pm$ 4.0}} & \underline{39.2 $\pm$ 1.5} & \textbf{50.6 $\pm$ 7.2} & 47.9 $\pm$ 4.5 & \textbf{36.5 $\pm$ 2.6} & 45.7 $\pm$ 1.9 & \textbf{\textcolor{red}{48.6 $\pm$ 13.5}} \\
$1 \times 10^{-1}$ / $0.4$ / $0.2$ & 34.1 $\pm$ 0.9 & 43.7 $\pm$ 9.5 & 57.3 $\pm$ 3.3 & \underline{46.7 $\pm$ 6.2} & 49.8 $\pm$ 6.7 & 58.2 $\pm$ 1.8 & \underline{50.0 $\pm$ 12.1} \\
$1 \times 10^{-1}$ / $1$ / $0.2$ & \underline{34.0 $\pm$ 0.7} & \textbf{\textcolor{red}{38.1 $\pm$ 1.1}} & 58.1 $\pm$ 5.1 & \textbf{39.5 $\pm$ 7.7} & 46.7 $\pm$ 8.1 & \textbf{\textcolor{red}{42.0 $\pm$ 7.5}} & 64.0 $\pm$ 2.0 \\
$1 \times 10^{-1}$ / $0.2$ / $0.4$ & 43.8 $\pm$ 6.9 & 48.4 $\pm$ 5.8 & \underline{54.5 $\pm$ 2.2} & 57.3 $\pm$ 3.3 & \underline{42.7 $\pm$ 1.6} & 58.5 $\pm$ 6.1 & 52.5 $\pm$ 11.7 \\
$1 \times 10^{-1}$ / $0.4$ / $0.4$ & 47.8 $\pm$ 2.2 & 45.7 $\pm$ 7.5 & 57.7 $\pm$ 4.2 & 47.8 $\pm$ 4.9 & 52.5 $\pm$ 4.7 & 53.8 $\pm$ 9.7 & 64.3 $\pm$ 4.6 \\
$1 \times 10^{-1}$ / $1$ / $0.4$ & 47.8 $\pm$ 3.1 & 42.6 $\pm$ 2.4 & 59.0 $\pm$ 1.9 & 51.5 $\pm$ 4.7 & 55.2 $\pm$ 5.0 & 47.8 $\pm$ 17.4 & 64.2 $\pm$ 2.5 \\
$1 \times 10^{-1}$ / $0.2$ / $1$ & 51.2 $\pm$ 5.9 & 47.1 $\pm$ 6.4 & 61.6 $\pm$ 6.7 & 61.8 $\pm$ 0.4 & 61.3 $\pm$ 4.2 & 68.5 $\pm$ 0.5 & 67.8 $\pm$ 1.6 \\
$1 \times 10^{-1}$ / $0.4$ / $1$ & 50.1 $\pm$ 2.9 & 48.7 $\pm$ 1.8 & 67.7 $\pm$ 0.8 & 60.9 $\pm$ 0.7 & 63.2 $\pm$ 3.2 & 69.9 $\pm$ 0.9 & 71.1 $\pm$ 0.3 \\
$1 \times 10^{-1}$ / $1$ / $1$ & 52.5 $\pm$ 2.4 & 50.5 $\pm$ 11.9 & 63.9 $\pm$ 1.4 & 59.0 $\pm$ 5.1 & 55.5 $\pm$ 9.2 & 70.1 $\pm$ 0.8 & 70.4 $\pm$ 1.4 \\
\bottomrule
\end{tabular}
\end{table*}

\begin{table*}[h!]
\caption{ Impact of layer-wise sparsity constraints on DHSA performance in non-IID setting. $\delta$ denotes global sparsity / $\delta^{max}_{cw1}$ denotes maximum sparsity for the first Conv layer / and $\delta^{max}_{fc}$ denotes maximum sparsity for the last FC layer. Constraint value of 1 indicates no constraint. Best attack performances (lowest test accuracy) are highlighted in \textbf{bold} / second-best values are \underline{underlined}.}
\label{tab:LayerConstraintsDIR-DHSA}
\begin{tabular}{@{}clllllll@{}}
\toprule
\textbf{Sparsity Pattern $(\delta / \delta^{max}_{cw1} / \delta^{max}_{fc})$} & \textbf{Bulyan} & \textbf{CC} & \textbf{M-Krum} & \textbf{RFA} & \textbf{TM} & \textbf{GAS-Bulyan} & \textbf{GAS-Krum} \\ \midrule
$5 \times 10^{-3}$ / $0.2$ / $0.2$ & \underline{35.5 $\pm$ 20.7} & 25.3 $\pm$ 1.1 & 42.4 $\pm$ 12.2 & \underline{24.1 $\pm$ 14.9} & \textbf{\textcolor{red}{10.0 $\pm$ 0.0}} & \underline{22.7 $\pm$ 18.0} & \underline{10.0 $\pm$ 0.0} \\
$5 \times 10^{-3}$ / $0.4$ / $0.2$ & 36.3 $\pm$ 13.6 & 24.4 $\pm$ 3.6 & 60.9 $\pm$ 2.7 & 40.6 $\pm$ 19.7 & 10.5 $\pm$ 0.7 & 30.1 $\pm$ 14.6 & \textbf{\textcolor{red}{10.0 $\pm$ 0.0}} \\
$5 \times 10^{-3}$ / $1$ / $0.2$ & 51.0 $\pm$ 22.8 & 27.4 $\pm$ 0.8 & 48.6 $\pm$ 11.3 & 47.0 $\pm$ 14.7 & 10.0 $\pm$ 0.0 & \textbf{16 $\pm$ 8.6} & 10.0 $\pm$ 0.0\\
$5 \times 10^{-3}$ / $0.2$ / $0.4$ & 47.0 $\pm$ 19.4 & \underline{18.2 $\pm$ 3.2} & 42.2 $\pm$ 14.6 & 24.1 $\pm$ 14.9 & 10.0 $\pm$ 0.0 & 35.9 $\pm$ 20.9 & 10.0 $\pm$ 0.0 \\
$5 \times 10^{-3}$ / $0.4$ / $0.4$ & 53.6 $\pm$ 5.7 & \textbf{\textcolor{red}{16.4 $\pm$ 1.1}} & \textbf{25.3 $\pm$ 6.1} & 21.2 $\pm$ 9.2 & 10.0 $\pm$ 0.0 & 27.3 $\pm$ 23.3 & 10.0 $\pm$ 0.0 \\
$5 \times 10^{-3}$ / $1$ / $0.4$ & 47.2 $\pm$ 23.0 & 26.4 $\pm$ 4.6 & 36.9 $\pm$ 21.4 & 27.2 $\pm$ 15.9 & 10.0 $\pm$ 0.0 & 27.7 $\pm$ 25.0 & 10.0 $\pm$ 0.0 \\
$5 \times 10^{-3}$ / $0.2$ / $1$ & 36.8 $\pm$ 13.6 & 21.1 $\pm$ 2.7 & \underline{30.9 $\pm$ 14.0} & \textbf{17.6 $\pm$ 10.7} & 10.0 $\pm$ 0.0 & 32.5 $\pm$ 24.1 & 10.0 $\pm$ 0.0 \\
$5 \times 10^{-3}$ / $0.4$ / $1$ & 50.5 $\pm$ 15.3 & 23.6 $\pm$ 6.1 & 51.7 $\pm$ 8.8 & 31.4 $\pm$ 20.7 & 10.0 $\pm$ 0.0 & 33.5 $\pm$ 18.2 & 10.0 $\pm$ 0.0 \\
$5 \times 10^{-3}$ / $1$ / $1$ & \textbf{33.3 $\pm$ 20.6} & 24.7 $\pm$ 3.0 & 45.4 $\pm$ 16.2 & 20.8 $\pm$ 13.5 & \underline{10.5 $\pm$ 0.7} & 11.0 $\pm$ 1.4 & \textbf{18.1 $\pm$ 11.4} \\ \midrule
$1 \times 10^{-2}$ / $0.2$ / $0.2$ & \underline{26.5 $\pm$ 9.6} & \underline{23.1 $\pm$ 3.3} & 42.1 $\pm$ 20.1 & \textbf{\textcolor{red}{10.0 $\pm$ 0.0}} & 10.0 $\pm$ 0.0 & 25.7 $\pm$ 22.2 & 10.0 $\pm$ 0.0 \\
$1 \times 10^{-2}$ / $0.4$ / $0.2$ & 34.5 $\pm$ 16.5 & \textbf{20.7 $\pm$ 1.4} & 38.0 $\pm$ 14.1 & \underline{15.1 $\pm$ 5.7} & \textbf{10.0 $\pm$ 0.0} & \underline{10.6 $\pm$ 0.8} & \textbf{\textcolor{red}{10.0 $\pm$ 0.0}} \\
$1 \times 10^{-2}$ / $1$ / $0.2$ & \textbf{24.0 $\pm$ 5.8} & 26.2 $\pm$ 2.6 & \underline{28.6 $\pm$ 6.8} & \textbf{11.6 $\pm$ 2.2} & 10.0 $\pm$ 0.0 & \textbf{10.0 $\pm$ 0.0} & 10.0 $\pm$ 0.0 \\
$1 \times 10^{-2}$ / $0.2$ / $0.4$ & 46.6 $\pm$ 10.9 & 32.3 $\pm$ 4.9 & 57.8 $\pm$ 8.3 & 35.8 $\pm$ 9.2 & 10.0 $\pm$ 0.0 & 42.0 $\pm$ 20.8 & 10.0 $\pm$ 0.0 \\
$1 \times 10^{-2}$ / $0.4$ / $0.4$ & 38.4 $\pm$ 26.9 & \underline{28.8 $\pm$ 1.3} & 55.5 $\pm$ 20.2 & 23.1 $\pm$ 13.4 & 10.0 $\pm$ 0.0 & 41.1 $\pm$ 21.9 & 10.0 $\pm$ 0.0 \\
$1 \times 10^{-2}$ / $1$ / $0.4$ & 36.6 $\pm$ 21.7 & 28.2 $\pm$ 2.6 & 51.1 $\pm$ 10.7 & 28.2 $\pm$ 20.0 & 10.0 $\pm$ 0.0 & 27.8 $\pm$ 23.9 & 10.0 $\pm$ 0.0 \\
$1 \times 10^{-2}$ / $0.2$ / $1$ & 46.6 $\pm$ 10.5 & 25.4 $\pm$ 4.2 & \textbf{28.6 $\pm$ 6.8} & 39.6 $\pm$ 18.5 & 10.0 $\pm$ 0.0 & 43.8 $\pm$ 24.1 & \underline{13.3 $\pm$ 4.7} \\
$1 \times 10^{-2}$ / $0.4$ / $1$ & 58.0 $\pm$ 14.5 & 29.1 $\pm$ 2.7 & \underline{44.2 $\pm$ 11.5} & 35.8 $\pm$ 19.1 & 10.0 $\pm$ 0.0 & 39.0 $\pm$ 20.6 & \textbf{11.1 $\pm$ 1.5} \\
$1 \times 10^{-2}$ / $1$ / $1$ & 46.0 $\pm$ 14.5 & \underline{24.1 $\pm$ 3.3} & 47.8 $\pm$ 18.1 & 26.1 $\pm$ 17.5 & \underline{10.0 $\pm$ 0.0} & 59.6 $\pm$ 10.7 & 10.0 $\pm$ 0.0 \\
\midrule
$5 \times 10^{-2}$ / $0.2$ / $0.2$ & \textbf{12.9 $\pm$ 0.4} & \underline{20.6 $\pm$ 4.2} & 32.8 $\pm$ 11.1 & 10.0 $\pm$ 0.0 & \textbf{10.0 $\pm$ 0.0} & 20.5 $\pm$ 14.8 & \textbf{10.8 $\pm$ 1.1} \\
$5 \times 10^{-2}$ / $0.4$ / $0.2$ & 19.8 $\pm$ 3.6 & 21.6 $\pm$ 3.2 & \underline{28.0 $\pm$ 3.2} & 10.0 $\pm$ 0.0 & 10.0 $\pm$ 0.0 & \textbf{\textcolor{red}{10.6 $\pm$ 0.9}} & 10.0 $\pm$ 0.0 \\
$5 \times 10^{-2}$ / $1$ / $0.2$ & \underline{14.6 $\pm$ 2.2} & \textbf{16.8 $\pm$ 4.6} & \textbf{24.0 $\pm$ 4.9} & 10.0 $\pm$ 0.0 & 10.0 $\pm$ 0.0 & \textbf{\textcolor{red}{10.0 $\pm$ 0.0}} & 10.0 $\pm$ 0.0 \\
$5 \times 10^{-2}$ / $0.2$ / $0.4$ & 29.3 $\pm$ 12.3 & 25.5 $\pm$ 1.9 & 34.1 $\pm$ 14.8 & \textbf{10.0 $\pm$ 0.0} & 10.0 $\pm$ 0.0 & \underline{13.7 $\pm$ 5.2} & 10.0 $\pm$ 0.0 \\
$5 \times 10^{-2}$ / $0.4$ / $0.4$ & 30.8 $\pm$ 7.4 & 23.0 $\pm$ 1.2 & 42.4 $\pm$ 12.2 & 10.0 $\pm$ 0.0 & 10.0 $\pm$ 0.0 & 56.8 $\pm$ 1.7 & 10.0 $\pm$ 0.0 \\
$5 \times 10^{-2}$ / $1$ / $0.4$ & 28.1 $\pm$ 12.6 & 22.1 $\pm$ 3.2 & 38.7 $\pm$ 10.0 & 10.0 $\pm$ 0.0 & 10.0 $\pm$ 0.0 & \underline{24.9 $\pm$ 21.1} & \underline{10.0 $\pm$ 0.0} \\
$5 \times 10^{-2}$ / $0.2$ / $1$ & 48.3 $\pm$ 13.0 & 26.1 $\pm$ 5.4 & 41.3 $\pm$ 11.6 & 10.0 $\pm$ 0.0 & 10.0 $\pm$ 0.0 & 72.8 $\pm$ 3.2 & 10.0 $\pm$ 0.0 \\
$5 \times 10^{-2}$ / $0.4$ / $1$ & 39.4 $\pm$ 10.3 & 28.5 $\pm$ 3.2 & 45.7 $\pm$ 4.0 & \underline{11.9 $\pm$ 2.7} & \underline{10.0 $\pm$ 0.0} & \textbf{\textcolor{red}{10.0 $\pm$ 0.0}} & 10.0 $\pm$ 0.0 \\
$5 \times 10^{-2}$ / $1$ / $1$ & 43.6 $\pm$ 17.5 & 23.7 $\pm$ 0.8 & 46.6 $\pm$ 11.3 & 10.0 $\pm$ 0.0 & 10.0 $\pm$ 0.0 & 73.2 $\pm$ 6.6 & 10.0 $\pm$ 0.0 \\
\midrule
$1 \times 10^{-1}$ / $0.2$ / $0.2$ & 13.2 $\pm$ 3.8 & \textbf{19.5 $\pm$ 4.1} & \textbf{14.5 $\pm$ 3.4} & 10.0 $\pm$ 0.0 & \textbf{10.0 $\pm$ 0.0} & \textbf{10.0 $\pm$ 0.0} & 10.0 $\pm$ 0.0 \\
$1 \times 10^{-1}$ / $0.4$ / $0.2$ & \textbf{\textcolor{red}{11.5 $\pm$ 2.1}} & 22.8 $\pm$ 5.3 & 22.0 $\pm$ 8.8 & 10.0 $\pm$ 0.0 & 11.0 $\pm$ 1.5 & 10.0 $\pm$ 0.0 & 10.0 $\pm$ 0.0 \\
$1 \times 10^{-1}$ / $1$ / $0.2$ & 14.5 $\pm$ 3.4 & \underline{20.6 $\pm$ 2.5} & 29.5 $\pm$ 14.8 & 10.0 $\pm$ 0.0 & 10.0 $\pm$ 0.0 & 10.0 $\pm$ 0.0 & 10.0 $\pm$ 0.0 \\
$1 \times 10^{-1}$ / $0.2$ / $0.4$ & 23.4 $\pm$ 11.9 & 24.2 $\pm$ 1.5 & \underline{22.0 $\pm$ 8.8} & 10.0 $\pm$ 0.0 & 10.0 $\pm$ 0.0 & 23.1 $\pm$ 18.5 & 10.0 $\pm$ 0.0 \\
$1 \times 10^{-1}$ / $0.4$ / $0.4$ & 27.2 $\pm$ 5.0 & 21.2 $\pm$ 2.2 & 33.6 $\pm$ 17.1 & 10.0 $\pm$ 0.0 & 10.0 $\pm$ 0.0 & \underline{16.1 $\pm$ 8.6} & 10.0 $\pm$ 0.0 \\
$1 \times 10^{-1}$ / $1$ / $0.4$ & \underline{12.8 $\pm$ 3.1} & 20.9 $\pm$ 1.5 & 42.6 $\pm$ 5.5 & 10.0 $\pm$ 0.0 & 10.0 $\pm$ 0.0 & \underline{11.9 $\pm$ 1.4} & 10.0 $\pm$ 0.0 \\
$1 \times 10^{-1}$ / $0.2$ / $1$ & 53.0 $\pm$ 21.8 & 32.7 $\pm$ 5.5 & 55.4 $\pm$ 7.2 & \textbf{10.0 $\pm$ 0.0} & 10.0 $\pm$ 0.0 & 74.5 $\pm$ 2.3 & \textbf{\textcolor{red}{9.8 $\pm$ 0.3}} \\
$1 \times 10^{-1}$ / $0.4$ / $1$ & 52.3 $\pm$ 16.5 & 24.8 $\pm$ 5.2 & 58.2 $\pm$ 3.4 & 30.9 $\pm$ 5.3 & 10.0 $\pm$ 0.0 & 76.4 $\pm$ 2.8 & 10.0 $\pm$ 0.0 \\
$1 \times 10^{-1}$ / $1$ / $1$ & 50.4 $\pm$ 11.0 & 24.3 $\pm$ 3.1 & 45.2 $\pm$ 6.9 & \underline{30.9 $\pm$ 5.3} & 10.0 $\pm$ 0.0 & 74.0 $\pm$ 2.7 & \underline{9.8 $\pm$ 0.3} \\
\bottomrule
\end{tabular}
\end{table*}

\begin{table}[t]
\centering
\caption{Performance comparison of sparsity mask formations, illustrating average test accuracy over all eight robust aggregators and the test accuracy against the best-performing robust aggregator. The best-performing mask formations (i.e., the one yielding the lowest test accuracy) are highlighted with \textbf{bold}, and the second-best ones are \underline{underlined}.}
\label{tab:mask_performance}
\resizebox{\columnwidth}{!}{%
\begin{tabular}{@{}ccccc@{}}
\toprule
\textbf{Sparsity Mask} & \multicolumn{2}{c}{\textbf{IID}} & \multicolumn{2}{c}{\textbf{non-IID}} \\
\midrule
 & \textbf{Average} & \textbf{Highest} & \textbf{Average} & \textbf{Highest} \\
\midrule

$\Pi_{r}(\delta_1)$         & 65.97 & \underline{36.7} & 40.39 & 24.1  \\
$\Pi_{r}^{+}(\delta_1)$     & 61.55 & 45.3  & \underline{20.86} & 10.1  \\
$\Pi_{ERK}^{+}(\delta_1)$   & \underline{54.51} & 26.7  & \textbf{\textcolor{red}{12.85}} & \textbf{\textcolor{red}{9.4}} \\
$\Pi_{F}^{+}(\delta_1)$     & \textbf{\textcolor{red}{48.85}} & \textbf{\textcolor{red}{15.5 }} & 26.48 & \underline{10 }\\

\midrule
$\Pi_{r}(\delta_2)$         & \textbf{60.85} & \textbf{23.47 } & 44.41 & \underline{27.44 } \\
$\Pi_{r}^{+}(\delta_2)$     & 64.40 & \underline{33.22 } & \underline{40.96} & \textbf{20.21 }\\
$\Pi_{ERK}^{+}(\delta_2)$   & 66.32 & 50.12 & 45.27 & 28.18 \\
$\Pi_{F}^{+}(\delta_2)$     & \underline{62.68} & 43.56  & \textbf{39.12} & 32.08  \\

\midrule
$\Pi_{r}(\delta_3)$         & \underline{64.26} & \underline{36.3 } & \underline{36.19} & \textbf{9.4 } \\
$\Pi_{r}^{+}(\delta_3)$     & \textbf{56.29} & \textbf{32} & \textbf{16.44} & \underline{10 }\\
$\Pi_{ERK}^{+}(\delta_3)$   & 64.39 & 47.1  & 44.66 & 27.9  \\
$\Pi_{F}^{+}(\delta_3)$     & 66.84 & 52.02  & 41.71 & 29.12  \\

\midrule
$\Pi_{r}(\delta_4)$         & \textbf{61.26} & \textbf{39.10 } & \underline{33.68} & \underline{24.98 }\\
$\Pi_{r}^{+}(\delta_4)$     & 63.04 & 48.61  & \textbf{32.23} & \textbf{23.13 } \\
$\Pi_{ERK}^{+}(\delta_4)$   & 63.07 & \underline{46.53}  & 60.63 & 33.93 \\
$\Pi_{F}^{+}(\delta_4)$     & \underline{62.99} & 48.91  & 38.63 & 26.89  \\

\midrule
$\Pi_{r}(\delta_5)$         & \underline{57.12} & \underline{40.2 } & \underline{26.41} & \textbf{10} \\
$\Pi_{r}^{+}(\delta_5)$     & \textbf{49.58} & \textbf{26.8 } & \textbf{18.95} & \textbf{10}  \\
$\Pi_{ERK}^{+}(\delta_5)$   & 60.83 & 50.7   & 47.11 & \underline{26.2} \\
$\Pi_{F}^{+}(\delta_5)$     & 63.73 & 49.29 & 35.49 & 27.29  \\

\midrule
$\Pi_{r}(\delta_6)$         & \underline{50.14} & \underline{28.2} & \underline{27.65} & \textbf{10 } \\
$\Pi_{r}^{+}(\delta_6)$     & \textbf{45.57} & \textbf{27.6 } & \textbf{21.25} & \textbf{10 } \\
$\Pi_{ERK}^{+}(\delta_6)$   & 64.36 & 49.2  & 41.66 & \underline{20.7} \\
$\Pi_{F}^{+}(\delta_6)$     & 60.91 & 44.91 & 28.17 & 23.30  \\

\bottomrule
\end{tabular}
}
\end{table}

\indent To address the first question, we consider three different approaches to generate the sparse mask: (i) random sparsity, (ii) ERK-based random sparsity, and (iii) FORCE. For random sparsity, we examine two variants. The first, denoted by $\Pi_{r}(\cdot)$, randomly selects nonzero positions for each layer according to a given sparsity ratio. The second, denoted by $\Pi^{+}_{r}(\cdot)$, first identifies the critical, unpruned (learnable) parameters \cite{prune_force} and sets the corresponding mask entries to be nonzero; then selects the remaining nonzero positions at random for each layer to match the target sparsity ratio. For ERK-based random sparsity, instead of employing the same sparsity ratio for each layer, the ratio for each layer is determined according to the {\em Erdos-Renyi-Kernel (ERK)} formulation \cite{prune6}. Finally, FORCE employs an iterative pruning framework to identify the nonzero positions in the sparsity mask.

In our numerical analysis, we consider three sparsity regimes, each with two sparsity ratios: high sparsity $\delta_{1}=5\times 10^{-3}$ and $\delta_{2}=1\times 10^{-2}$, moderate sparsity $\delta_{3}=5\times 10^{-2}$ and $\delta_{4}=1\times 10^{-1}$, and finally low sparsity, $\delta_{5}=2\times 10^{-1}$ and $\delta_{6}=5\times 10^{-1}$.

Following our main simulation setup in Section~\ref{subsec:sim_setup}, we consider an image classification task on the CIFAR-10 dataset ~\cite{cifar10}, and train a ResNet-20 model~\cite{resnet}. In line with the common practice ~\cite{CC,rop}, we train ResNet-20 for 100 epochs with a local batch size of 32 and an initial learning rate of $\eta=0.1$, which is reduced at epoch 75 by a factor of $0.1$. Finally, we set the local momentum parameter to $\beta=0.9$. We assume that the Byzantines can collude to share their local data with each other. They can access or predict benign model updates shared with the PS, while remaining oblivious to the local datasets of benign clients. Further, following prior works \cite{CC,rop,ndss2022_Attack}, we set Byzantine ratio $\gamma=0.2$, and set the number of clients to $k=25$, of which $k_{m}=5$ are malicious. 

We consider eight different robust aggregators, namely, Bulyan \cite{Bulyan}, CC \cite{CC}, CM \cite{Trimmed_mean}, M-Krum \cite{Krum}, RFA \cite{RFA}, and TM \cite{Trimmed_mean} and GAS \cite{GAS} with two variations built on Bulyan and M-Krum, respectively. All experiments are conducted under both IID and non-IID (following the Dirichlet distribution $\mathrm{Dir}(\alpha)$ with $\alpha=1$) dataset distributions. The corresponding results for IID and Non-IID data distributions are illustrated in Table~\ref{tab:IID-otherPrunes} and Table~\ref{tab:nonIID}, respectively.

One immediate observation from Table~\ref{tab:IID-otherPrunes} and Table~\ref{tab:nonIID} is that when the critical layers of the NN architecture are taken into account rather than employing a purely random sparse mask $\Pi_{r}(\cdot)$, the proposed HSA has a higher impact on model convergence, with an average $5-10\%$ reduction in the test accuracy. The impact of the critical layers is more pronounced at high sparsity levels, since in the case of low and moderate sparsity, the probability of critical layer weights being already selected in a binary random sparse mask is higher.

Another interesting and counterintuitive observation from Table~\ref{tab:IID-otherPrunes} and Table~\ref{tab:nonIID} is that, ERK-based random mask $\Pi_{ERK}(\cdot)$, which partially takes the network topology into account, underperforms relative to random sparse masks. The overall effectiveness of these different mask formations is summarized in the Table~\ref{tab:mask_performance}.

\subsection{Sparsity Distribution over Network Layers}
To address the second and third research questions on the importance of layer-wise distribution and the optimal sparsity ratio for the strongest attack, we modify the iterative network pruning strategy by introducing additional sparsity constraints for each layer. We consider two strategies: (i) imposing a fixed maximum sparsity constraint on each layer to prevent accumulation of non-zero positions, and (ii) identifying layers that typically retain high density of non-zero positions after pruning (e.g., the first convolutional layer and fully connected layers) and applying additional sparsity constraints specifically to these layers. In Figure \ref{fig:layers_prune_comperison}, we illustrated how FORCE and ERK based random pruning overemphasized certain layers, particularly first convolutional layer and the penultimate FC layer, the figure also illustrates how layer-wise constraint helps to redistribute non-zero weight more uniformly.
In Appendix \ref{sec:figsForce}, in Figs, \ref{fig:layers_prune_comperison005}, \ref{fig:layers_prune_comperison01}, \ref{fig:layers_prune_comperison05}, \ref{fig:layers_prune_comperison1},
we illustrate the remaining weights for layers of the ResNet-20 architecture for $\delta=5\times10^-3$, $\delta=1\times10^-2$, $\delta=5\times10^-2$ ,$\delta=1\times10^-1$ global sparsity regimes respectively, using FORCE with specific layer constraints.

\subsubsection{Simulation Results for HSA}
The results in Table~\ref{tab:ForceAblationsIID} for IID scenarios, as well as Table~\ref{tab:forceAblationsDir} and Table~\ref{tab:LayerConstraintsDIR} for non-IID scenarios. indicate that both approaches improve training stability, evidenced by smaller standard deviations and higher accuracy values across attack types. Furthermore, the layer-specific constraint consistently achieves higher success rates and smoother convergence, as reflected by lower standard deviations in Table~\ref{tab:forceAblationsDir}, making the improvement particularly evident in the non-IID case. These results emphasize the importance of maintaining critical layers when datasets are heterogeneous.

The impact of layer-wise sparsity constraints is demonstrated in Table~\ref{tab:LayerConstraintsIID}, where constraining the first and last layers ($\delta^{max}_{cw1}$, $\delta^{max}_{fc}$) significantly improves attack success rate across all sparsity regimes and aggregators. Examining different global sparsity levels reveals several key observations. First, at high sparsity ($\delta=5\times10^{-3}$), constraining the last layer to 0.4 while allowing flexibility in the first layer ($\delta^{max}_{cw1}=1$) yields strong results, for TM, achieving 50.1\% compared to 70.4\% with full flexibility corresponding to a 29\% improvement. Second, at moderate sparsity ($\delta=5\times10^{-2}$), dual constraints on both layers prove most effective, where RFA achieves 41.7\% with ($\delta^{max}_{cw1}=0.2, \delta^{max}_{fc}=0.4$) compared to 54.1\% with no constraints, while CC achieves its best overall performance of 34.6\% with ($\delta^{max}_{cw1}=1, \delta^{max}_{fc}=1$). Third, at higher sparsity ratios ($\delta=1\times10^{-1}$), the benefits become even more pronounced; M-Krum drops from 66.8\% to 50.2\%, TM from 40.8\% to 36.6\%, and Bulyan from 43.4\% to 24.1\% when layer-wise constraints are applied. Notably, constraining the first layer alone ($\delta^{max}_{cw1}=0.2$ or $0.4$) while leaving the last layer unconstrained ($\delta^{max}_{fc}=1$) generally yields suboptimal results, suggesting that both boundary layers play critical roles in defence evasion.

Finally, to address the third question, examining the impact of uniform layer constraints ($\delta^{max}$) reveals critical insights across both IID and non-IID settings. In the IID case (Table~\ref{tab:ForceAblationsIID}), tighter constraints significantly enhance attack effectiveness: at $\delta=5\times10^{-3}$, applying $\delta^{max}=0.25$ reduces test accuracy to 31.3\% (Bulyan), 33.9\% (CC), and 35.7\% (CM) compared to 40.6\%, 39.6\%, and 39.5\% with no constraint ($\delta^{max}=1$), representing improvements of 23\%, 14\%, and 10\% respectively. This trend intensifies at higher sparsity: at $\delta=1\times10^{-1}$, constraints yield 25.8\% (Bulyan) versus 55.4\% unconstrained, a remarkable 53\% improvement. The non-IID results (Table~\ref{tab:forceAblationsDir}) demonstrate even more dramatic effects: at $\delta=5\times10^{-2}$, constraining to $\delta^{max}=0.25$ achieves 14.0\% (Bulyan), 20.8\% (CC), and 10\% (RFA and TM) compared to 34.2\%, 28.0\%, and 45.2\% without constraints, improvements of 59\%, 26\%, and 78\%, respectively. At $\delta=1\times10^{-1}$, constraints produce near-complete model collapse with accuracies reaching 10\% (random guessing for 10 classes) across multiple aggregators. Notably, M-Krum exhibits vulnerability at more sparse regimes in both settings (15.5\% at $\delta=5\times10^{-3}$ in IID, 21.4\% at $\delta=1\times10^{-2}$ in non-IID), while GAS demonstrates superior robustness, maintaining 52.5-74.8\% accuracy in IID even under tight constraints. These findings underscore that controlling layer-wise sparsity distribution is crucial for attack success, with benefits amplified in heterogeneous data scenarios.

\subsubsection{Simulation results for DHSA}

We illustrate the performance of DHSA with certain layer constraints in (Tables~\ref{tab:LayerConstraintsIID-DHSA} and \ref{tab:LayerConstraintsDIR-DHSA}). In the IID setting, imposing tight layer constraints significantly amplifies attack effectiveness. For Bulyan, constraining both the first convolutional layer and final FC layer to $\delta^{max}_{cw1} = \delta^{max}_{fc} = 0.2$ at $\delta = 1 \times 10^{-1}$ reduces accuracy from 52.5\% (unconstrained, $\delta^{max} = 1$) to 31.4\%, a 21.1 percentage point degradation. Similarly, CC suffers a 12.4 percentage point drop (from 50.5\% to 38.1\%) under the same constraints at $\delta = 1 \times 10^{-1}$, $\delta^{max}_{cw1} = 1$, $\delta^{max}_{fc} = 0.2$. For TM, constraining the first layer at $\delta = 5 \times 10^{-2}$ reduces accuracy by 23.5 percentage points (57.6\% to 34.1\%). In contrast, M-Krum maintains relative robustness, with worst-case accuracy at 42.2\% under $\delta = 5 \times 10^{-3}$, $\delta^{max}_{cw1} = 0.4$, $\delta^{max}_{fc} = 0.2$, representing only an 18.0 percentage point drop from its unconstrained baseline (60.2\%).

The non-IID setting exposes catastrophic vulnerabilities with layer constraints inducing complete model collapse. TM achieves exactly 10.0\% $\pm$ 0.0\% across all 36 sparsity configurations, representing a 46.8-66.8 percentage point degradation from typical benign performance. For RFA, accuracy plummets from 47.0\% (at $\delta = 5 \times 10^{-3}$, $\delta^{max}_{cw1} = 1$, $\delta^{max}_{fc} = 0.2$) to 10.0\% when both constraints tighten to 0.2, a 37.0 percentage point collapse. Bulyan experiences dramatic instability, at $\delta = 5 \times 10^{-3}$, relaxing $\delta^{max}_{cw1}$ from 0.4 to 1 while maintaining $\delta^{max}_{fc} = 0.4$ reduces accuracy from 53.6\% to 47.2\%, but tightening both to 0.2 drives it further down to 35.5\%. CC shows sensitivity to first-layer constraints, with accuracy improving from 16.4\% to 27.4\% when $\delta^{max}_{cw1}$ increases from 0.4 to 1 at $\delta = 5 \times 10^{-3}$, $\delta^{max}_{fc} = 0.2$. M-Krum exhibits the strongest resilience, maintaining 42.4\%-60.9\% accuracy in the low sparsity regime ($\delta = 5 \times 10^{-3}$), but still degrades to 14.5\%-33.6\% at $\delta = 1 \times 10^{-1}$, demonstrating that aggregators cannot withstand extreme sparsity under data heterogeneity. 

Lastly, standard deviations in non-IID settings reach $\pm 25.0$, indicating an instability due to the optimization process, compared to $\pm 1-13$ in IID settings. We address this issue later in the Appendix \ref{sec:IQR}.

\section{Further Ablations on Cascaded Robust Aggregators} \label{app:AggregatorCombs}

In Section \ref{s:back}, through extensive numerical experiments, we have shown that existing robust aggregators are often effective against particular type of Byzantines. To elucidate,  TM is tailored to identify index-wise statistical anomalies, whereas M-Krum is designed to detect and eliminate outliers based on norm distance. Consequently, while TM is more resilient to local attacks with strong perturbations, it is less effective against attacks with smaller but consistent perturbations as in ALIE. However, these small perturbations may accumulate over indices, hence, can be detected by a defence mechanism leveraging Euclidean-wise distance as an anomaly detection metric. This is aligned with M-Krum's effectiveness against ALIE. At this point one can raise the question whether a universal defence method can be formed by sequentially combining two different robust aggregators, each employing a different metric for anomaly detection.

Motivated by this, in this section, our objective is to answer this question by conducting extensive numerical experiments and analysing the performances of different sequentially combined robust aggregator pairs. We want to remark that the Bulyan method is an example of a sequential aggregation/sanitization framework. Hence, we start our analysis by revisiting the Bulyan method.

\textbf{Revisiting Bulyan:} Bulyan \cite{Bulyan} method first executes M-Krum \cite{Krum} for client selection/outlier elimination then executes TM \cite{Trimmed_mean} for the aggregation. Numerical experiments in Section \ref{sec:exp_results} indicates that the Bulyan method does not consistently outperform both M-Krum and TM. It can even underperform compared to M-Krum and TM. To clarify, against ALIE, in the IID split illustrated in Table \ref{tab:Main-IID}, Bulyan achieves $\% 9$ and $ \% 25$ lower test accuracy compared to M-Krum and TM, respectively. Similarly in the non-IID split illustrated in Table \ref{tab:nonIID-main}, Bulyan achieves $\% 15$ and $ \% 8$ lower test accuracy compared to M-Krum and TM, respectively.\\
\indent We observe that, for both IID and non-IID splits, Bulyan outperforms TM under the Min-Max attack, which uses larger perturbations than Min-Sum and ALIE. This is expected since, due to the larger adversarial perturbations, Bulyan’s first stage can successfully eliminate outliers via selection mechanism of  M-Krum, which is also consistent with the strong performance of M-Krum itself against Min-Max. Excluding HSA variants, M-Krum either outperforms or matches Bulyan's performance. However, under HSA and its variants, M-Krum is completely evaded in both IID and non-IID settings, whereas Bulyan maintains certain robustness through the coordinate-wise sanitization mechanism inherent in TM. 
\begin{remark}
   Bulyan does not provide consistent improvement over TM and M-Krum individually, but mitigates worst-case failures when Byzantines completely evade either component. Specifically, Bulyan offers more stable performance by enhancing the worst-case accuracy against the most effective Byzantine attacks, preventing complete defence collapse.
\end{remark} 
A critical observation from Tables \ref{tab:Main-IID} and \ref{tab:nonIID-main} reveals that Bulyan underperforms both TM and M-Krum against ALIE and Min-Sum attacks. We recall that these attacks craft malicious updates to be statistically and geometrically more close to the benign updates. Hence, we argue that the reason behind the performance degradation is due to the false elimination of the benign ones through the stages of sanitization, thus in contrary to the design objective, the Byzantine-to-benign ratio increases over the stages. In other words, the client selection mechanism orchestrated according to M-Krum eliminates benign clients rather than the malicious ones, which degrades the performance of later TM execution.\\
\indent  To verify this argument, we consider two performance metrics, namely, escape and impact ratios, and we compare the performance of M-Krum, TM, and Bulyan. The results are illustrated in Table \ref{tab:Bulyan-Escape},  which clearly shows that all attacks, except Min-Max, perfectly evade  M-Krum, that is, all the Byzantines pass the elimination stage while the beings ones are falsely eliminated, which implies $\%100$ escape ratio, and increase the impact ratio from $\%20$ to $\%28$. We also observe a similar trend in TM, however, the most important observation from  Table \ref{tab:Bulyan-Escape} is that compared to TM, impact ratio of Bulyan approximately doubles for ALIE, HSA, and Min-Ma,x since in the first stage of the Bulyan, M-Krum based client selection eliminates benign clients. On the other hand, in the case of Min-Max, on average, M-Krum is able to eliminate $\%40$ of the Byzantines, and able to reduce the impact ratio to $\%13$, which also explains the performance of Bulyan against Min-Max where the impact ratio is  only $\%4$ and a higher accuracy compared to both TM and M-Krum is achieved.
\begin{table*}[t]
\centering
\caption{Illustration of escape ratio, impact ratio and achieved accuracy of ALIE, HSA, Min-Sum, and Min-Max under robust aggregators M-Krum, TM, and Bulyan on CIFAR-10 classification task. For the client configuration ,we consider $k=25$ clients, $k_{m}=5$ of them being malicious. (Natural impact ratio of \% 20)}
\label{tab:Bulyan-Escape}
\resizebox{\textwidth}{!}{%
\begin{tabular}{cccc|ccc|ccc|}
\hline
\multirow{2}{*}{\textbf{ATK}} & \multicolumn{3}{c|}{\textbf{M-Krum}} & \multicolumn{3}{c|}{\textbf{TM}} & \multicolumn{3}{c|}{\textbf{Bulyan}} \\ \cline{2-10} 
 & \textbf{Escape \%} & \textbf{Impact \%} & \textbf{Acc. \%} & \textbf{Escape \%} & \textbf{Impact \%} & \textbf{Acc. \%} & \textbf{Escape \%} & \textbf{Impact \%} & \textbf{Acc. \%} \\ \hline
ALIE & 100 & 28 & 55.8 & 100 & 33 & 72.8 & 88 & 73 & 47.1 \\
HSA & 100 & 28 & 16.3 & 99.7 & 33 & 33.4 & 87 & 70 & 31.3 \\
Min-Sum & 100 & 28 & 55.1 & 84 & 28 & 47.5 & 73 & 48 & 41.1 \\
Min-Max & 46 & 13 & 61 & 37 & 12 & 45 & 6 & \textbf{4} & 69.4 \\ \hline
\end{tabular}%
}
\end{table*}

\begin{table}[h!]
\caption{Performance comparison of M-Krum+CliM and CliM+M-Krum hybrid aggregators against M-Krum+TM (Bulyan) on CIFAR-10 classification task in both IID and non-IID settings. Values show test accuracy differences relative to the Bulyan baseline.}
\label{tab:CliMBulyan}
\resizebox{\columnwidth}{!}{%
\begin{tabular}{@{}llccc@{}}
\toprule
 & \multicolumn{1}{c}{\textbf{Accuracy}} & \textbf{Attack} & \textbf{Hybrid Aggregators} & \textbf{Bulyan} \\ \midrule
\multirow{8}{*}{IID} & 72.61 $\pm$ 1.74 & ALIE & M-Krum+CliM & \textcolor{blue}{+25.51} \\
 & 79.84 $\pm$ 0.21 & ALIE & CliM+M-Krum & \textcolor{blue}{+32.74} \\ \cmidrule(l){2-5} 
 & 41.16 $\pm$ 1.9 & HSA & M-Krum+CliM & \textcolor{blue}{+9.86} \\
 & 53.94 $\pm$ 10.4 & HSA & CliM+M-Krum & \textcolor{blue}{+21.73} \\ \cmidrule(l){2-5} 
 & 48.09 $\pm$ 1.18 & Min-Sum & M-Krum+CliM & \textcolor{blue}{+6.99} \\
 & 82.91 $\pm$ 0.16 & Min-Sum & CliM+M-Krum & \textcolor{blue}{+41.81} \\ \cmidrule(l){2-5} 
 & \multicolumn{1}{c}{50} & Min-Max & M-Krum+CliM & \textcolor{red}{-17.2} \\
 & 70.03 $\pm$ 3.29 & Min-Max & CliM+M-Krum & \textcolor{blue}{+2.83} \\ \midrule
\multirow{8}{*}{non-IID} & \multicolumn{1}{c}{55.41 $\pm$ 9.17} & ALIE & M-Krum+CliM & \textcolor{blue}{+20.71} \\
 & \multicolumn{1}{c}{30.44 $\pm$ 4.3} & ALIE & CliM+M-Krum & \textcolor{red}{-4.26} \\ \cmidrule(l){2-5} 
 & 18.5 $\pm$ 5.8 & HSA & M-Krum+CliM & \textcolor{blue}{+8.6} \\
 & 64.81 $\pm$ 0.2 & HSA & CliM+M-Krum & \textcolor{blue}{+54.91} \\ \cmidrule(l){2-5} 
 & \multicolumn{1}{c}{38.08 $\pm$ 1.91} & Min-Sum & M-Krum+CliM & \textcolor{blue}{+16.48} \\
 & \multicolumn{1}{c}{52.13 $\pm$ 1.55} & Min-Sum & CliM+M-Krum & \textcolor{blue}{+30.53} \\ \cmidrule(l){2-5} 
 & \multicolumn{1}{c}{82.58 $\pm$ 0.28} & Min-Max & M-Krum+CliM & \textcolor{blue}{+0.38} \\
 & \multicolumn{1}{c}{44.66 $\pm$ 5.52} & Min-Max & CliM+M-Krum & \textcolor{red}{-37.54} \\ \bottomrule
\end{tabular}%
}
\end{table}
\begin{remark}
    The sequential execution of M-Krum and TM, as proposed in Bulyan,  may lead to a performance loss compared to marginal execution of each defence mechanism, due to false elimination of the benign (non-Byzantine) values and undesired increase in the impact ratio.
\end{remark}
We have shown that robust aggregators that are designed to expose and eliminate malicious values may suffer from false elimination of benign values, especially when they are executed sequentially, as in the case of Bulyan. Hence, we propose to replace TM with its alternative version where, instead of eliminating suspicious values, an index-wise sanitization, through clipping, is performed as illustrated in Algorithm~\ref{code:clim}.
\begin{algorithm}[t]
    \small
    \caption{CliM: Cliped Mean}
    \label{code:clim}
    \begin{algorithmic}[1]
     \State \textbf{Inputs:} Client updates $\{\mathbf{m}_1, \mathbf{m}_2, \ldots, \mathbf{m}_n\}$
     \For{each coordinate $i = 1, 2, \ldots, d$}
     \State $\mathcal{M}_{j} =\left\{ \mathbf{m}_{k}[i]: k\in \mathcal{K}\right\} $
     \State $\mathcal{V}_{i}= f_{sort-eliminate}(\mathcal{M}_{j},k_m)$
     \EndFor
     \For{each client $k\in \mathcal{K}$}
        \For{each coordinate $i = 1, 2, \ldots, d$}
        \State $\hat{\mathbf{m}}_{k}[i] = \max(\min(\mathbf{m}_{k}[i],\max(\mathcal{V}_{i})),\min(\mathcal{V}_{i}))$
        \EndFor
     \EndFor
     \State \textbf{return} $\{\hat{\mathbf{m}}_1, \hat{\mathbf{m}}_2, \ldots, \hat{\mathbf{m}}_n\}$
    \end{algorithmic}
\end{algorithm}
After introducing CliM, we analyse the performance of the sequential execution of CliM and M-Krum, and the results are illustrated in Table \ref{tab:CliMBulyan}. Here, we remark that CliM can also be employed for an index-wise sanitization method, thus, it can be executed before M-Krum.

On non-IID scenarios, If we employ CliM as the last aggregator similar to the TM, we can achieve performance gain against all attacks. Furthermore, on ALIE and Min-Sum, which have the highest \% impact in the Bulyan, we can improve the test accuracy by \% 20, nearly doubling the accuracy of the Bulyan.

\textbf{Cascaded aggregators with CC:} For a more extensive analysis, we designed a setup involving three prominent robust aggregation methods: CC, M-Krum, and TM, each employing a distinct sanitization/ outlier-detection strategy. CC performs sanitization by projecting model updates onto a $\tau$-ball centered around a reference point, ensuring that updates do not deviate arbitrarily far from it. M-Krum/M-KRUM also utilizes a Euclidean distance-based anomaly detection but relies on pairwise distances to identify outliers rather than a single reference point. Finally, TM conducts an index-wise investigation instead of Euclidean distance, assuming the attacks are not visible globally but locally. We consider 4 different hybrid aggregators namely, $CC+TM$, $M-Krum+CC$, $CC+M-Krum$, and finally $M-Krum+CC+TM$. These combinations are selected to observe the interactions between different defence mechanisms: history-based sanitization by projection (CC), euclidean distance based elimination among the clients (M-Krum), and coordinate-wise (TM) elimination.\\
\indent In our analysis, we consider three of the most potent attacks: ALIE, Min-Sum, and Min-Max. It is worth noting that all three attacks are built upon the same foundational strategy of adding scaled index-wise standard deviations as perturbations to a reference model update; they differ primarily in how the scaling coefficients are determined, which seeks a trade-off between imperceptibility and the strength.\\
\indent The results in Table \ref{tab:Hybrid-aggregation} clearly indicate that a naive sequential combination of robust aggregators does not guarantee improved robustness across different attack types; for instance, the M-Krum+CC combination is highly effective against the Min-Max attack but performs poorly against the Min-Sum attack, highlighting the lack of a universally robust hybrid defence. Furthermore, the most important observation is that no hybrid aggregator consistently outperforms all of its constituent aggregators when used individually.\\
\indent Similar to our analyses for Bulyan, we observe that cascaded solutions are often only helps to mitigate the worst-case performance of the underlying individual robust aggregators, to elucidate cascaded solution $M-Krum + CC $ enhance the performance of CC against Min-Max but does not outperform M-Krum, in a similar way, cascaded solution $CC + M-Krum $ offers a better robustness against Min-Sum compared to  M-Krum but underperforms CC.\\
\indent Another important observation is that in the case of cascaded defenses, the use of CC at the initial stage may make the attack imperceptible but effective rather than sanitizing it, which make the subsequent defense strategy less effective against Byzantines. To elucidate, as illustrated in Table \ref{tab:Hybrid-aggregation}, applying CC before Multi-Krum reduces accuracy by $\% 44$  and $\% 14$ compared to using Multi-Krum alone, against the Min-Max attack and ALIE, respectively.
\begin{remark}
    The order of aggregation methodologies is critical for the identification and neutralization of malicious updates. Employing a norm-based sanitation method such as CC as a first-stage defense can inadvertently shield malicious updates from subsequent norm-based elimination methods like Multi-Krum, thereby diminishing their effectiveness.
\end{remark}

\begin{table}[h!]
\centering
\caption{Performance evaluation of hybrid aggregation defenses against ALIE and HSA attacks on CIFAR-10 non-IID classification task using ResNet-20. Difference columns show performance degradation relative to standalone aggregators: M-Krum CC and TM. Blue (positive) values indicate hybrid defense performs better (higher accuracy) than the individual aggregator; red (negative) values indicate worse performance.}
\label{tab:Hybrid-aggregation}
\resizebox{\columnwidth}{!}{%
\begin{tabular}{@{}cccccc@{}}
\toprule
\textbf{Accuracy} & \textbf{Attack} & \textbf{Hybrid aggregation} & \textbf{M-Krum-Dif} & \textbf{CC-Dif} & \textbf{TM-Dif} \\ \midrule
54.66 $\pm$ 2.42  & ALIE            & M-Krum+CC                     & \textcolor{blue}{+4.96} & \textcolor{red}{-5.04} & - \\
53.02 $\pm$ 0.08  & ALIE            & CC+TM                       & - & \textcolor{red}{-6.68} & \textcolor{blue}{+10.92} \\
41.9 $\pm$ 0.66   & ALIE            & M-Krum+CC+TM                  & \textcolor{red}{-7.8} & \textcolor{red}{-17.8} & \textcolor{red}{-0.2} \\
35.94 $\pm$ 1.14  & ALIE            & CC+ M-Krum                     & \textcolor{red}{-13.76} & \textcolor{red}{-23.76} & - \\ \hline
17.73 $\pm$ 4.46  & HSA            & M-Krum+CC                     & \textcolor{blue}{+7.93} & \textcolor{blue}{+0.13} & - \\
12.64 $\pm$ 0.12  & HSA            & CC+TM                       & - & \textcolor{red}{-4.96} & \textcolor{blue}{+2.64} \\
11.72 $\pm$ 1.36   & HSA            & M-Krum+CC+TM                  & \textcolor{blue}{+1.92} & \textcolor{red}{-5.88} & \textcolor{blue}{+1.72} \\
20.63 $\pm$ 3.06  & HSA            & CC+ M-Krum                     & \textcolor{blue}{+10.83} & \textcolor{blue}{+3.03} & - \\ \hline
85.29 $\pm$ 0.02  & Min-Max          & M-Krum+CC                     & \textcolor{red}{-1.61} & \textcolor{blue}{+51.29} & - \\
83.6 $\pm$ 0.44   & Min-Max          & M-Krum+CC+TM                  & \textcolor{red}{-3.3} & \textcolor{blue}{+49.6} & \textcolor{blue}{+65.6} \\
43.04 $\pm$ 1.23  & Min-Max          & CC+ M-Krum                     & \textcolor{red}{-43.86} & \textcolor{blue}{+9.04} & - \\
26.76 $\pm$ 0.4   & Min-Max          & CC+TM                       & - & \textcolor{red}{-7.24} & \textcolor{blue}{+8.76} \\ \hline
35.74 $\pm$ 1.55  & Min-Sum          & CC+ M-Krum                     & \textcolor{blue}{+8.54} & \textcolor{red}{-2.36} & - \\
33.6 $\pm$ 1.16   & Min-Sum          & CC+TM                       & - & \textcolor{red}{-4.5} & \textcolor{blue}{+1.1} \\
30.98 $\pm$ 0.3   & Min-Sum          & M-Krum+CC                     & \textcolor{blue}{+3.78} & \textcolor{red}{-7.12} & - \\
22.02 $\pm$ 0.95  & Min-Sum          & M-Krum+CC+TM                  & \textcolor{red}{-5.18} & \textcolor{red}{-16.08} & \textcolor{red}{-10.48} \\ \bottomrule
\end{tabular}
}
\end{table}

\begin{table*}[h!]
\centering
\caption{Accuracy and impact ratios of ALIE, HSA, Min-Sum, and Min-Max for CIFAR-10 classification task. CIFAR-10 dataset distributed IID over $k=25$ clients, $k_{m}=5$ of them being malicious. (Natural impact ratio of \% 20)}
\label{tab:ImpactRatios}
\resizebox{\textwidth}{!}{%
\begin{tabular}{@{}cccc|ccc|ccc|ccc|ccc@{}}
\toprule
\multirow{2}{*}{\textbf{ATK}} & \multicolumn{3}{c|}{\textbf{Krum}} & \multicolumn{3}{c|}{\textbf{TM}} & \multicolumn{3}{c|}{\textbf{Bulyan}} & \multicolumn{3}{c|}{\textbf{CC+KRUM}} & \multicolumn{3}{c}{\textbf{CC+TM}} \\ \cmidrule(l){2-16} 
 & \textbf{Escape \%} & \textbf{Impact \%} & \textbf{Acc. \%} & \textbf{Escape \%} & \textbf{Impact \%} & \textbf{Acc. \%} & \textbf{Escape \%} & \textbf{Impact \%} & \textbf{Acc. \%} & \multicolumn{1}{c}{\textbf{Escape \%}} & \multicolumn{1}{c}{\textbf{Impact \%}} & \multicolumn{1}{c|}{\textbf{Acc. \%}} & \multicolumn{1}{c}{\textbf{Escape \%}} & \multicolumn{1}{c}{\textbf{Impact \%}} & \multicolumn{1}{c}{\textbf{Acc. \%}} \\ \midrule
ALIE & 100 & 28 & 55.8 & 100 & 33 & 72.8 & 88 & 73 & 47.1 & 100 & 28 & 70.66 & 99 & 33 & 69.26 \\
HSA & 100 & 28 & 16.3 & 99.7 & 33 & 33.4 & 87 & 70 & 31.3 & 89 & 25 & 30.86 & 98 & 32 & 38.4 \\
Min-Sum & 100 & 28 & 55.1 & 84 & 28 & 47.5 & 73 & 48 & 41.1 & 98 & 27 & 44.22 & 76 & 25 & 45.35 \\
Min-Max & 46 & 13 & 61 & 37 & 12 & 45 & 6 & \textbf{4} & 69.4 & 52 & 15 & 53.22 & 32 & 11 & 44.25 \\ \bottomrule
\end{tabular}%
}
\end{table*}

\begin{table}[h!]
\centering
\caption{Performance comparison of CliM-based hybrid aggregators versus vanilla TM-based combinations on CIFAR-10 non-IID classification task. Values indicate test accuracy improvements (blue) or degradations (red) relative to M-Krum+TM (Bulyan) and CC+TM baselines under ALIE, Min-Max, and Min-Sum attacks.}
\label{tab:TMcapped}
\resizebox{\columnwidth}{!}{%
\begin{tabular}{@{}ccccc@{}}
\toprule
\textbf{Accuracy} & \textbf{Attack} & \textbf{hybrid Aggregators} & \textbf{Krum+TM (Bulyan)} & \textbf{CC+TM} \\ \midrule
55.96 $\pm$ 1.36 & ALIE & CC+CliM & {--} & {\color{blue} +2.94} \\
55.41 $\pm$ 9.17 & ALIE & M-Krum+CliM & {\color{blue} +20.71} & {--} \\
43.46 $\pm$ 24.28 & ALIE & CliM+CC & {--} & {\color{red} {--}9.56} \\
30.44 $\pm$ 4.3 & ALIE & CliM+Krum & {\color{red} {--}4.26} & {--} \\
{--} & ALIE & {--} & 34.7 & 53.02 \\ \midrule
82.58 $\pm$ 0.28 & Min-Max & M-Krum+CliM & {\color{blue} +0.38} & {--} \\
44.66 $\pm$ 5.52 & Min-Max & CliM+Krum & {\color{red} {--}37.54} & {--} \\
27.95 $\pm$ 2.75 & Min-Max & CC+CliM & {--} & {\color{blue} +1.19} \\
27.77 $\pm$ 0.47 & Min-Max & CliM+CC & {--} & {\color{blue} +1.01} \\
{--} & Min-Max & {--} & 82.2 & 26.76 \\ \midrule
54.24 $\pm$ 0.68 & Min-Sum & CliM+CC & {--} & {\color{blue} +20.64} \\
52.13 $\pm$ 1.55 & Min-Sum & CliM+Krum & {\color{blue} +30.53} & {--} \\
38.08 $\pm$ 1.91 & Min-Sum & M-Krum+CliM & {\color{blue} +16.48} & {--} \\
33.02 $\pm$ 0.96 & Min-Sum & CC+CliM & {--} & {\color{red} {--}0.58} \\
{--} & Min-Sum & {--} & 21.6 & 33.6 \\ \bottomrule
\end{tabular}%
}
\end{table}

\section{Complexity and Running Time Analysis of aggregators} \label{app:Complexity}

In this section, we go over the runtime of the robust aggregator in practical scenarios. We experimented with $k=[25,50,100]$ total clients and calculated the average aggregation time on ResNet-20 architecture which has a total of $\sim7\times10^{5}$ trainable parameters ($d\cong 700K$). All simulations run on the NVIDIA A40 GPU and averaged over 1000 aggregations. We show those results in Table \ref{tab:aggr_times}. Out of all aggregators, FedAvg \cite{FedAVG1}, SignSGD \cite{SignSGD} CM, and TM \cite{Trimmed_mean} are considerably faster than all aggregators with asymptotic runtime of $O(dk)$. Following that, CC \cite{CC} and RFA and FL-Trust \cite{FLtrust} which also theoretically have asymptotic runtime of $O(dk)$. Finally, as expected, M-Krum and Bulyan are the slowest robust aggregators with an asymptotic runtime of $O(dk^2)$. Our simulations for the robust aggregators also show the expected linear/polynomial increase in the runtime as the $k$ increases. Theoretically, \cite{GAS} should achieve the same performance as the aggregator that is employed. However, due to the sub-optimal implementations, GAS with $p=1000$ considerably slows down the M-Krum and Bulyan.

\begin{table}[]
\centering
\caption{Average aggregation times of Byzantine-robust aggregators for varying client populations ($k=25$, $k=50$, $k=100$) on ResNet-20 architecture with $\sim 7 \times 10^{5}$ parameters, measured in milliseconds (ms). Results averaged over 1000 aggregations on NVIDIA A40 GPU.}
\label{tab:aggr_times}
\begin{tabular}{@{}cccc@{}}
\toprule
\textbf{AGGR} & \textbf{$k=25$} & \textbf{$k=50$} & \textbf{$k=100$} \\ \midrule
FedAVG \cite{FedAVG1} & 0.3 & 0.5 & 0.7 \\
Bulyan \cite{Bulyan} & 50 & 225 & 1073 \\
CC \cite{CC} & 2.3 & 4.5 & 8.9 \\
CM \cite{Trimmed_mean} & 0.8 & 1 & 2.6 \\
FL-Trust \cite{FLtrust} & 27 & 34 & 43 \\
GAS-Bulyan \cite{GAS} & 239 & 323 & 333 \\
GAS-Krum \cite{GAS} & 146 & 226 & 232 \\
M-Krum \cite{Krum} & 49 & 228 & 956 \\
RFA \cite{RFA} & 6.1 & 12.1 & 24.2 \\
SignSGD \cite{SignSGD} & 0.3 & 0.5 & 0.7 \\
TM \cite{Trimmed_mean} & 0.9 & 1.6 & 3.5 \\ \bottomrule
\end{tabular}
\end{table}

\section{Additional Numerical Results for Robust Aggregators. }\label{sec:ablations on Robust Aggregators}
In the scope of this work we mostly investigate the well known fundamental benchmark defence frameworks such as M-Krum, RFA, TM, CC, etc. However, in this section, we revisit some of the SoTA defence strategies such as FoundationFL \cite{FoundationFL} and LASA \cite{lasa}. Here, through counter-examples and detailed simulations we aim to show how the recent robust aggregation frameworks that utilize different strategies, such as sign investigation, sparsification gives a false sense of security due to the assumption that malicious clients follows the conventional Byzantine methods. In other words we will show how such defence mechanism are vulnerable to the attacks that are redesigned according to the proposed defence mechanism. We remark that, similarly, in \cite{rop}, the authors have shown that, although CC has been effective against existing Byzantine attacks, with certain simple modifications, CC can be easily evaded.\\
\textbf{LASA:} First, we will analyse LASA \cite{lasa} (Algorithm \ref{alg:lasa_defense}), which utilize model sparsification to mitigate Byzantine attacks. 
\begin{definition}[Positive Direction Purity]
For a vector $x \in \mathbb{R}^d$, the positive direction purity $\rho$ of $x$ is defined as
\begin{equation}
\rho := \frac{1}{2} \times \left( 1 + \frac{\sum_{i=1}^{d} \operatorname{sgn}([x]_i)}{\sum_{i=1}^{d} |\operatorname{sgn}([x]_i)|} \right), \quad 0 \le \rho \le 1,
\label{eq:pdp}
\end{equation}
where $\operatorname{sgn}(\cdot)$ is the function to take the sign of each element and $[ \cdot ]_i$ is the $i$-th coordinate of a vector.
\end{definition}

\begin{definition}[MZ-score]
For a set of values $X := \{x_1, \dots, x_n\}$ with median $\operatorname{Med}(X)$ and standard deviation $\sigma$, the MZ-score $\lambda_i$ of $x_i \in X$ is defined as
\begin{equation}
\lambda_i := \frac{x_i - \operatorname{Med}(X)}{\sigma}.
\label{eq:mzscore}
\end{equation}
\end{definition}


\begin{algorithm}[h!]
\caption{LASA}
\label{alg:lasa_defense}
\begin{algorithmic}[1]

\State \textbf{Inputs:} Set of $n$ local model updates $\{\Delta_i\}_{i=1}^n$, number of model layers $L$, sparsification parameter $k$, magnitude-based radius $\lambda_m$, direction-based radius $\lambda_d$

\For{$i = 1$ to $n$}
    \State $\hat{\Delta}_i \leftarrow f_{\text{top-}k}(\Delta_i)$
\EndFor

\For{layer $l = 1$ to $L$}
    \State $S \leftarrow \emptyset$ 
    
    \State $\omega^l \leftarrow \{\|\hat{\Delta}_i^l\|_2\}_{i=1}^n$ 
    \State $\rho^l \leftarrow \{\text{PDP}(\hat{\Delta}_i^l)\}_{i=1}^n$ \Comment{via Eq.~\eqref{eq:pdp}}

    \For{$i = 1$ to $n$}
        \State $\lambda_{i,m}^l \leftarrow \text{MZ-score}(\omega_i^l, \omega^l)$ \Comment{via Eq.~\eqref{eq:mzscore}}
        \State $\lambda_{i,d}^l \leftarrow \text{MZ-score}(\rho_i^l, \rho^l)$ \Comment{via Eq.~\eqref{eq:mzscore}}

        \If{$|\lambda_{i,m}^l| \le \lambda_m$ \textbf{and} $|\lambda_{i,d}^l| \le \lambda_d$}
            \State $S \leftarrow S \cup \{i\}$
        \EndIf
    \EndFor

    \If{$S = \emptyset$} 
        \State $S \leftarrow \{1, \dots, n\}$
    \EndIf

    \State $\bar{\Delta}^l \leftarrow \frac{1}{|S|} \sum_{i \in S} \hat{\Delta}_i^l$
\EndFor

\State \Return $\bar{\Delta} = [\bar{\Delta}^1, \dots, \bar{\Delta}^L]$

\end{algorithmic}
\end{algorithm}

By modifying HSA to target less-utilized parameter locations, we successfully evade LASA's defenses. Our counter-attack (Algorithm \ref{alg:lasa_sparse_attack}) identifies parameter indices with the smallest magnitudes from benign clients (line 12) and allocates attack budget using a single sparsity ratio $\beta$: the bottom $\beta$ fraction of locations receives the large perturbation $z_{\text{large}}$, while all remaining coordinates receive the smaller perturbation $z_{\max}$. We then apply layer-wise stigmatization and inverted mean sign masking to bypass LASA's detection mechanisms (lines 14-19).  

\begin{algorithm}[t]
    \small
    \caption{Counter-attack Formation for LASA}
        \label{alg:lasa_sparse_attack}
    \begin{algorithmic}[1]
     \State \textbf{Inputs:} Benign gradients $\{\mathbf{g}_1, \ldots, \mathbf{g}_B\}$, layer boundaries $\mathcal{L}$, sparsity ratio $\delta$, attack scales $z_{\max}$, $z_{\text{large}}$, attack ratio $\beta$
     \State \textbf{Outputs:} Adversarial perturbation $\mathbf{p}$
     \State benign gradients: $G \leftarrow \text{stack}(\{\mathbf{g}_1, \ldots, \mathbf{g}_{k_b}\})$ 
    \State $\mathbf{s} \leftarrow \text{sign}(\text{mean}(G, \text{dim}=0))$
    \State $\sigma \leftarrow \text{std}(G)$ 

     \For{each gradient $\mathbf{g}_i$ in $G$}
        \State $I_i \leftarrow \text f_{bottom-k}(|\mathbf{g}_i|, \lfloor \delta \cdot d \rfloor)$
     \EndFor
     \State $\mathbf{c} \leftarrow \mathbf{0}^d$
     \For{each index set $I_i$}
        \State $\mathbf{c}[I_i] \mathrel{+}= 1$
     \EndFor
     \State $\mathbf{m} \leftarrow \sigma \cdot z_{\max} \cdot \mathbf{1}^d$
     \State $J \leftarrow \text f_{bottom-k}(\mathbf{c}, \lfloor \beta \cdot d \rfloor)$ 
     \State $\mathbf{m}[J] \leftarrow \sigma \cdot z_{\text{large}}$ 
     \State $\mathbf{m} \leftarrow \mathbf{m} \odot \mathbf{s}$ 
     \State Split $\mathbf{m}$ into layers: $\{\mathbf{m}^{(\ell)}\}_{\ell \in \mathcal{L}}$
     \State Split $G$ into layers: $\{G^{(\ell)}\}_{\ell \in \mathcal{L}}$
     \For{each layer $\ell$}
        \State $\mathbf{n}^{(\ell)}[i] \leftarrow \left\| G^{(\ell)}_{i,:} \right\|_2 \quad \forall i$
        \State $\tilde{n}^{(\ell)} \leftarrow \text{median}(\mathbf{n}^{(\ell)})$ 
        \State $\mathbf{p}^{(\ell)} \leftarrow -\frac{\mathbf{m}^{(\ell)}}{\|\mathbf{m}^{(\ell)}\|_2} \cdot \tilde{n}^{(\ell)}$ 
     \EndFor
     \State $\mathbf{p} \leftarrow \text{concat}(\{\mathbf{p}^{(\ell)}\}_{\ell \in \mathcal{L}})$
     \State \Return $\mathbf{p}$
    \end{algorithmic}
    \end{algorithm}

\begin{table*}[]
\caption{Test accuracy (\%) of state-of-the-art robust aggregators against HSA, DHSA, and the proposed C-LASA counter-attack on the CIFAR-10 classification task under both IID and non-IID data distributions. Results are reported as mean $\pm$ standard deviation over multiple runs. Entries marked with ``-'' indicate that the corresponding attack--aggregator combination was not evaluated. We employ HSA configuration of $\delta = 5 \times 10^{-2}$, $\delta^{max}_{cw1} = 0.25$,  $\delta^{max}_{fc}= 0.25$}
\label{tab:newAGGs}
\resizebox{\textwidth}{!}{%
\begin{tabular}{@{}cccccccccc@{}}
\toprule
Dist & ATK / DEF & FoundationFL\cite{FoundationFL} & LASA\cite{lasa} & DnC\cite{shejwalkar2021manipulating}& FedREDefence\cite{xie2024fedredefense} & FL-Defender\cite{jebreel2023fl} & Flame\cite{nguyen2022flame} & FLDetector\cite{zhang2022fldetector} & SkyMask\cite{yan2024skymask} \\ \midrule
\multirow{3}{*}{IID} & HSA & 83.12 $\pm$ 0.7 & 85 $\pm$ 1.3 & 32.64 $\pm$  5.4 & 51.78 $\pm$ 2.1 & 86.12 $\pm$ 0.3 & 48.12 $\pm$ 3.6 & 21 $\pm$ 1.2 & 66.02 $\pm$ 3.6 \\
 & DHSA & 67.6 $\pm$ 4.4 & 83 $\pm$ 1.8 &  &  &  &  &  &  \\
 & C-LASA & - & 55  $\pm$ 4.5 & - & - & - & - & - & - \\ \midrule
\multirow{3}{*}{non-IID} & HSA & 11.2 & 70.3 $\pm$ 3.1 & 10.0 $\pm$ 0.0 & 10.0 $\pm$ 0.0 & 10.0 $\pm$ 0.0 & 16.57 $\pm$ 2.3 & 10.0 $\pm$ 0.0 & 10.6 $\pm$ 0.5 \\
 & DHSA & 12.4 & 73.4  $\pm$ 2.6 &  &  &  &  &  &  \\
 & C-LASA & - & 10.8 $\pm$ 0.6 & - & - & - & - & - & - \\ \bottomrule
\end{tabular}%
}
\end{table*}

Table~\ref{tab:newAGGs} reports average test accuracies under our proposed attacks. LASA effectively defends against HSA and DHSA because it was designed to counter structurally similar attacks (Min-Sum, Min-Max, ALIE), achieving only $\sim$15\% accuracy reduction in non-IID scenarios. However, our C-LASA counter-attack exposes LASA's vulnerability: it causes complete model divergence in non-IID settings and reduces test accuracy by 30\% in IID scenarios. Against FoundationFL, both DHSA and HSA struggle in IID settings but successfully diverge the model in non-IID scenarios.

\subsection{Evaluation Against Additional State-of-the-Art-Defenses}
We further evaluated the robustness of our proposed HSA and DHSA attacks against recently proposed state-of-the-art (SOTA) defence mechanisms. 
\begin{itemize}
    \item \textbf{DnC (Divide-and-Conquer) \cite{shejwalkar2021manipulating}:} This defense leverages spectral analysis to identify and remove Byzantine updates. It performs randomized dimensionality reduction followed by singular value decomposition (SVD) to filter out updates that significantly deviate from the principal components of the benign gradient distribution.
    \item \textbf{FLAME \cite{nguyen2022flame}:} Designed primarily to mitigate backdoors, FLAME utilizes a combination of HDBSCAN-based clustering, adaptive weight clipping, and differential privacy (DP) noise injection to neutralize potentially malicious local model updates.
    \item \textbf{FLDetector \cite{zhang2022fldetector}:} This mechanism attempts to detect malicious clients by analyzing the consistency of local model updates across multiple training rounds, focusing on the temporal trajectory of the learning process.
    \item \textbf{FedREDefense \cite{xie2024fedredefense}:} A very recent framework that utilizes model update reconstruction error to distinguish between benign and malicious updates, effectively filtering out poisoning attempts that do not align with the reconstructed global distribution.
    \item \textbf{SkyMask \cite{yan2024skymask}:} An attack-agnostic defense that employs fine-grained learnable masks to protect the global model. It identifies and masks parameters that are susceptible to adversarial manipulation while maintaining the model's utility.
    \item \textbf{FL-Defender \cite{jebreel2023fl}:} This framework focuses on combating targeted attacks by integrating multi-layered filtering and distance-based analysis to protect the integrity of the federated learning process.
\end{itemize}

\section{Addressing Numerical Instability in Simulations} \label{sec:IQR}
As discussed in Section \ref{sec:perf_HSA_noniid} and illustrated in Fig.~\ref{fig:z_scalesDir}, DHSA exhibits reduced $z_1$ values in non-IID scenarios compared to IID settings, which adversely affects attack performance. The root cause of this degradation lies in the variance explosion phenomenon at sparse mask locations under data heterogeneity.

\indent In non-IID federated learning, client datasets exhibit significant statistical heterogeneity, leading to high variance in gradient estimates across different clients. When HSA targets specific network parameters through the sparse mask $\mathbf{c}$, the variance $\boldsymbol{\sigma}_t$ at these locations can grow disproportionately large due to divergent local optimization trajectories. Specifically, for the masked indices where $\mathbf{c}[i]=1$, even though we only perturb a small fraction of parameters (sparse attack), the product $z_2 \mathbf{c} \odot \boldsymbol{\sigma}_t$ in \eqref{eq:DHSA-opt} can produce extremely large values when individual variance estimates $\boldsymbol{\sigma}_t[i]$ become inflated.

\begin{figure}
    \centering    \includegraphics[width=0.8\linewidth]{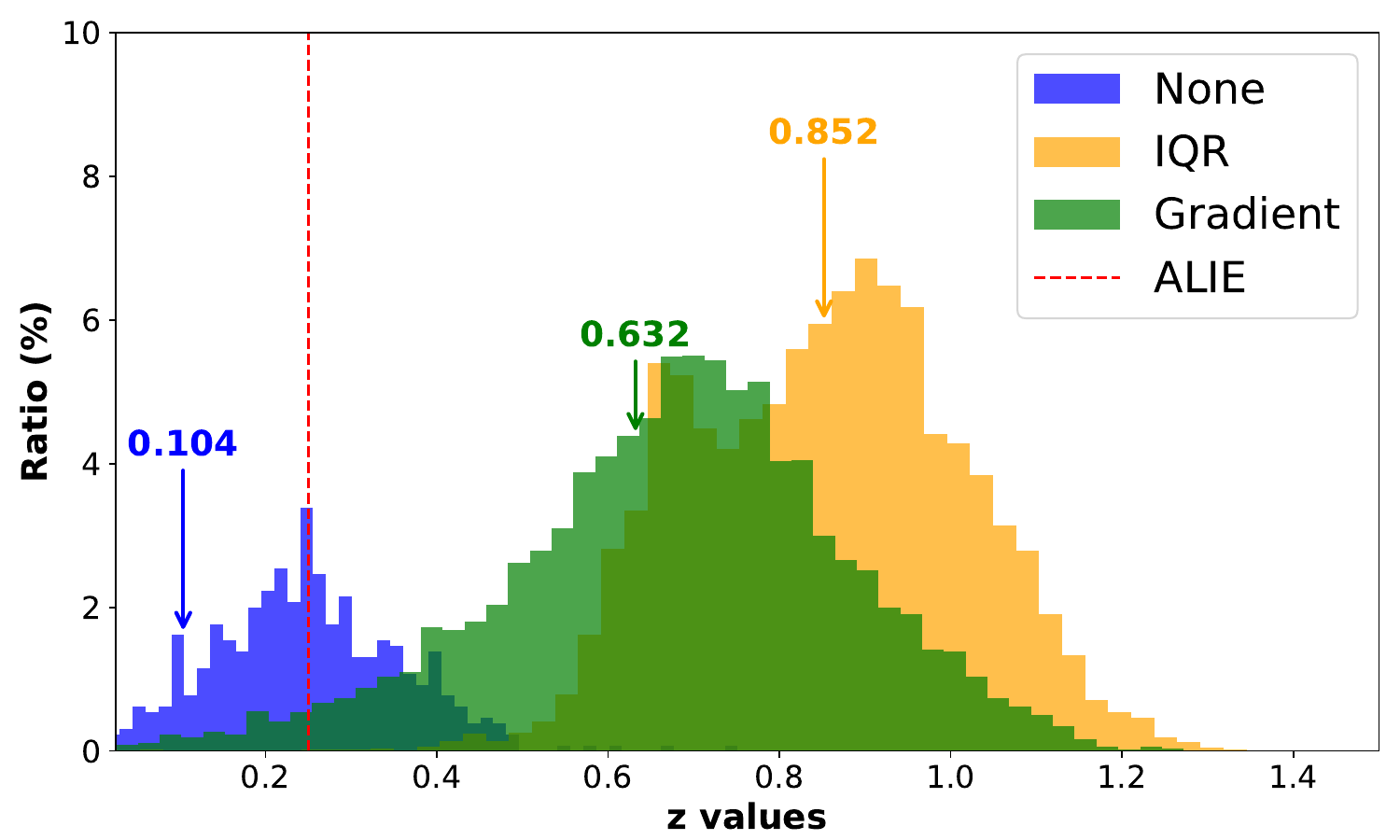}
    \caption{Histogram of optimized $z$ values generated by DHSA using IQR and Gradient-based variance stabilization methods in the \textit{non-IID} setting against the CM aggregator.}
    \label{fig:z_cm_dir}
    \vspace{-5mm}
\end{figure}

\indent This variance explosion directly undermines the DHSA optimization in \eqref{eq:DHSA-opt}. Recall that DHSA maximizes $z_1$ subject to the constraint that the $L_2$ distance between the malicious update and benign updates remains below $\Delta_{th}$. When a few coordinates in $z_2 \mathbf{c} \odot \boldsymbol{\sigma}_t$ exhibit exploded values, they consume the majority of the perturbation budget, forcing the optimizer to drastically reduce $z_1$ to satisfy the distance constraint. Consequently, the attack loses its effectiveness on the remaining $(1-\mathbf{c})$ coordinates, which constitute the majority of the parameter space.

\indent To address this numerical instability, we employ two complementary outlier detection and mitigation strategies: the Interquartile Range (IQR) method and the gradient-based elbow method. Both approaches identify and regularize anomalously large variance estimates at sparse mask locations before they are used in attack construction.

\begin{figure}
    \centering    \includegraphics[width=0.8\linewidth]{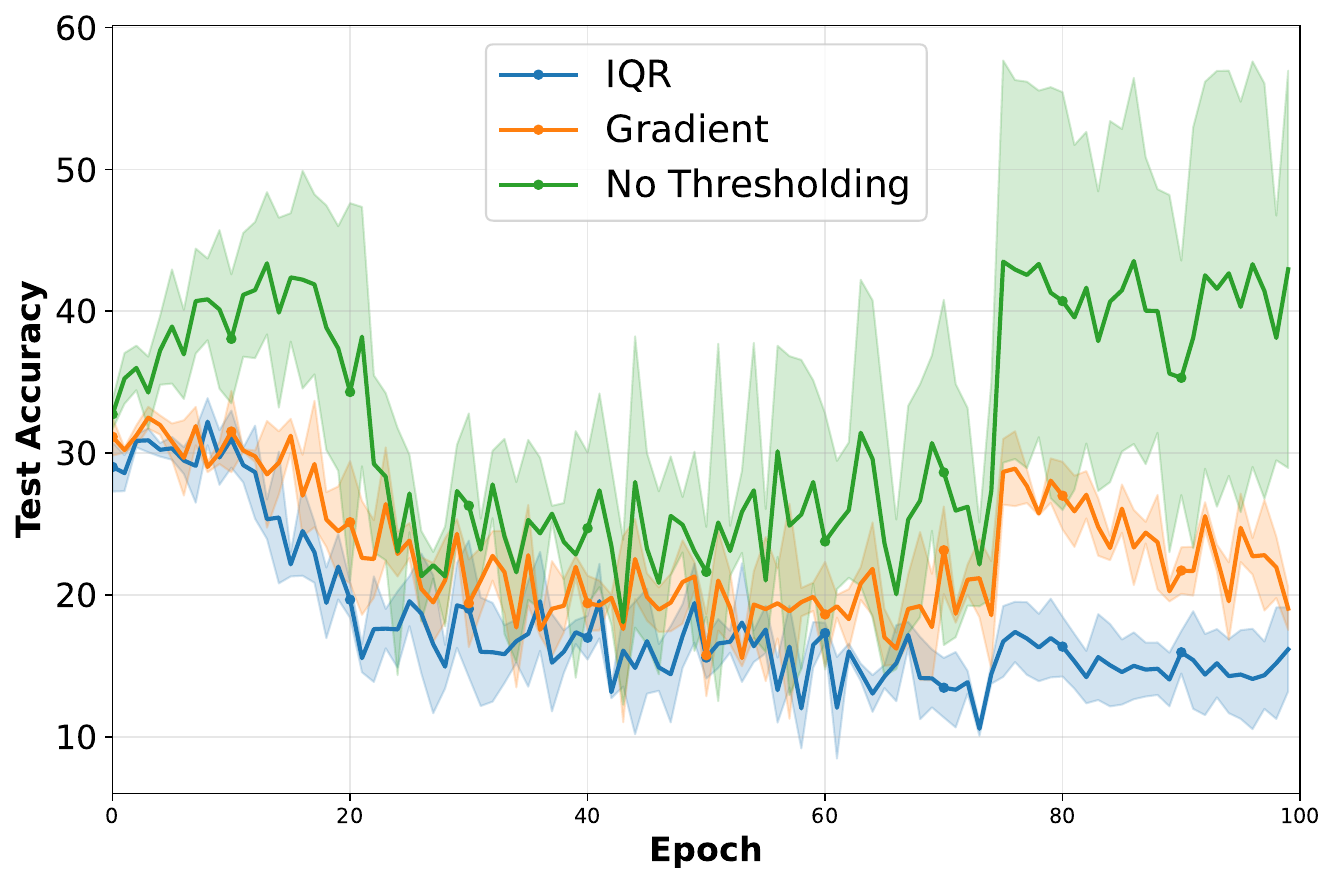}
    \caption{Test accuracy under DHSA using IQR and Gradient-based variance stabilization methods in the \textit{non-IID} setting against the CM aggregator. Shaded regions represent the standard deviation across multiple runs.}
    \label{fig:cm_std_th}
\end{figure}

\indent \textbf{IQR-based thresholding:} The IQR method \cite{IQR-stats} provides a robust statistical approach to detect outliers. For the variance values at masked locations $\{\boldsymbol{\sigma}_t[i] : \mathbf{c}[i]=1\}$, we compute the first quartile $Q_1$ and third quartile $Q_3$, and define the interquartile range as $\text{IQR} = Q_3 - Q_1$. Values exceeding the upper fence $Q_3 + 1.5 \times \text{IQR}$ are considered outliers and clipped to this threshold. This technique is particularly effective when the majority of variance estimates are well-behaved, and only a small fraction exhibits extreme values due to data heterogeneity among clients.

\indent \textbf{Gradient-based elbow method:} While IQR relies on fixed quantile-based thresholds, the elbow method \cite{L-method,L-method2}  adaptively identifies the transition point where variance values begin to grow abnormally. We sort the variance values and compute consecutive differences. The elbow point is identified where the ratio of consecutive differences changes most dramatically, indicating a sharp increase in variance. This adaptive threshold is especially useful when the proportion of outliers varies across training iterations.

\indent In Fig. \ref{fig:z_cm_dir}, we illustrate the optimized $z$ values generated by the DHSA attack. Both IQR and Gradient-based methods demonstrate substantial improvements over vanilla DHSA, with IQR achieving higher and more stable $z$ values. Ultimately, both methods preserve attack effectiveness by ensuring that individual coordinate perturbations remain bounded while maintaining sufficient attack strength across the sparse mask. By preventing variance explosion at targeted locations, DHSA can allocate the perturbation budget more efficiently, thereby increasing $z_1$ values and improving attack success rates in non-IID scenarios, as demonstrated in Fig. \ref{fig:cm_std_th}. Against the CM aggregator, vanilla DHSA reduces test accuracy by only $\sim 35\%$ with high standard deviation (shown as shaded regions). In contrast, both IQR and Gradient-based methods provide more stable attacks while achieving an additional $\sim 35\%$ reduction in test accuracy compared to vanilla DHSA. Overall, in non-IID scenarios, Byzantine clients can significantly enhance the attack performance of both HSA and DHSA by employing variance stabilization techniques to ensure numerical stability.

\section{Illustrations of FORCE sparsification with Layer Constraints} \label{sec:figsForce}

\begin{figure*}[t!]
    \centering
    \includegraphics[width=\textwidth]{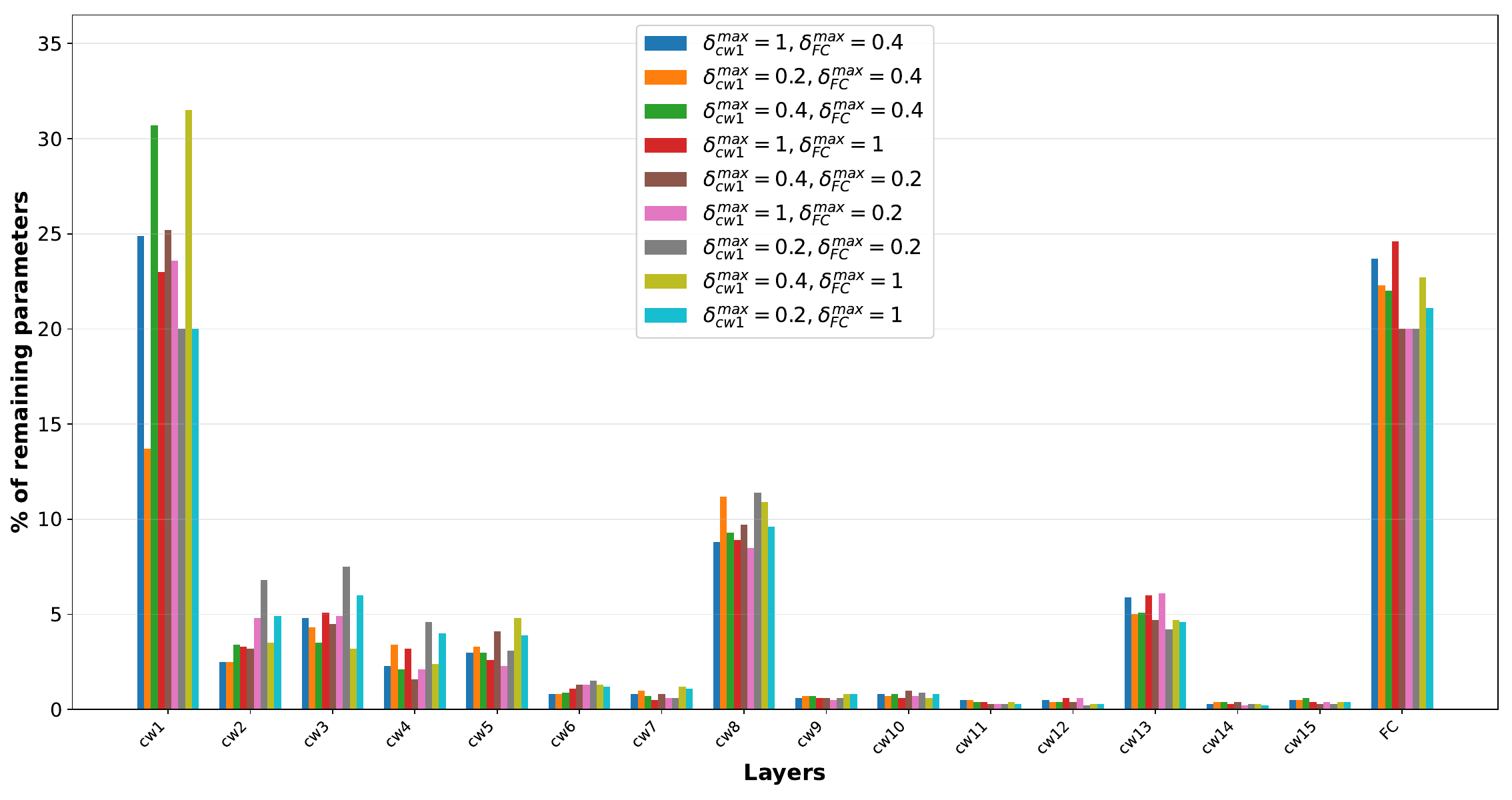}
    \caption{Layer-wise sparsity density distribution for $\delta=5\times10^{-3}$ global pruning factor. The figure illustrates how the pruning budget is distributed across network layers, highlighting the impact of constraining maximum sparsity in layers ($\delta^{max}_{cw1}$ for the first Conv layer, $\delta^{max}_{fc}$ for the last FC layer)}.
    \label{fig:layers_prune_comperison005}
\end{figure*}

\begin{figure*}[t!]
    \centering
    \includegraphics[width=\textwidth]{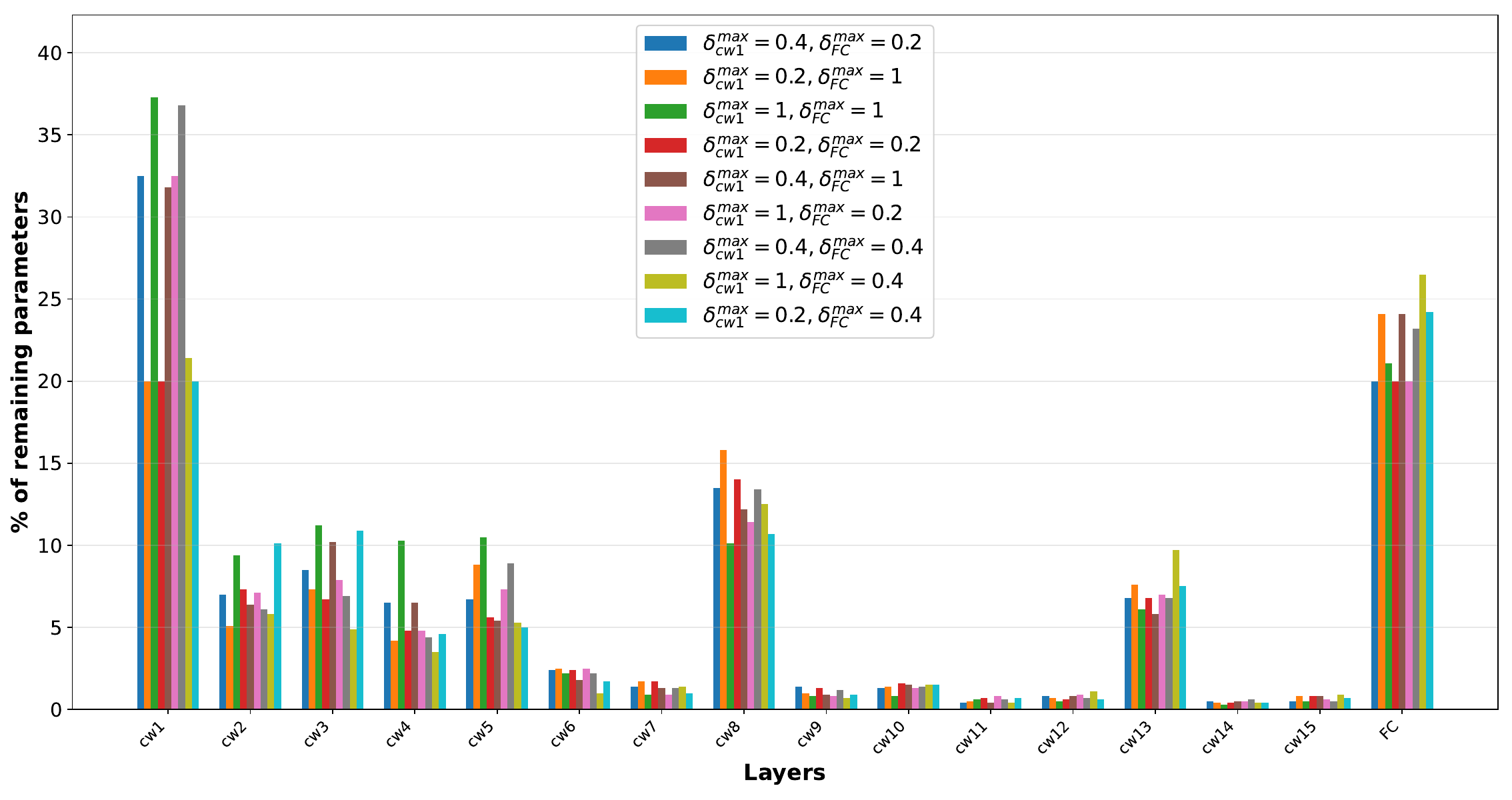}
    \caption{Layer-wise sparsity density distribution for $\delta=1\times10^{-2}$ global pruning factor. The figure illustrates how the pruning budget is distributed across network layers, highlighting the impact of constraining maximum sparsity in layers ($\delta^{max}_{cw1}$ for the first Conv layer, $\delta^{max}_{fc}$ for the last FC layer)}.
    \label{fig:layers_prune_comperison01}
\end{figure*}

\begin{figure*}[t!]
    \centering
    \includegraphics[width=\textwidth]{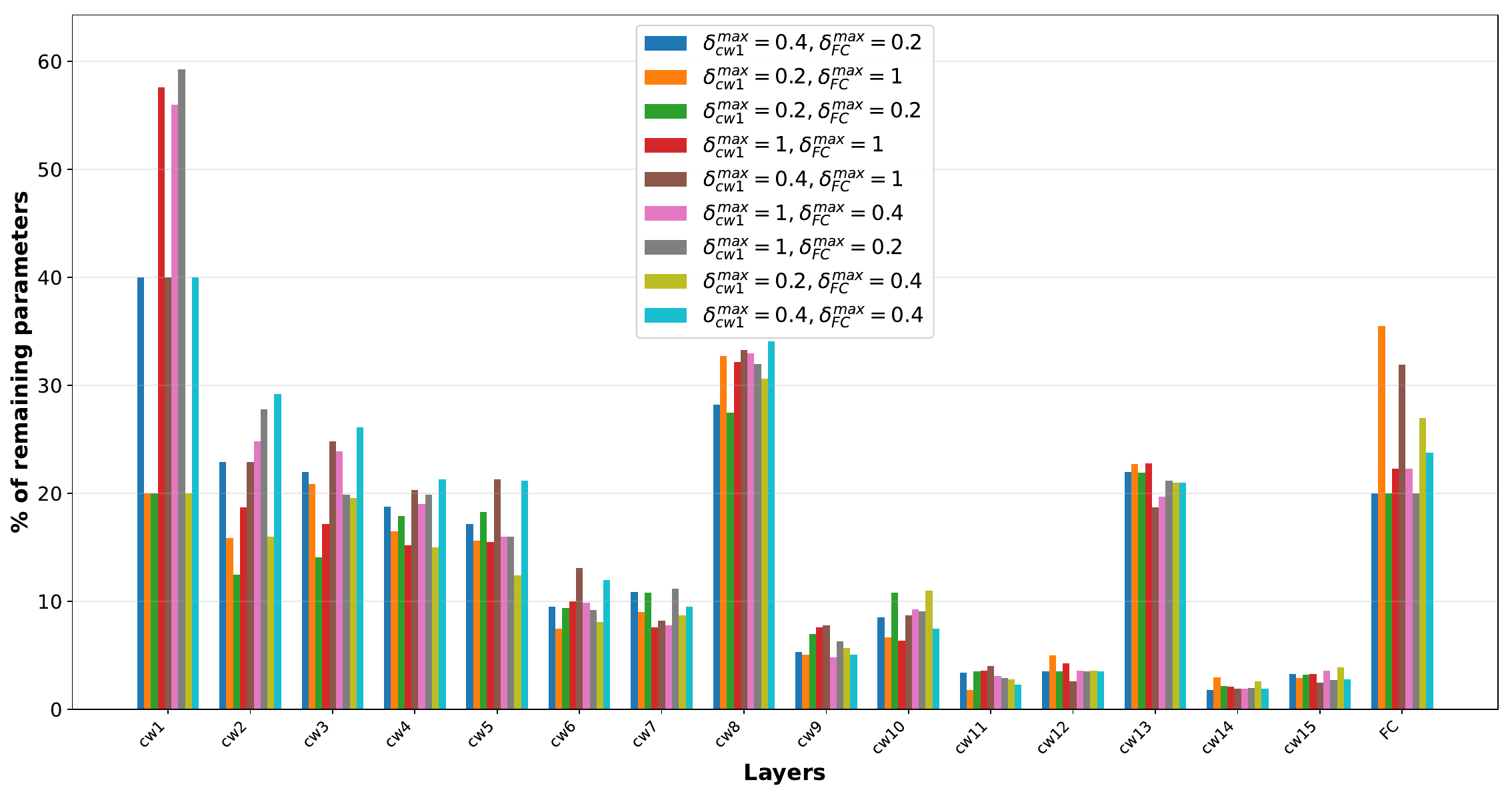}
    \caption{Layer-wise sparsity density distribution for $\delta=5\times10^{-2}$ global pruning factor. The figure illustrates how the pruning budget is distributed across network layers, highlighting the impact of constraining maximum sparsity in layers ($\delta^{max}_{cw1}$ for the first Conv layer, $\delta^{max}_{fc}$ for the last FC layer)}.
    \label{fig:layers_prune_comperison05}
\end{figure*}

\begin{figure*}[t!]
    \centering
    \includegraphics[width=\textwidth]{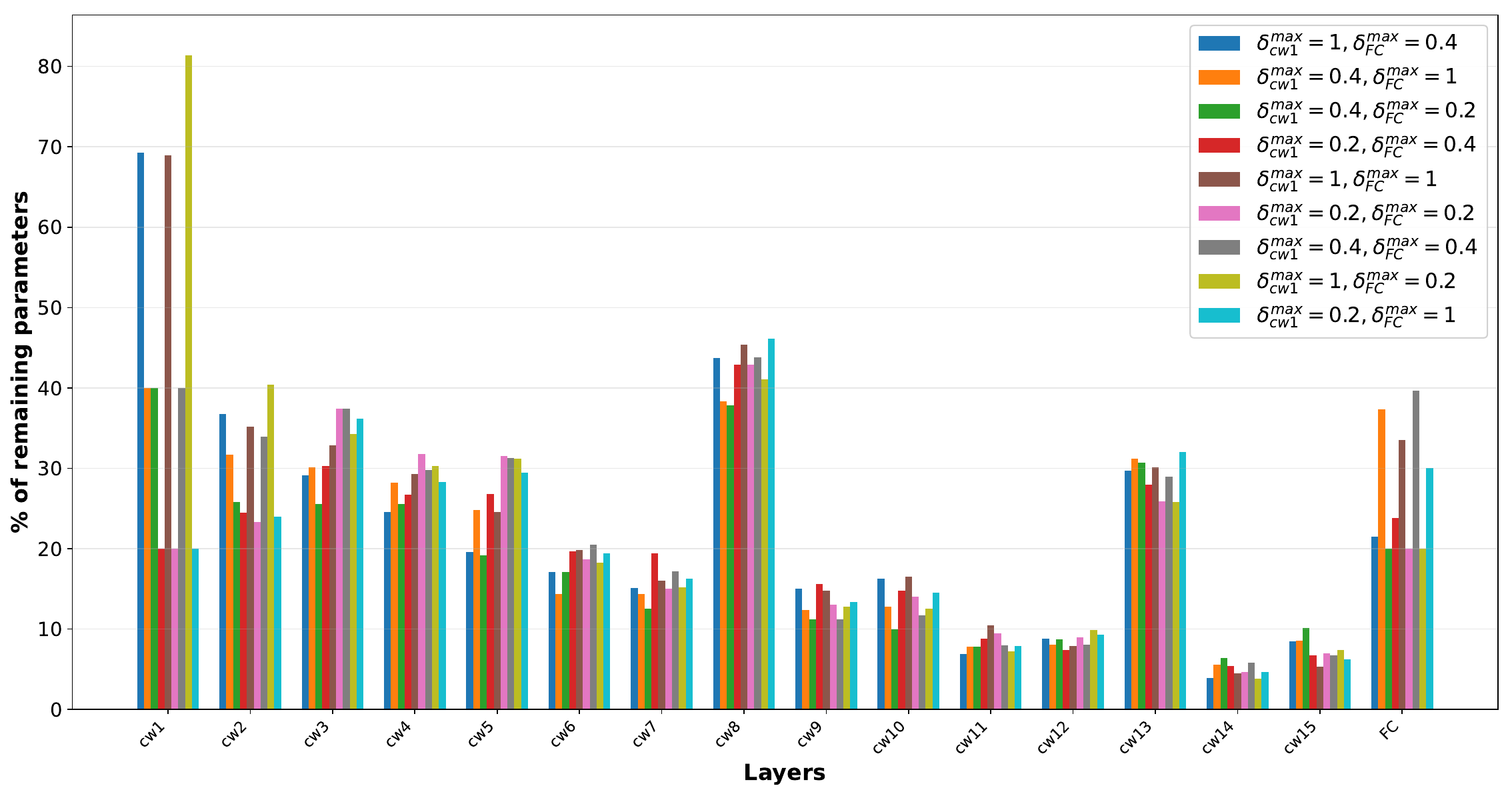}
    \caption{Layer-wise sparsity density distribution for $\delta=1\times10^{-1}$ global pruning factor. The figure illustrates how the pruning budget is distributed across network layers, highlighting the impact of constraining maximum sparsity in layers ($\delta^{max}_{cw1}$ for the first Conv layer, $\delta^{max}_{fc}$ for the last FC layer)}.
    \label{fig:layers_prune_comperison1}
\end{figure*}

\end{document}